%% file: arxiv.tex
\documentclass[10pt,twocolumn,letterpaper]{article}
\usepackage{iccv}

\usepackage{times}
\usepackage{epsfig}
\usepackage{graphicx}
\usepackage{amsmath}
\usepackage{amssymb}

\usepackage[pagebackref=true,breaklinks=true,letterpaper=true,colorlinks,bookmarks=false]{hyperref}

\iccvfinalcopy
\def\iccvPaperID{3740} 

\usepackage{mmstyle}
\usepackage{subfig}
\usepackage{dsfont}
\usepackage[dvipsnames]{xcolor}

\usepackage{pifont}
\usepackage{soul}
\newcommand{\xmark}{\ding{55}}%
\usepackage{multirow}

\usepackage{algorithm, algpseudocode} 
\algnewcommand\algorithmicinput{\textbf{INPUT:}}
\algnewcommand\INPUT{\item[\algorithmicinput]}
\algnewcommand\algorithmicoutput{\textbf{OUTPUT:}}
\algnewcommand\OUTPUT{\item[\algorithmicoutput]}

\graphicspath{{figures/}}

\begin{document}

\title{\vspace{-0pt}Transcript to Video: Efficient Clip Sequencing from Texts}

\author{Yu Xiong$^{1}$ \quad Fabian Caba Heilbron$^2$ \quad Dahua Lin$^1$\\	
		$^1$The Chinese University of Hong Kong \quad
		$^2$Adobe Research\\	
		{\tt\small \{xy017,dhlin\}@ie.cuhk.edu.hk}\hspace{10pt}	
		{\tt\small caba@adobe.com}
}
\twocolumn[{ 
 \renewcommand\twocolumn[1][]{#1} 
 \maketitle 
 \begin{center} 
 \centering 
 \vspace{-16px}
 \includegraphics[width=0.88\linewidth]{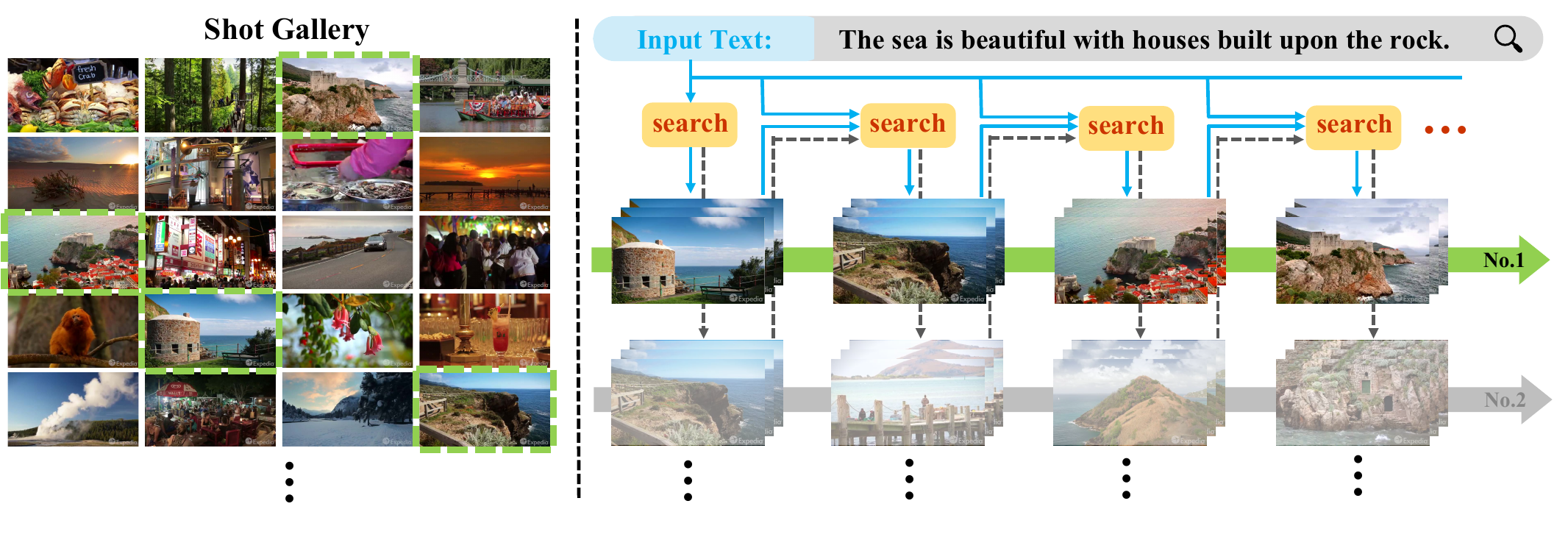} 
 \vskip -0.4cm 
  \vspace{-6px}
 \captionof{figure}{\small
 \textbf{Automatic multi-shot clip creation (or clip sequencing)} from texts. 
 Our goal is to create multi-shot video clips from a gallery of shots given input text queries. 
 The shot gallery (left) is a large set of stock footage comprising varied video content. 
 At inference time (right), our Transcript-to-Video model takes as input a query text and retrieves a series of shots from the gallery.
 Transcript-to-Video ensures that the generated sequence of shots has temporal coherence with 
 a plausible sequencing style.
 }
 \label{fig:teaser} 
 \end{center} 
}]

\input{articles/abstract.tex}
\input{articles/introduction.tex}

\input{articles/related.tex}
\input{articles/benchmark.tex}
\input{articles/method.tex}

\input{articles/experiment.tex}

\input{articles/conclusion.tex}
\input{articles/acknowledgement.tex}

{\small
\bibliographystyle{ieee_fullname}
\bibliography{egbib}
}

\clearpage

\appendix
\renewcommand\thefigure{\thesection\arabic{figure}} 
\setcounter{figure}{0} 
\renewcommand\thetable{\thesection\arabic{table}} 
\setcounter{table}{0} 

{\Large \noindent\textbf{Supplementary Material}\par}
\vspace{10pt}

We provide supplementary materials both in \emph{this PDF file} and in \emph{the video file}.
In this file, we show: 
\begin{itemize}
    \item Details of user studies in Sec.~\ref{supp:user-study}.
    \item Experiment section with implementation details and more experimental analysis in Sec.~\ref{supp:exp}.
    \item Statistics and examples from WeTravel in Sec.~\ref{supp:dataset}.
    \item Visualization results of proposed methods in Sec.~\ref{supp:visualization}.
    \item Other extra information in Sec.~\ref{supp:extra}.
\end{itemize}
For the demo video file, it comprises three parts:
\begin{itemize}
    \item A story video generated by multiple query sentences; the story video transcript could be found at Sec.~\ref{supp:extra}.
    \item Introduction of our task and framework.
    \item Visualization results of proposed methods.
\end{itemize}
Please visit our project page for demo video: \url{http://www.xiongyu.me/projects/transcript2video/}

\setcounter{table}{0} 
\setcounter{figure}{0} 
\input{supp_articles/userstudy.tex}

\setcounter{table}{0} 
\setcounter{figure}{0} 
\input{supp_articles/exp.tex}

\setcounter{table}{0} 
\setcounter{figure}{0} 
\input{supp_articles/dataset.tex}
\clearpage

\setcounter{table}{0} 
\setcounter{figure}{0} 
\input{supp_articles/visualization.tex}

\setcounter{table}{0} 
\setcounter{figure}{0} 
\input{supp_articles/extra}

\input{supp_articles/fig_htm_vs_wt}

\input{supp_articles/fig_greedy_vs_beam}
\input{supp_articles/fig_p_vs_i}
\input{supp_articles/fig_all}
\input{supp_articles/fig_mil_nce_i}
\input{supp_articles/fig_whole_test_set}

\input{supp_articles/fig_vimeo}

\end{document}

%% file: articles/abstract.tex

\begin{abstract}
\vspace{-8pt}
Among numerous videos shared on the web, well-edited ones always attract more attention. 
However, it is difficult for inexperienced users to make well-edited videos 
because it requires professional expertise and immense manual labor. 
To meet the demands for non-experts, we present Transcript-to-Video -- a weakly-supervised framework 
that uses texts as input to automatically create video sequences from an extensive collection of shots. 
Specifically, we propose a Content Retrieval Module and a Temporal Coherent Module to 
learn visual-language representations and model shot sequencing styles, respectively. 
For fast inference, we introduce an efficient search strategy for real-time video clip sequencing. 
Quantitative results and user studies demonstrate empirically that the proposed learning framework 
can retrieve content-relevant shots while creating plausible video sequences in terms of style. 
Besides, the run-time performance analysis shows that our framework can support real-world applications. 
We will release codes and models. Project page: \url{http://www.xiongyu.me/projects/transcript2video/}
\end{abstract}

%% file: articles/introduction.tex
\vspace{-20pt}
\section{Introduction}
\vspace{-3px}
\label{sec:introduction}

%
In recent years, a large number of Apps flooded the marketplace with the feature of video sharing.
In the era of ``everybody could be a content creator'', the technique of selecting,
editing, and piecing together separate sections of footage to form a video becomes the key to video making.
Mastering superb video editing skills, film editors and professional creators can always 
precisely express abstract concepts and comprehensive content by assembling series of shots 
into videos~\cite{gfl_grammer_film_lang,afiTrain,edwin,wikimontage,wikisovietmontage}.
With the emergence of video sharing, this kind of video editing technique is also becoming 
a standard tool for aspiring creators to tell their stories. 
However, and despite their increasing popularity, compositing multi-shot clips is still 
a challenging task that requires video editing expertise and substantial manual work.
This paper aims to advance computational video editing by developing a method (see Fig.~\ref{fig:teaser}) 
for automatic clip sequencing from texts.
To be specific, to automatically compose multiple shots into a video clip, 
using text as the query and a hefty stock video gallery.

%
Over the past few years, multiple studies have been devoted to computational video editing. 
Some recent works have focused on the machine-assisted creation of 
video sequences~\cite{dialog_montage,quickcut,write_a_video} using transcripts. 
The pioneering work QuickCut~\cite{quickcut} offers an interactive editing tool 
to navigate and quickly create story outlines from raw footage. 
QuickCut shows tremendous benefits at speeding up the editing process; 
however, it still demands excessive work from expert video editors' to select and arrange the final video shots. 
Closely related to our work, Write-A-Video~\cite{write_a_video} proposes a method 
to retrieve relevant shots using Visual Semantic Embedding~\cite{vsepp}, 
and a rule-based optimization to order shots. 
Unfortunately, Write-A-Video requires users to specify keywords on query sentences, 
which are then matched across a similarly tagged footage collection. 
These constraints limit the vocabulary that can be used in the queries and the overall approach's scalability. 
Despite the progress, we argue that automatic clip sequencing from  text remains challenging, 
especially when the input queries and gallery shots are unlabeled. 
To tackle these challenges, we propose a weakly-supervised framework for 
learning vision-language embeddings and sequencing styles on a newly constructed unlabeled video dataset. 
Unlike previous works that are semi-automated, our method presents an efficient and automated
inference pipeline, making the application of clip sequencing more practically appealing.

%
In this work, we develop a new learning-based framework for automated clip sequencing. 
Primarily, a good multi-shot clip that illustrates query text should have the following properties:
(a) it should cover the semantics of the given text visually, and
(b) it should ensure a smooth transition between shots. 
Additionally, to be practical, it should also efficiently generate sequences in response to the query, 
even when the gallery set is large.
We achieve this goal by proposing two weakly supervised modules that learn 
a visual-language embedding and a temporal smoothness preserving representation during training. 
To be specific, the training framework integrates two key components. 
First, a \emph{Content Retrieval Module} for shot retrieval that learns 
a joint text-video embedding via noise-contrastive estimation \cite{milnce}. 
And second, a \emph{Temporal Coherence Module} that learns how to preserve 
shot-level transition smoothness using a new pretext task. 
Furthermore, we propose a modified beam search inference to create multi-shot clip in real-time.

%
It is not a secret that weakly supervised models are data-hungry \cite{milnce,mmt}. 
Therefore, to feed our models, we need a large-scale (unlabeled) dataset of well-edited multi-shot clips. 
However, and despite significant progress in video understanding datasets \cite{howto100m,youcook2,vlog}, 
no existing data source fits our demand. For instance, 
there might exist datasets depicting human actions \cite{anet,knet} or instructional videos \cite{howto100m,how2}, 
but they either are single-clip or contain lots of talking heads, respectively. 
These properties limit their value to weakly supervised learning of visual-language representation and shot sequencing styles. 
We cope with the lack of datasets by introducing WeTravel, a novel dataset of travel videos 
containing rich, diverse, and extensive edited video content.
The construction pipeline leverages publicly available videos, automated transcripts, 
and off-the-shelf shot-boundary detectors to collect
shot sequences paired with noisily and weakly aligned text.

%
\noindent\textbf{Contributions.} Our key idea is to leverage weak supervision and 
large-scale data to build an automated method for clip sequencing. 
Our work brings three contributions:\\
\textbf{(1)} To the best of our knowledge, we are the first to study the task of 
automated clip sequencing from a learning-based perspective. 
To do so, we establish a benchmark for multi-shot clip creation with a new dataset called WeTravel.\\
\textbf{(2)} We propose a weakly supervised framework that effectively 
learns both a joint visual-language representation and shot sequencing styles while efficiently assembling multi-shot videos during inference.\\
\textbf{(3)} We conduct throughout studies on new downstream tasks based on WeTravel, 
where we validate our approach's effectiveness and efficiency over baseline methods. 
We notice similar findings in perceptual studies involving experienced and non-experienced editors.

%% file: articles/related.tex

\section{Related Work}
\label{sec:related}
\vspace{-3px}

 \begin{figure*}[!t]
    \vspace{-7pt}
	\centering
	\includegraphics[width=0.9\linewidth]{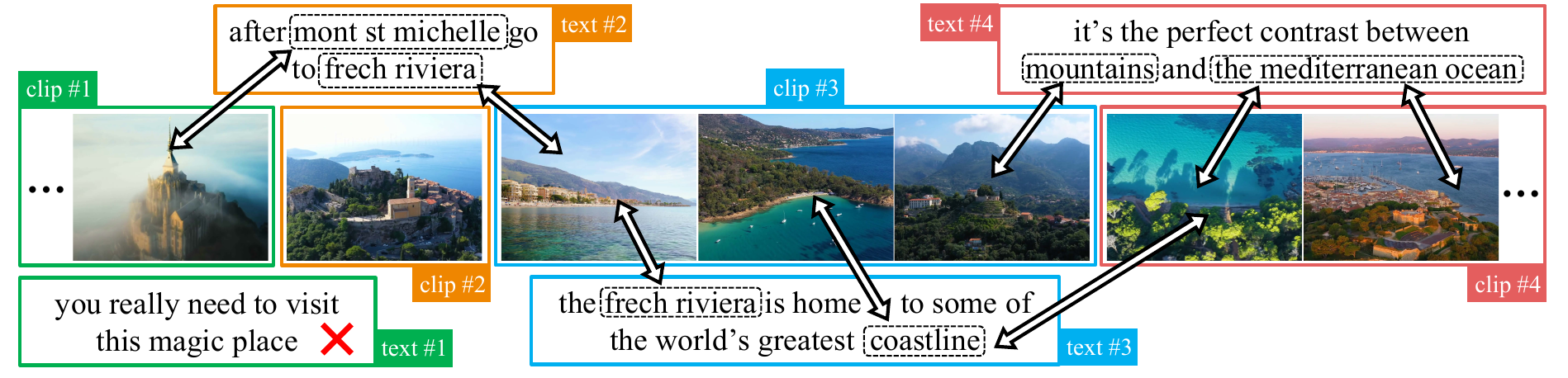}
	\vspace{-10pt}
	\caption{\small
    	\textbf{The WeTravel dataset.} We showcase a typical example from the dataset. 
    	Each color represents a text-clip pair. Note that the text descriptions are not 
    	directly aligned with its corresponding clip in many cases, but rather with a neighbor. 
    	For instance, the last transcript text (red) mentions the concept of ``mountains'', 
    	which we observe in the previous clip. 
    	Moreover, there might exist cases where no one-to-one mapping exists 
    	between the text and the visuals (illustrated with ``\xmark'').
	}
	\label{fig:benchmark_sample}
	\vspace{-13pt}
\end{figure*}

%
\noindent{\textbf{Video Summarization.}}
Video summarization ~\cite{rl_vs_zhou2017deep,vs_gong2014diverse,vs_sharghi2018improving,vs_li2018local,tvsum,vs_otani2019rethinking} 
aims to shorten a long video while keeping the most salient information. 
To reduce the subjectivity in the results, recent approaches allow a text query to 
control the summary generation~\cite{query_vs,query_control_vs, choi2018contextually}.
Despite addressing a similar goal, multi-shot clip creation is different from query-based video summarization since: 
(i) video summarization approaches are frame-based and the gallery frames are from a single long video; 
(ii) it does not require temporal coherence across shots 
since the generated summaries directly inherit the frame order in the original videos.

%
\noindent\textbf{Transcript-based Semi-Automatic Video Editing.}
Although traditional and recent approaches have achieved significant speedups 
to video editing experiences, they still have constraints that limit their scalability.
From one frontier, earlier works required either editing templates~\cite{vssr, ahanger1998automatic} 
or complex film and movie scripts~\cite{shen2009s,dialog_montage} as input to the sequencing system.
From the other angle, existing works in computational video editing have addressed 
the task of shot assembly but from an interactive (human-in-the-loop) perspective. 
For instance, ~\cite{quickcut} relies on the editors’ expertise to arrange the final shots, 
and ~\cite{write_a_video} requires users to specify keywords manually. 
This work addresses these limitations by proposing new downstream tasks for automatic clip sequence retrieval, generation, and completion.

%
\noindent{\textbf{Vision Language Representation Learning.}}
Visual Semantic Embeddings (VSE)~\cite{vse,vsepp} are widely adopted approaches 
for cross-modality tasks~\cite{movieqa,tvqa,set_complete,venugopalan2014translating,anet_cap,findit}
to learn joint visual-text representations.
With the emerging of self-supervised learning, a series of methods~\cite{cbt,hero,coot,milnce} 
have been proposed to train models on unlabeled datasets~\cite{howto100m}. 
These methods draw inspiration from prior fully supervised approaches~\cite{liu2019use,mithun2018learning,jsfusion,mme,mmt} 
that learn from manually annotated video description datasets~\cite{msrvtt,msvd,lsmdc,tvr,anet_cap,didemo,youcook2}.
However, unlabeled datasets rely on automated transcripts (and narrations) rather than ground-truth descriptions. 
This creates a semantic misalignment between the text and the visual information. 
%
To tackle this problem, \cite{milnce} introduced the MIL-NCE loss, 
a way to cope with temporal misalignment using Noise-Contrastive Estimation~\cite{nce} 
and Multiple Instance Learning~\cite{mil}. 
Our WeTravel videos also exhibit misalignment issues. 
We thus leverage the MIL-NCE loss to deal with it
but re-imagine the model architecture to: 
(a) model multi-shot clips; 
(b) dynamically and sequentially capture multiple concepts within one text query.

%
\noindent{\textbf{Pretext Tasks for Video Representation.}}
To learn video representations, without labels, numerous methods 
have been proposed using various pretext tasks,
including: colorization~\cite{vondrick2018tracking}, 
spatial-temporal contrastive learning~\cite{yang2020video,qian2020spatiotemporal}, 
future prediction~\cite{vr_anticipating,cpc,dpc,mpc},
temporal cycle consistency~\cite{dwibedi2019temporal}, 
motion prediction~\cite{wang2019self,zhan2019self}, 
frame-level temporal sorting~\cite{fernando2017self,lee2017unsupervised,misra2016shuffle,xu2019self}, 
and recognizing temporal transforms~\cite{video_temp_trans}.
Most of these methods focus on down-stream tasks such as action recognition~\cite{anet,knet,tsn,i3d,c3d,s3d}, 
constraining them to learn representations at the frame or short clip level, 
without considering shot-level temporal coherence.
Unlike previous works, we propose the pretext task of classifying artificial distortions on edited videos to learn sequencing styles from professionally edited videos.

\begin{figure*}[!ht]
\vspace{-14pt}
\centering
	\subfloat[\small Parallel Encoder (PE)\label{fig:mil_nce_parallel}]{
		\includegraphics[width=0.45\linewidth]{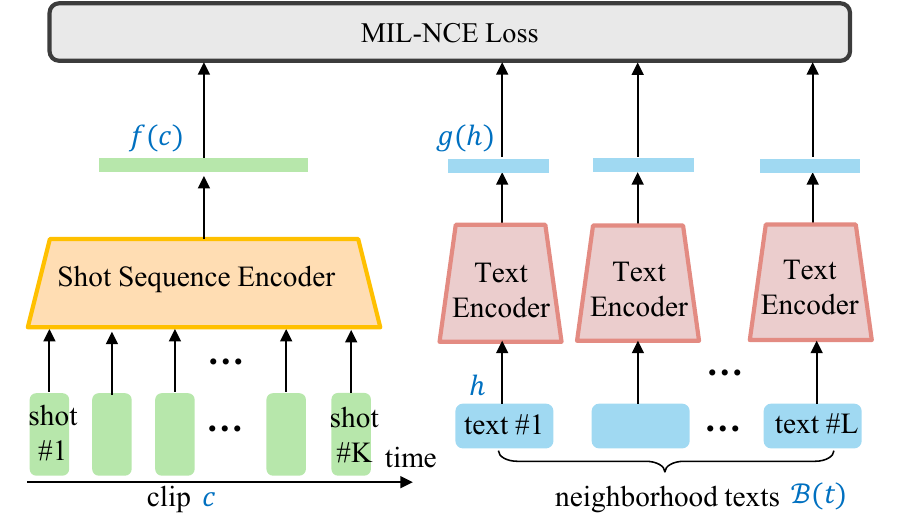}
		\vspace{-5pt}} \hfill
	\subfloat[\small Adaptive Encoder (AE) \label{fig:mil_nce_interactive}]{
		\includegraphics[width=0.42\linewidth]{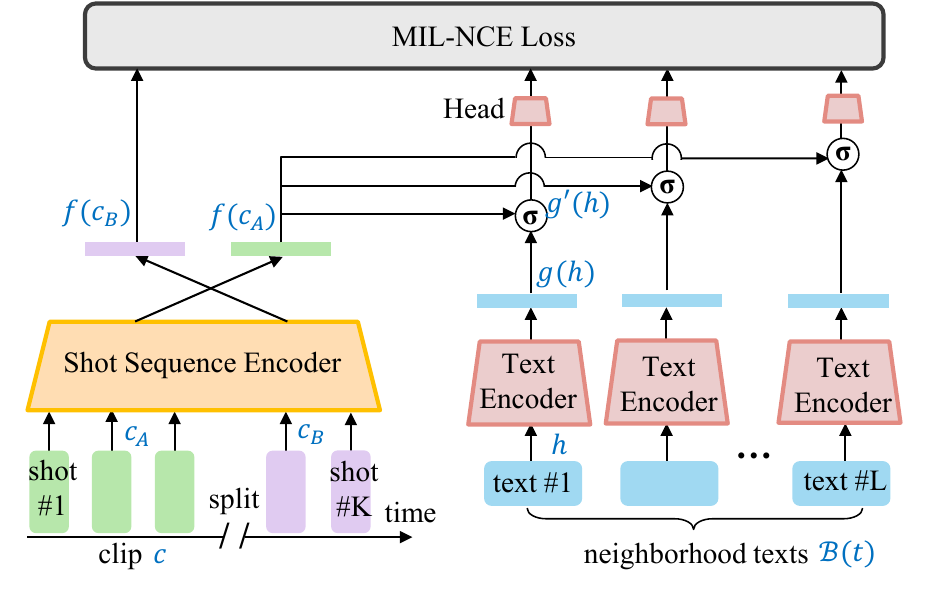}
		\vspace{-5pt}}
	\vspace{-10pt}
	\caption{\small
    	\textbf{Content Retrieval Module variants.} 
    	The Parallel Encoder (PE) and Adaptive Encoder (AE) learn task-specific text-video embedding. 
    	Both encoders receive as input a clip $c$ and neighborhood texts $\cB(t)$. 
    	\textbf{(a)} PE aims at capturing a hierarchical frame-shot-clip visual representation.
    	It encodes the clip $c$ and each text $h \in \cB(t)$ into clip feature $f(c)$ and text feature $g(h)$ respectively. 
    	\textbf{(b)} AE first splits the clip and encodes sub-clips as $f(c_A)$ and $f(c_B)$.
    	Then a Context Interaction Function $\sigma(\cdot)$ updates each encoded text $g(h)$ so that $g'(h)=\sigma(f(c_A),g(h))$;
    	it aims to mask out already retrieved visual content from the text at the feature level.
    	Furthermore, MLP heads are introduced for further encoding text features. 
    	Both encoders are trained with MIL-NCE loss~\cite{milnce} to tackle visual-text misalignment.
	}
	\label{fig:mil_nce}
	\vspace{-10pt}
\end{figure*}

%% file: articles/benchmark.tex

\vspace{-5px}
\section{WeTravel Dataset}
\label{sec:benchmark}
\vspace{-3px}
%
Our goal is to find a large number of narrated videos with devoted editing efforts.
Since similar video retrieval tasks have been introduced, 
there are many video datasets that could be considered~\cite{knet,anet,howto100m,gu2018ava}; 
yet, none of them fit our needs to learn and benchmark text-based clip sequencing.
Towards this goal, we collect WeTravel,
a dataset containing $31.2$K narrated travel videos and over $3.4$M shots. 
To create our dataset, we leverage publicly available travel videos on YouTube, 
an accurate shot boundary detector, 
and the widespread availability of YouTube video transcripts obtained with Automatic Speech Recognition. 
In this section, we describe WeTravel at a glance. 
More details can be found in the \textit{supplementary}.
We will release the dataset with the URLs to the videos.

%
\noindent\textbf{Why travel videos?}
According to our goal, the closest fit would be the HowTo100M ~\cite{howto100m} dataset, 
which contains instructional videos paired with transcripts. 
However, instructional videos have two limitations. 
First, being instructional, these videos contain a vast number of talking-to-camera shots. 
Second, given that they depict the actions required to complete a goal, 
the shots from instructional videos are very long. Travel videos, 
instead, tend to be carefully edited to engage audiences, 
which results in a large ratio of creatively sequenced clips per video. 
Moreover, travel videos are usually narrated, 
either by a first-person or a voice-over added during editing. 
Besides, some travel videos are carefully edited with relatively aligned transcripts,
allowing us to benchmark clip sequencing tasks. 

%
\noindent\textbf{Source.}
To gather the data, we query YouTube using the following template: 
``[travel / travel guide] + [country / continent]''. 
These queries typically obtain between $50$ to $600$ results; 
we remove duplicated videos across queries by tracking their 
YouTube ID.
By further manually checking a subset, 
we found that only less than 0.4\% of videos are duplicated. 
Similar to \cite{howto100m}, we download
and parse the transcript for each video.

%
\noindent\textbf{Pre-processing.}
The \textit{shot} is the basic unit for multi-shot clip creation. 
Hence, we first apply a shot-boundary detector across all retrieved videos; 
we use ~\cite{pyscenedet} for that. 
Even if travel videos contain a large number of creatively sequenced clips, 
there might be sections within a video that might contain talking-head shots. 
Thus, we devise a heuristic to find and filter these types of clips. 
We apply a face detector on the middle-frame of each shot 
and remove those with faces larger than $30\%$ of the video height. 
We also remove videos with long average shot duration 
to discard typical failure cases of the shot detector 
or irrelevant (single-scene / rarely-edited) videos. 
Furthermore, and concerning the transcript, 
we merge words with a temporal gap of less than one second 
into a single text chunk because Automatic Speech Recognition does not produce punctuation.
Fig.~\ref{fig:benchmark_sample} shows a typical multi-shot video-text pairs from WeTravel. 

%% file: articles/method.tex
\vspace{-8px}
\section{Methodology}
\label{sec:method}
\vspace{-4px}

%
In this section, we present our framework for text-based automated clip sequencing.
As shown in Fig.~\ref{fig:teaser}, given a gallery with $N$ shots, 
we create a multi-shot clip of $M$ shots according to an input query text $T$.
Our goal is to retrieve relevant shots to the query and compose valid shot sequencing styles (\eg, avoid extreme shot-scale changes between shots).
Hence, we here describe our training framework, which comprises a \emph{Content Retrieval Module} (CRM) 
to learn vision-language representation (Sec.~\ref{subsec:crm}), 
and a \emph{Temporal Coherence Module} (TCM) to model shot-level sequencing styles (Sec.~\ref{subsec:tcm}).
We also describe our beam search approach (Sec.~\ref{subsec:inference}), which combines 
the proposed modules at inference while assuring efficiency.

%
\subsection{Vision Language Representation Learning}
\label{subsec:crm}
\vspace{-2px}
A key requirement in text-based multi-shot clip creation is to 
retrieve shots with relevant content to the queries.
To achieve this goal, we train joint vision-language embeddings from unlabeled videos.
First of all, the model's design should consider the task itself, 
which requires a representation that covers a frame-shot-clip hierarchy. 
To do so, we propose a Parallel Encoder that separately encodes clip and text features 
for the shot-based narrated clips.
Furthermore, unlike traditional video retrieval that searches for the video-text matches in a single step,
our task requires adaptive retrieval of shots in an online fashion to compose shot sequences.
Given these requirements a natural question arises, 
how can we create shot sequences that fully cover all the visual concepts in the query text?
Our Adaptive Encoder offers a solution for that question; it dynamically transforms text queries (features) leveraging 
the visual information of already retrieved shots.
An additional challenge is 
the natural misalignment between text and video, 
as illustrated earlier in Fig.~\ref{fig:benchmark_sample}. 
We can observe that concepts like ``mountains'' in the $4$th text is 
found in the $3$rd clip, instead of its counterpart clip. 
To cope with text-visual misalignment, we adopt the MIL-NCE loss \cite{milnce}.

%
\noindent\textbf{Parallel Encoder.}
Originally, a clip-text pair $(c, t)$ serves as positive pair where
$t$ is a transcript text covered by a video clip $c =  \{s_i\}_{i=1}^{K}$ with $K$ complete shots.
The core of Multiple Instance Learning is to expand the single pair
to a set of candidate pairs within a positive bag, denoted as $\cP = \{(c, h) | h \in \cB(t)\}$, where $\cB(t)$
is the neighbor text set of $t$ with $L$ texts. 
This formulation dovetails with our observation that a video clip's visual concepts may lie within 
its nearby texts. 
As in Fig.~\ref{fig:mil_nce_parallel}, for each clip, the \emph{Shot Sequence Encoder} generates clip-level feature $f(c)$ by 
(i) encoding each shot with a shared weight S3D~\cite{s3d} hence 
$\phi(c)=\{\phi(s_i)\}_{i=1}^{K}$;
(ii) assembling the shot-level feature by a Consensus Aggregation function 
$\cG(\cdot)$ to produce clip-level feature $f(c)=\cG(\phi(c))$.
As for each candidate text $h$ in the positive bag, 
a text encoder is used to obtain the feature $g(h)$.
We follow the same text encoder structure in ~\cite{milnce} to compute $g(h)$.
Finally, the loss function is given as:
\vspace{-10px}
\begin{equation}
\small
\hspace{-0.3cm} \cL = \frac{1}{n}\sum_{i=1}^{n} \log \frac{\sum\limits_{(c, h) \in \cP_i} e^{f(c)^\top g(h)} }{\sum\limits_{(c, h) \in \cP_i} e^{f(c)^\top g(h)} + \sum\limits_{(c', h') \sim \cN_i} e^{f(c')^\top g(h')}}\hspace{-0.2cm}
\vspace{-3px}
\end{equation}
where $n$ is the batch size; $\cN_i$ denotes the set for sampling negative pairs.
Specifically, the negative pairs are formed by selecting texts and clips from different positive bags.

%
\noindent\textbf{Adaptive Encoder.}
It's often the case that one query text conveys multiple visual concepts.
For example, to generate a multi-shot clip for the sentence 
``\emph{Take a walk through dense forests to the beach}'',
one may first display some shots of dense forests 
and then those containing a beach with a forest nearby.
That means, given the retrieved shots, the current query text feature 
should focus on completing the context that is missing in the already retrieved shots.
Following this idea, we present the Adaptive Encoder with a \emph{Context Interaction Function} 
that dynamically updates the text embedding according to existing shots. 
As shown in Fig.~\ref{fig:mil_nce_interactive}, this function $\sigma(\cdot)$ computes:
\vspace{-4px}
\begin{equation} \hspace{-0.3cm}
\sigma\big(f(c_A), g(h)\big)=g(h) - m\big(f(c_A), g(h)\big)\cdot \mW^\top f(c_A) \hspace{-0.2cm}
\vspace{-4px}
\end{equation}
Specifically, in Adaptive Encoder, we randomly split a video clip into two parts
with the first part $c_A$ captures the context in retrieved shots
while the second part $c_B$ as the missing one.
Here $f(c_A)$ is the feature for the first part, 
which is sent to $\sigma(\cdot)$ for adjusting the text feature.
$g(h)$ denotes the old text feature, and $\mW$ is the projection matrix
that implemented by a fully-connected layer to project visual features into the text space. 
We define $m(a,b)=\max(0, \text{cos}(a,b))$ to measure the similarity between two feature vectors.
Intuitively, $m(\cdot,\cdot)$ measures the retrieval coherence so as to control the degree of text feature update.
Finally, the updated text feature $g'(h)=\sigma\big(f(c_A), g(h)\big)$ 
is sent to an MLP head for further encoding, and then, 
with the feature of the second part of the clip $f(c_B)$ together, 
are used to calculate the loss.

%
 \begin{figure}[t]
	\centering
	\includegraphics[width=0.95\linewidth]{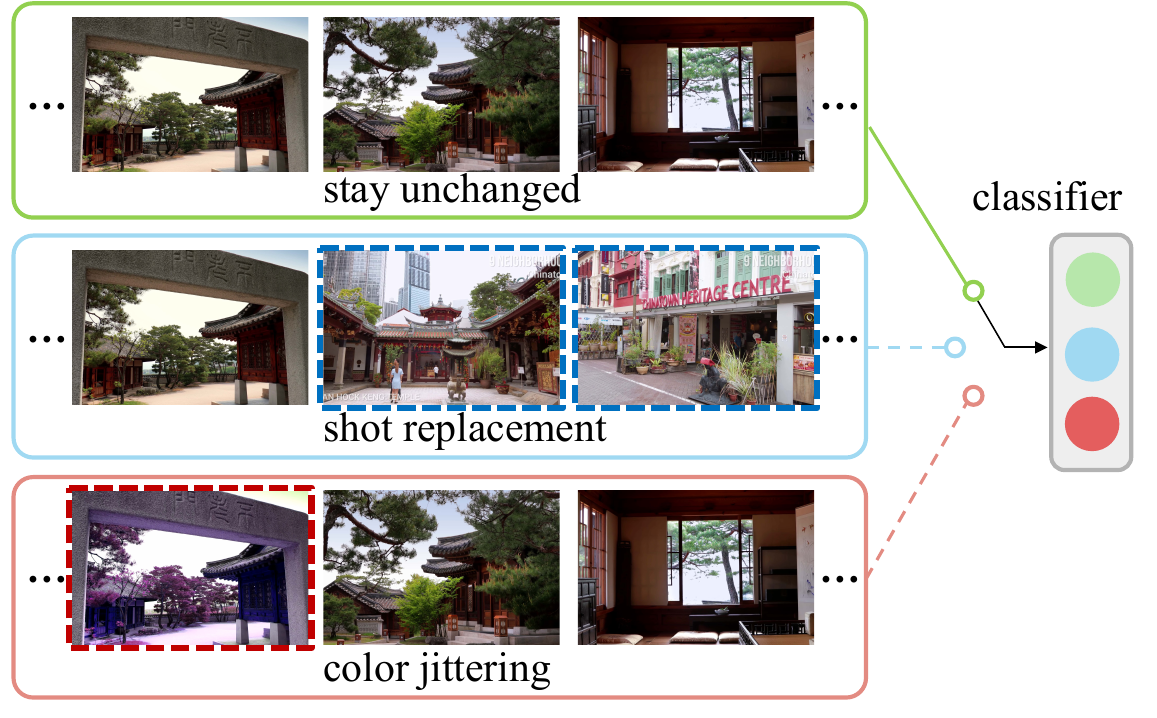}
	\vspace{-13pt}
	\caption{\small
    	\textbf{Learning shot sequencing styles.} 
    	The shot ordering and composition matters, 
    	so we learn to assemble clips by discriminating real shot sequences against altered ones
    	via a distortion classification pretext task. 
    	The task consists of identifying the applied transformation to an input video. 
    	We propose three possible distortions: 
    	(a) stay unchanged; 
    	(b) shot replacement, which randomly changes shots in the original video; 
    	and (c) apply color jittering to multiple random shots in the original multi-shot clip. 
    }
	\label{fig:dist_cls}
	\vspace{-15pt}
\end{figure}

%
\vspace{-3px}
\subsection{Learning Shot Sequencing Styles}
\vspace{-4px}
\label{subsec:tcm}
%
Apart from visual-semantic relevance, a well-composed multi-shot clip 
should also exhibit coherent temporal structures,
\ie smooth shot transitions in color, shot scale and angle, camera movement, among others ~\cite{christianson1996declarative,cutting2011act,editgrammar}.
However, as technology improves and creative preferences change, 
editing styles will evolve and new editing rules will emerge~\cite{gfl_grammer_film_lang,holway2013steadicam,editgrammar}.
This evolving paradigm makes it a challenge to model sequencing styles. 
Our key idea is to learn sequencing styles in a data-driven fashion -- 
we do so by synthesizing bad (or fake) sequencing styles, and learning to classify a set of distortions applied to compose the fake sequence.
In practice, we leverage the multi-shot clips in our WeTravel videos to learn this task.

%
Towards learning valid sequencing styles, we introduce a \emph{Temporal Coherence Module} (TCM) 
shown in Fig.~\ref{fig:dist_cls}, which first applies 
and then classifies the distortion on each video clip.
At first, given a video clip with $K$($\geq$2) shots, 
we sample a distortion operation $d$ from 
$D=\{\text{unchanged}, \text{shot replacement}, \text{color jittering}\}$ according to a probability $p(D)$.
Afterward, the distortion classifier receives the temporal transformed clip 
as input to predict which distortion was applied on it. 
Specifically, the distortions are:
(1) stay unchanged: not apply any distortion and output the original clip;
(2) shot replacement: randomly replace $k \in [1,K-1]$ shots from other videos.
(3) color jittering: apply channel shuffle and channel multiplying on $k \in [1,K-1]$ shots. 
For detailed model implementation, please refer to the supplementary.

%
\begin{figure*}[t]
\vspace{-5pt}
	\centering
	\includegraphics[width=0.9\linewidth]{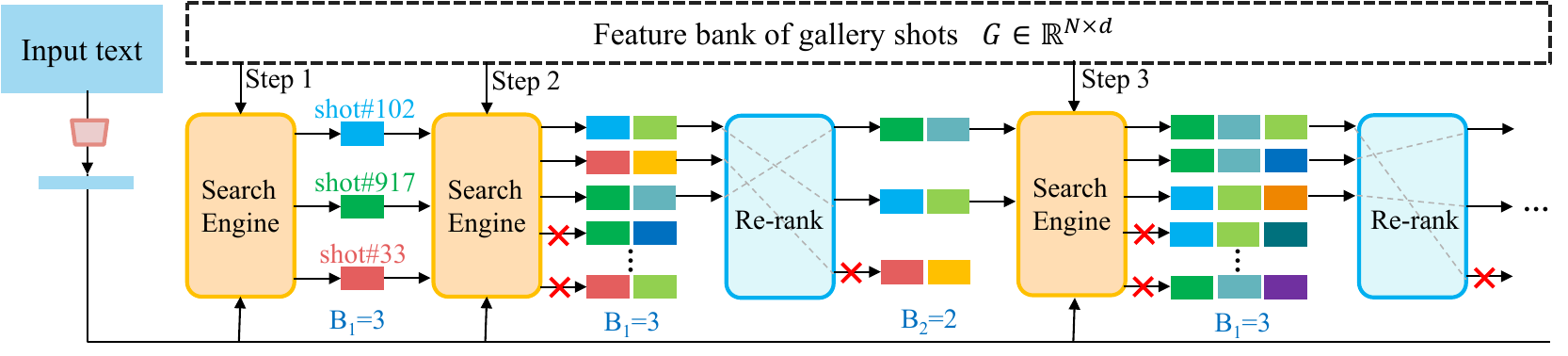}
	\vspace{-10pt}
	\caption{\small
        \textbf{Beam search pipeline.}
    	Here we illustrate the clip sequencing process. 
    	At Step 1, the system retrieves $B_1$=3 candidate shots from the gallery $G$.
    	At Step 2, once the search engine receives 3 incomplete shot sequences and the query feature,
    	it generates 3 updated query features. 
    	kNN uses the query features to retrieve multiple candidate shots for each of the query,
    	put candidates together, and sort out top $B_1$=3 candidate sequences. 
    	The Re-rank module further sorts the sequences by temporal coherence scores 
    	and only keep the top $B_2$ results.
    	The whole procedure in Step 2 is repeated in future steps.
	}
	\label{fig:inference}
	\vspace{-10pt}
\end{figure*}

%
\vspace{-3px}
\subsection{Efficient Inference Pipeline}
\vspace{-4px}
\label{subsec:inference}
Unlike video retrieval that searches on a gallery set only once, 
automated clip sequencing follows an iterative process 
that continuously sends conditioned queries to the search engine to retrieve one shot at a time.
Suppose we have a gallery set of shot features such that 
$G=\{s_i\}_{i=1}^N$ and $s_i \in \mathbb{R}^d$, 
the time complexity for assembling $M$ shots by the Brute-force search method is $\cO(N^Md)$.
This is because the number of clips generated by permutation and combination is $N^M$ 
if we follow the standard video retrieval strategy.
Brute-force then becomes too computationally expensive to be practical.
%
An alternative way is to use a greedy search that retrieves top-M shots 
and arranges them according to their similarity scores.
Unfortunately, greedy search often leads to suboptimal results~\cite{showtell}. 

%
To efficiently create multi-shot clips while making full use of the above pre-trained representations, 
we introduce our beam search based pipeline as illustrated in Fig.~\ref{fig:inference}.
Given a previously encoded text $g^{t-1}(h)$ as query feature at time step $t$,
as well as $B_2$ incomplete shot sequence from the previous stage,
the workflow of the inference pipeline at time step $t$($>$2) is described as follows:
(1) It first generates $B_2$ new query features $\{g_i^t(h)\}_{i=1}^{B_2}$ for all incomplete shot sequences
using the models mentioned in Sec.~\ref{subsec:crm}.
(2) kNN algorithm retrieves $B_1$ ($B_1 > B_2$) top candidate shots from the gallery. 
The retrieved shots are then appended to their corresponding incomplete shot sequences.
Like traditional Beam search, each query in a beam retrieves multiple results. We put them together, sort them 
according to similarity score, and take the top-$B_1$ as the top candidate sequences.
(3) Re-rank all the candidate sequences by temporal coherence scores obtained 
from TCM mentioned in Sec.~\ref{subsec:tcm} and keep the top $B_2$ as the  next stage's input.
Please refer to the supplementary video for animated explanation.

%
Suppose the inference time for generating a new text feature is 
$\tau_c(B_2)$ and that for distortion classification is $\tau_t(B_1)$, 
then the overall time complexity of the inference pipeline is $\cO(M\cdot(\tau_c(B_2)+\tau_t(B_1)+Nd))$.
The model inference time $\tau_c(B_2)$ and $\tau_t(B_1)$ can be bounded 
to an acceptable time by adjusting the beam size parameters $(B_1, B_2)$.
Theoretically, our pipeline is more efficient than Brute-force search. 
We empirically demonstrate the run-time in Sec.~\ref{subsec:exp-ablation}.

%% file: articles/experiment.tex

\vspace{-5px}
\section{Experiments}
\label{sec:experiment}
\begin{figure}[t]
    \vspace{-3pt}
	\centering
	\includegraphics[width=0.95\linewidth]{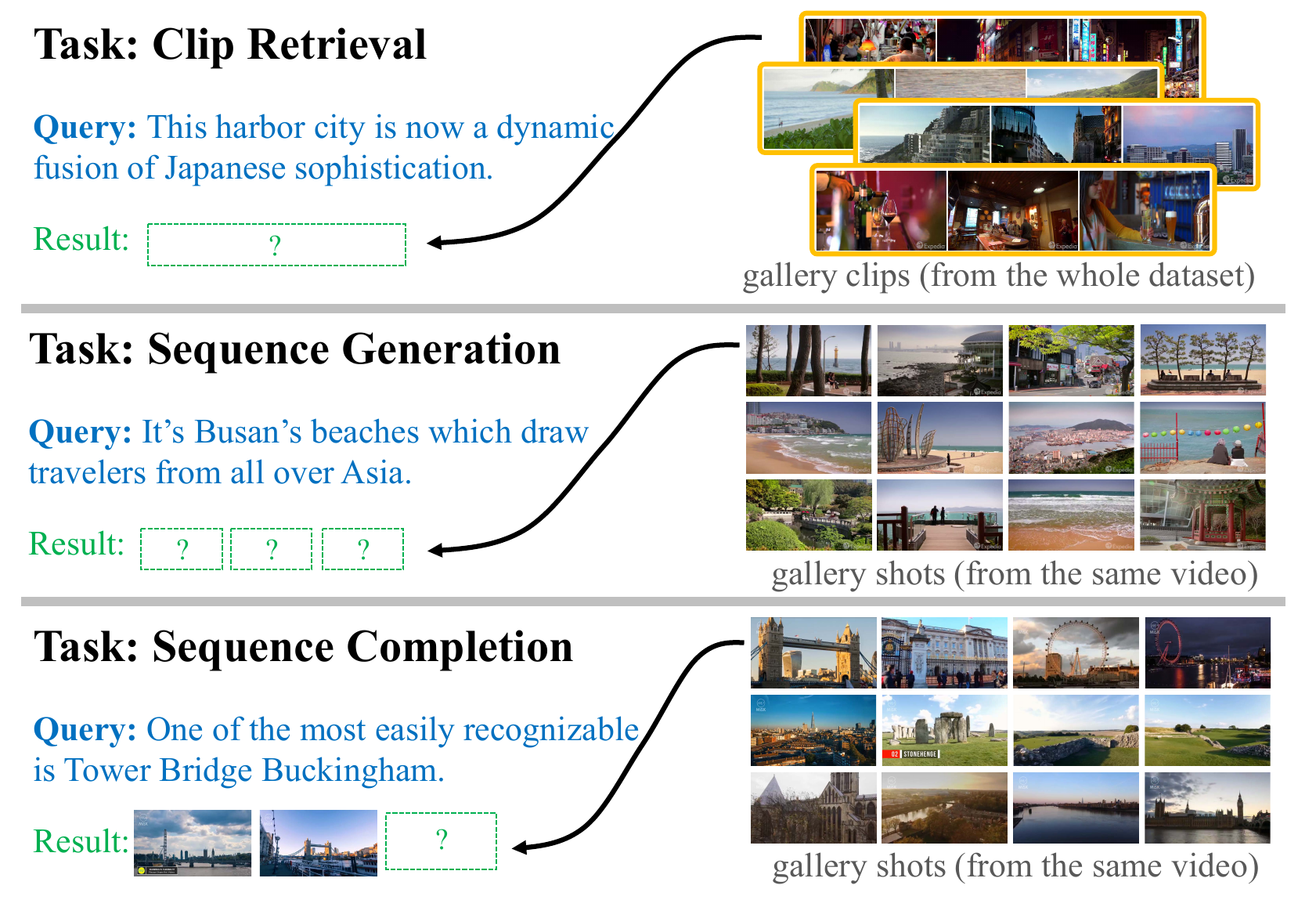}
	\vspace{-13pt}
	\caption{\small
    	\textbf{Illustration of the downstream tasks:} 
    	Clip Retrieval; Sequence Generation; Sequence Completion.
    }
	\label{fig:task_definition}
	\vspace{-15pt}
\end{figure}

\vspace{-4px}
%
In this section, we conduct experiments on several downstream tasks (see Fig.~\ref{fig:task_definition})
designed for text-based clip sequencing quantitative studies.
Since evaluating the quality of generated multi-shot clips might be subjective,
we further conduct user studies involving both inexperienced users 
and professional video editors as qualitative analysis.
Please refer to \emph{supplementary} for visualization and details.

%
\subsection{Experiment Setup}
\label{subsec:exp-setup}
\vspace{-4px}

%
\noindent\textbf{Dataset.}
We randomly split the WeTravel dataset into \emph{train, val, test} 
subsets using $6$:$1$:$3$ ratios. 
Note that for tasks like video retrieval, 
well-written captions and relatively aligned clip-text pairs are needed. 
Among all the videos in the testing set, 
we found that those from the YouTube channel \emph{Expedia} closely follow our requirements. 
Accordingly, we create an additional test set called WT-expedia, 
different from the original test set called WT-test, 
where all the videos are from the Expedia channel. 
Finally, there are $130$ videos with $4.3$K captions and $13$K shots in WT-expedia.

%
\noindent\textbf{Feature extraction.}
It is not a surprise that video representation learning consumes massive computation resources. 
Original MIL-NCE models~\cite{milnce} are trained on $64$ Cloud TPU v3 with $128$ videos per TPU, 
$32$ frames per video for about 3 days. 
That is beyond the computational budget of this paper's authors.
We instead conduct our experiments using pre-trained features. 
For each shot with $16$ sampled frames, 
we extracted the feature from the \emph{mixed5c} layer of S3D~\cite{s3d} 
pre-trained on HowTo100M~\cite{howto100m}. 
To encode text data, we use a Google-News pre-trained Word2Vec~\cite{w2v}.

%
\noindent\textbf{Training vision-language models.}
We use a two-layer MLP for encoding shot features, 
and AvgPooling over the time-step dimension as the Consensus Aggregation Function. 
Given a positive pair of video clip and text,
\emph{at most} $3$ closest neighbor texts are sampled as positive candidates, 
\ie, any overlapping sentence within a 3-second interval from the query text.
The number of shots in each video clip is restricted to $10$.
The max length of a text is $16$ words. 
More details in the \textit{supplementary material} include data processing details, model structures, and training schemes.

%
\begin{table}[!t]
\vspace{-10pt}
\small
\begin{tabular}{c|c|cccc}
\hline
Method    & Data & R@1           & R@10          & R@50           & MedR         \\ \hline
Random    & -    & 0.02          & 0.23          & 1.17           & 2144         \\
MIL-NCE~\cite{milnce} & HTM & 0.20  & 1.52  & 4.76           & 1266         \\
Parallel & HTM  & 0.23          & 1.89          & 7.11           & 943          \\
VSE~\cite{vse}       & +WT  & 0.09          & 0.89          & 4.34           & 1149         \\
VSE++~\cite{vsepp}     & +WT  & 0.02          & 0.35          & 1.52           & 1935         \\
Parallel & +WT  & \textbf{0.63} & 3.50          & 12.20          & \textbf{548} \\
Adaptive & +WT  & 0.58          & \textbf{3.59} & \textbf{12.22} & 570          \\ \hline
\end{tabular}
\vspace{-10pt}
\caption{\small
    \textbf{Clip retrieval results.} 
    The query is a transcript sentence; the candidates are multi-shot clips.
    Overall, training with WeTravel (WT) yields better performance than doing so with HowTo100M (HTM). 
    We also notice our models significantly improve upon the frame-based model, VSE, and VSE++ baselines.
}
\label{tab:exp-vid-rtv}
\vspace{-13pt}
\end{table}

%
\noindent\textbf{Training the temporal coherence module.}
We also use feature-level representations to train the temporal coherence module via the pretext distortion classification task.
For each shot, S3D features and RGB histograms are extracted.
The shot replacement randomly replaces shot features from other videos,
while color jittering is implemented by histogram-level channel shuffle
and histogram stretching.
We randomly crop a sequence with various shots from a video 
and apply distortions with equal probability on at most half of the shots.
Before passing it to the model, we first multiply each feature 
by the similarity between itself and its neighbor features. 
We do this to add temporal correlation priors to the model, 
which helps stabilize the training process.
Later on, two separated two-layer LSTMs encode S3D features and histogram features, respectively.
The outputs of each LSTM are fed through a two-layer MLP for distortion prediction.
We demonstrate empirically that our TCM is able to successfully learn to: (i) classify the distortion type, and (ii) rank real vs. fake sequencing styles. Due to space limitations, please refer to the \textit{supplementary material} for in depth analysis of these results.

%
\begin{table*}[!t]
\vspace{-8pt}
    \centering
    \small
\begin{tabular}{ccccc||ccccc||c}
\hline
\multicolumn{5}{c||}{Configurations}                                                                                                                                                               & \multicolumn{5}{c||}{sequence generation}           & \multicolumn{1}{|c}{seq. comp.} \\ \hline
\multicolumn{1}{c|}{Loss}                   & \multicolumn{1}{c|}{Arch.}                         & \multicolumn{1}{c|}{Data}                              & Search & Distortion   & Recall & AOP-1 & AOP-2 & AOP-3 & AOP-S & Acc.      \\ \hline
\multicolumn{1}{c|}{Random}                   & \multicolumn{1}{c|}{-}                            & \multicolumn{1}{c|}{-}                                 & -      & -            & 2.05   & 1.92  & 0     & 0     & 1.92  & 0.74      \\ \hline
\multicolumn{1}{c|}{MIL-NCE~\cite{milnce}}                  & \multicolumn{1}{c|}{Parallel}                     & \multicolumn{1}{c|}{HTM} & greedy & -            & 5.89   & 5.63  & 0.28  & 0     & 5.90  & 2.20      \\ \hline
\multicolumn{1}{c|}{VSE~\cite{vse}}                      & \multicolumn{1}{c|}{\multirow{3}{*}{Parallel}}    & \multicolumn{1}{c|}{\multirow{3}{*}{+WT}}              & greedy & -            & 3.94   & 3.76  & 0.11  & 0     & 3.87  & 1.25      \\
\multicolumn{1}{c|}{VSE++~\cite{vsepp}}                    & \multicolumn{1}{c|}{}                             & \multicolumn{1}{c|}{}                                  & greedy & -            & 2.78   & 2.69  & 0.03  & 0     & 2.72  & 0.95      \\
\multicolumn{1}{c|}{MIL-NCE~\cite{milnce} }                 & \multicolumn{1}{c|}{}                             & \multicolumn{1}{c|}{}                                  & greedy & -            & 7.28   & 6.89  & 0.47  & 0.08  & 7.43  & 2.63      \\ \hline
\multicolumn{1}{c|}{\multirow{4}{*}{MIL-NCE~\cite{milnce}}} & \multicolumn{1}{c|}{\multirow{2}{*}{Parallel}}    & \multicolumn{1}{c|}{\multirow{2}{*}{+WT}}              & beam   & repl.        & 7.33   & 6.96  & 0.63  & \textbf{0.11}  & 7.70  & 4.01      \\
\multicolumn{1}{c|}{}                         & \multicolumn{1}{c|}{}                             & \multicolumn{1}{c|}{}                                  & beam   & repl. + jit. & 7.39   & 7.06  & 0.62  & 0.10  & \textbf{7.77}  & 4.14      \\ \cline{2-11} 
\multicolumn{1}{c|}{}                         & \multicolumn{1}{c|}{\multirow{2}{*}{Adaptive}} & \multicolumn{1}{c|}{\multirow{2}{*}{+WT}}              & beam   & repl.        & 7.37   & 7.04  & \textbf{0.63}  & 0.05  & 7.72  & 4.24      \\
\multicolumn{1}{c|}{}                         & \multicolumn{1}{c|}{}                             & \multicolumn{1}{c|}{}                                  & beam   & repl. + jit. & \textbf{7.42}   & \textbf{7.09}  & 0.57  & 0.05  & 7.71  & \textbf{4.42}      \\ \hline
\end{tabular}
\vspace{-10pt}
    \caption{\small
        \textbf{Sequence generation and completion results.} 
        HTM denotes HowTo100M~\cite{howto100m}
        and +WT means finetuning on WeTravel; 
        repl. denotes shot replacement while jit. denotes color jittering.
        Findings:
        (a) MIL-NCE is the best loss; 
        (b) fine-tuning on WT is beneficial; 
        (c) Adaptive Encoder boosts performance for sequence completion;
        (d) beam search with TCM boost performance in both tasks.
    }
    \label{tab:exp-merge}
\vspace{-10pt}
\end{table*}

%
\noindent\textbf{Baseline models.}
For vision-language related tasks, we adopt VSE~\cite{vse} and VSE++~\cite{vsepp}
as baselines. We train them with a pairwise ranking loss
following the same settings as our Parallel/Adaptive encoders.
For clip retrieval, we also add the frame-based MIL-NCE model as a baseline; we follow its original paper \cite{milnce} details to extract text and video features.

%

%
\vspace{-4px}
\subsection{Quantitative Results}
\label{subsec:exp-quantitative}
\vspace{-4px}
We conduct experiments on several downstream tasks to show the capability of the proposed framework.
We also report the run-time performance of the inference pipeline.

%
\noindent\textbf{Clip retrieval.}
This task aims at searching a video clip from a candidate pool given a transcript text as the query
(see Fig.~\ref{fig:task_definition}).
We conduct experiments on clip retrieval using language queries on WT-expedia.
Specifically, the query is a transcript sentence; 
The GT is a single clip (with multiple shots) temporally-aligned with
such sentence. There are 4287 queries in WT-expedia.
We adopt top-K recall and median rank as evaluation metrics.
As shown in Tab.~\ref{tab:exp-vid-rtv}, we observe that:
(i) Models trained with MIL-NCE loss surpass VSE models trained with pairwise ranking loss.
(ii) Original frame-based encoder in ~\cite{milnce} processes 
a relatively short video clip as a frame stream. 
This structure is insufficient for shot sequence with varied lengths and frequent shot transitions.
(iii) Our encoders finetuned on WeTravel archive better results than 
those pre-trained only on HowTo100M~\cite{howto100m}.
These results show the importance of gathering WeTravel 
and its effect on improving clip retrieval performance. 
The performance of Parallel Encoder and Adaptive Encoder are comparable here since clip 
retrieval task does not require update on text features.

%
\noindent\textbf{Sequence generation.}
To study the effectiveness of combining the CRM and TCM modules, 
we design the task of sequence generation (see Fig.~\ref{fig:task_definition}). 
The sequence generation task is similar to our goal of text-based clip sequencing.
For each query text, we generate from scratch a shot sequence with length $M$ ($M$=$3$ here)
departing from a gallery that contains all the shots from \emph{the same video}.
In addition to recall, we also compute the \emph{Average $k$-th Order Precision (AOP-$k$)}
as an additional evaluation metric.
AOP measures how many sub-clips with $k$ shots in the generated video ($M$ shots, $M\geq k$) match
those in the original (ground truth) video.
The definition of AOP is given by:
\vspace{-4pt}
\begin{equation} \small
\text{AOP-}k = \frac{1}{N}\sum_{i=1}^{N} \frac{\sum_{s\in \cS_k^i} \mathds{1}(s \in \cU_k^i) }{|\cS_k^i|}\cdot \min(1,\frac{|\cS_k^i|}{|\cU_k^i|})
\vspace{-4pt}
\end{equation}
$\cS_k$ is the set of sub-clips with a length of $k$ for each generated clip
while $\cU_k$ is that for each ground-truth clip. $\mathds{1}(\cdot)$ is the indicator function.
We set the beam size parameters as ($B_1$=$4$, $B_2$=$2$).
Apart from previous observations that MIL-NCE loss and WeTravel fine-tuning matters,
the results shown in Tab.~\ref{tab:exp-merge} further indicate that
the beam search strategy is helpful for shot assembly
when comparing Parallel Encoder using greedy and beam search.

%
\noindent\textbf{Sequence completion.}
We also conduct experiments on sequence completion
given a text with an incomplete shot sequence, 
retrieving the missing shot within \emph{the same video} (see Fig.~\ref{fig:task_definition}).
Here we show the sequence completion result on completing the last shot
of a sequence. The beam size is set as $B_1$=$6$ in our experiments.
We can observe from Tab.~\ref{tab:exp-merge} (the last column) that 
(i) the accuracy improves a lot when applying beam search with CRM and TCM.
(ii) AE is better than PE at completing missing shots.

%
\vspace{-2pt}
\subsection{Ablation Studies}
\label{subsec:exp-ablation}
\vspace{-4pt}

%
\noindent\textbf{Inference time.}
To evaluate the efficiency of the proposed inference pipeline, 
we present the search engine's run-time performance using the full models.
We generate a sequence with $4$ shots, and the beam sizes are ($B_1$=$5$,$B_2$=$3$).
We use batch size $1$ to simulate real-world applications.
For GPU, we infer on single Titan X.
For CPU, we record the \emph{single thread} run-time performance on a 2.6GHz Intel Xeon CPU.
We show both inference time of GPU and CPU in Tab.~\ref{tab:infer-time}.
The results show that the video clips can be generated in real-time
even when the gallery is large,
and the run-time increases linearly concerning gallery size.

%
\begin{table}[t]
\small
       \centering
        \begin{tabular}{c|ccccc}
        \hline
        Gallery Size & 10K  & 100K & 500K & 1M    & 2M    \\ \hline
        GPU          & 16.4 & 25.3 & 73.5 & 133.3 & 243.9 \\
        CPU          & 46.3 & 158.0 & 671.1 & 1369.9 & 2777.8 \\ \hline
        \end{tabular}
        \vspace{-10px}
        \caption{\small
            \textbf{Inference time} in milliseconds \emph{\wrt} the gallery size. 
            We report single GPU and single thread CPU runtime.
            We can observe that the inference time scales linearly \wrt the gallery size.
        }
        \vspace{-8px}
        \label{tab:infer-time}
\end{table}

%
\noindent\textbf{Choice of beam sizes.}
We conduct experiments by changing beam sizes ($B_1$,$B_2$) for the Adaptive Encoder, 
and measure its influence on sequence generation/sequence completion results (Tab.~\ref{tab:exp-ablation}). 
The results show that tuning the beam sizes brings additional performance gains.

%
\noindent\textbf{Number of positive texts.}
We conduct ablations on how the max number of candidate texts, $L$, 
influences the performance of visual-language embeddings on sequence completion and clip retrieval (Tab.~\ref{tab:exp-ablation}).
We observe that multiple instance learning matters.
However, adding too many candidate texts introduces extra noise, leading to inferior results.

%
\begin{table}[t]
\small
    \begin{minipage}[]{0.23\linewidth}\centering
            \begin{tabular}{c|c}
            \hline
            $B_1$ & Acc. \\ \hline
            4  & 3.81 \\
            5  & 4.14 \\
            6  & \textbf{4.42} \\
            7  & 4.32 \\
            8  & 4.29 \\ \hline
            \end{tabular}
    \hfill
    \end{minipage}
    \begin{minipage}[]{0.34\linewidth}\centering
            \begin{tabular}{c|c}
            \hline
            ($B_1$,$B_2$) & AOP-S \\ \hline
            (3,2)  & 7.55 \\
            (4,2)  & \textbf{7.71} \\
            (5,3)  &  7.68 \\
            (6,3)  & 7.63 \\
            (7,4)  & 7.70 \\ \hline
            \end{tabular}
    \hfill
    \end{minipage}
    \begin{minipage}[]{0.4\linewidth}\centering
            \begin{tabular}{c|c|c}
            \hline
            \multirow{2}{*}{\shortstack{max\\L}} & \multicolumn{1}{c|}{\multirow{2}{*}{Acc.}} & \multicolumn{1}{c}{\multirow{2}{*}{R@1}} \\
                               & \multicolumn{1}{l|}{}                      & \multicolumn{1}{l}{}                     \\ \hline
            1                  & 3.99                                       & 0.53                                     \\
            3                  & \textbf{4.42}                                       & \textbf{0.58}                                     \\
            5                  & \textbf{4.42}                                       & 0.51                                   \\
            7                  & 4.39                                       & 0.49                                     \\ \hline
            \end{tabular}
    \hfill
    \end{minipage}
    \vspace{-10px}
    \caption{\small
        \textbf{Ablation studies}. 
        \textit{Left}: beam size $B_1$ on sequence completion; 
        \textit{Mid}: beam sizes ($B_1$, $B_2$) on sequence generation; 
        \textit{Right}: influence of max candidate text number in a 
        positive pair (max $L$) on sequence completion (Acc.) and clip retrieval (R@1).
    }
    \vspace{-12px}
    \label{tab:exp-ablation}
\end{table}

%
\vspace{-2pt}
\subsection{User Studies}
\label{subsec:exp-userstudy}
\vspace{-3pt}
To verify the quality of generated video sequences, 
we further conduct user studies with two audience groups: 
inexperienced people and video editing experts.
We collect $50$ query texts ($16$ user-generated and $34$ from the testing set)
to generate clips from WT-expedia gallery.
The texts cover diverse topics such as lifestyle, building, animal, and food.
A total of $200$ videos are generated for the $50$ cases with each case containing $4$ videos from $4$ models.
For all models, the sequence length is $4$ and ($B_1$=$5$,$B_2$=$3$) if applicable.
The users are asked to watch videos (in random order) to answer three 6-point
Likert-scale questions with choices ranging from \emph{strongly disagree} to \emph{strongly agree}:

%
\begin{figure}[t]
    \vspace{-3px}
	\centering
	\includegraphics[width=0.99\linewidth]{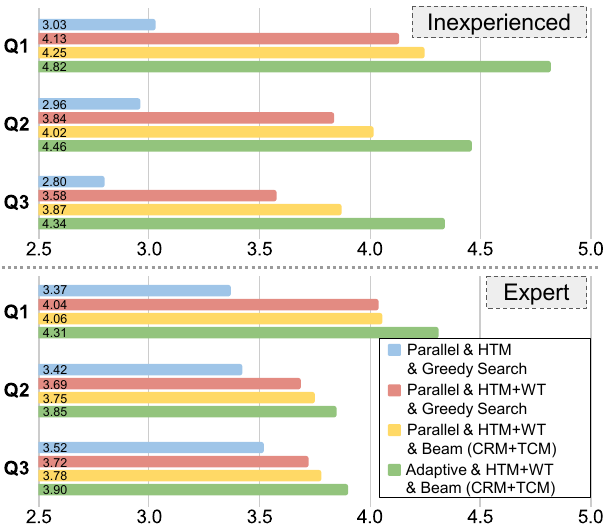}
	\vspace{-7px}
	\caption{\small
    	\textbf{User study results.} 
    	We conduct a user study with two sets of participants, 
    	inexperienced users and expert editors. 
    	We report the average score of 6-point ratings on: 
    	\textbf{(Q1)} asked the relevance of the content, 
    	\textbf{(Q2)} the temporal smoothness across shot transitions, and 
    	\textbf{(Q3)} the overall quality of the composed clips. 
	}
	\label{fig:user_study}
	\vspace{-10px}
\end{figure}

\noindent\textbf{Q1.} \emph{The video completely covers the info of the description. There's no much missing or totally irrelevant message.}

\noindent\textbf{Q2.} \emph{The transitions between shots are smooth and pleasant. (in terms of color, scale, movement, order, etc)}

\noindent\textbf{Q3.} \emph{The video appears professionally edited.}

To ensure the quality, we set up a consistency check by providing duplicated videos.
$18$ inexperienced users were invited and $16$ results are considered valid after consistency check. 
To make the task more comfortable, they are only required to score $10$ out of $50$ cases. 
For the expert group, we invite $6$ experts to watch and score all the cases.
The experts are all professional video editors with many years of full-time editing experience; some of them has impressive badges such as being an Emmy-nominated filmmaker.
From Fig.~\ref{fig:user_study}, we can observe that:
(i) experts are harsher with all the methods, 
(ii) models trained on HTM+WT score better than the baseline without fine-tuning on WT, 
(iii) beam search plus TCM provides slightly better quality than other models (Q2),
(iv) results from AE are better than PE.

%% file: articles/conclusion.tex
\section{Conclusion}
\label{sec:conclusion}
\vspace{-5pt}
In this paper, we introduced the first attempt for 
learning-based automated multi-shot clip creation for text.
For training, our framework integrates a \emph{Content Retrieval Module} 
to bridge the gap between video and text
and a \emph{Temporal Coherence Module} to model sequencing styles.
At inference, the search engine generates shot sequences in real-time.
To facilitate future research, we constructed WeTravel -- 
an unlabeled dataset of edited videos paired with transcripts.
Based on WeTravel, both quantitative analysis and user studies 
showed the effectiveness and efficiency of the framework.
Despite our contributions, many open challenges remain, e.g., 
how to model other essential editing elements such as music? 
or how to improve smoothness across shot transitions by modeling complex relations in the text? 
We believe further explorations include modeling joint vision-language-audio representations 
and improving shot-level temporal coherence using text priors.
We hope our work, and the WeTravel dataset, 
paves the way for further explorations on clip sequencing research.

%% file: articles/acknowledgement.tex
\section{Acknowledgement}
\label{sec:acknowledgement}
\vspace{-5pt}
This work is partially supported by GRF 14205719 and Centre for
Perceptual and Interactive Intelligence.

%% file: supp_articles/userstudy.tex
\section{User Study}
\label{supp:user-study}

\begin{figure*}[t]
	\centering
	\includegraphics[width=\linewidth]{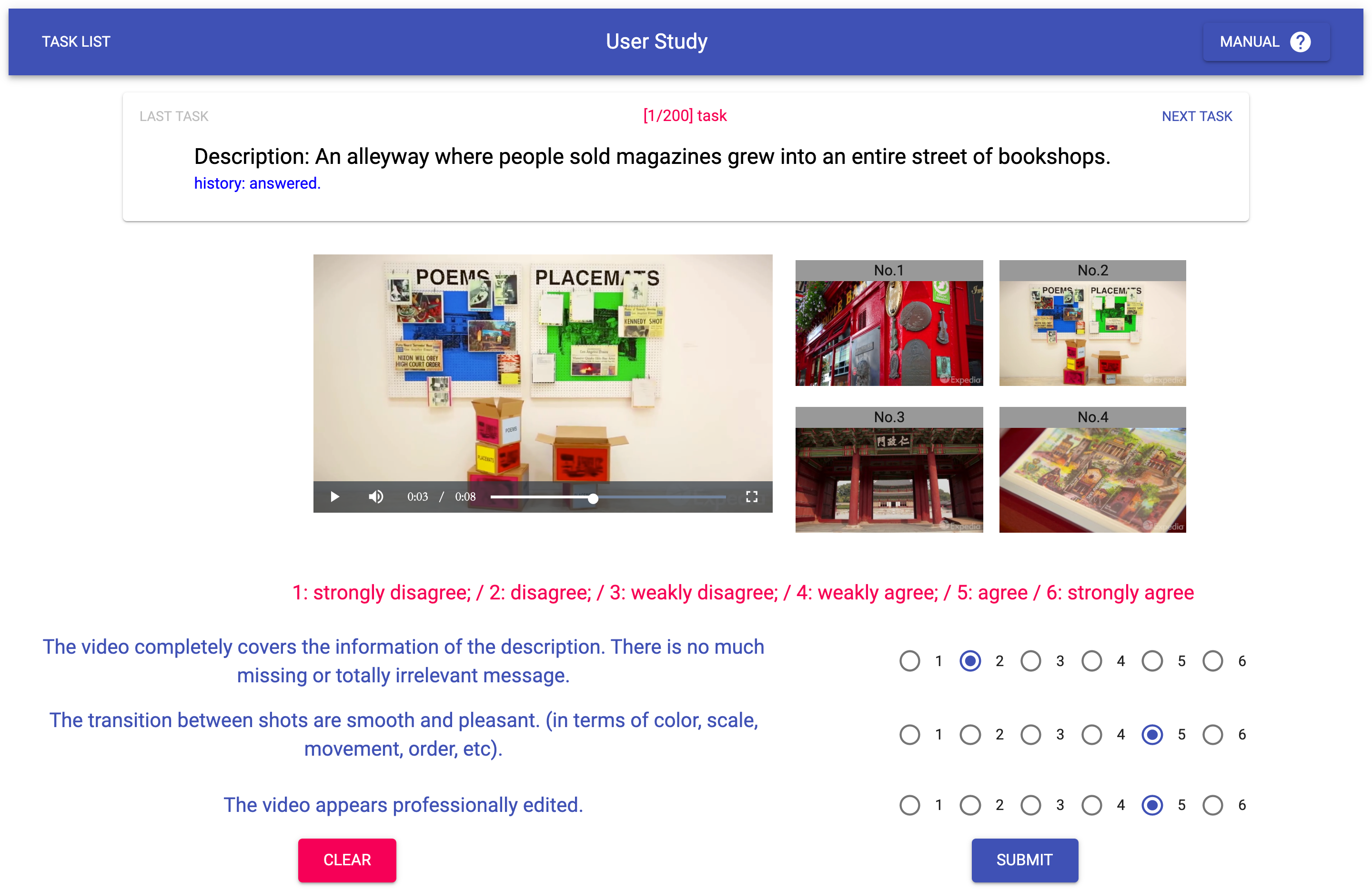}
	\caption{\small
		\textbf{User study interface}.
		From the top to the bottom are:
		(1) Information board that mainly showcases the current text query (the description).
		(2) Video player and shot keyframe board.
		(3) Question answering board.
	}
	\label{fig:supp_user_study_interface}
\end{figure*}

We conduct user studies involving both inexperienced users and experts.
The experts are all professional video editors with years of working experience.
Below are their profile information:
\begin{itemize}
    \item \textbf{Expert 1}: He is a full-time filmmaker since 2005. He started as an editor at a TV station but afterward decided to work as a freelance editor. He has experience editing full-length feature films, documentaries, and commercials. Currently, his editorial work focuses on YouTube videos. He produces, manages, and edits content for two YouTube channels, and he has his own YouTube channel.
    \item \textbf{Expert 2}: He is an Emmy-nominated filmmaker. He specializes in all things video production, whether it be any pre-production work, producing, cinematography, and post-production work. He is 1/4th of a production company.
    \item \textbf{Expert 3}: She is a video production artist with over a decade of experience. She specializes in taking clients from start to finish in creating marketing videos on various platforms. Genres include documentary-style introduction videos, social media campaigns, YouTube content and more.
    \item \textbf{Expert 4}: He is presently employed as an Manager Video Editor at a university. He has 6 years of experience working and editing digital files and he is proficient with Adobe Creative Suite and Davinci Studio. In 6 years in the industry, he has acquired valuable experience in video editing, color grading, providing HQ sound effects, and cinematic results.
    \item \textbf{Expert 5}: With over 5 years of video production experience, he brought fresh new ideas and creativity to all video production projects. He allows for revisions at every stage of the project: scripting, design, storyboard, and draft.
    \item \textbf{Expert 6}: He is a skilled media content creator with a Bachelor's in New Media Communications and expertise in a wide range of technical, creative, and collaborative fields. He specializes in video editing, although he has a deep background in overall video production, including writing, directing, shooting, and editing. He has self-produced over fifty promotional shorts for bands and music venues, as well as over thirty videos for corporate clients.
\end{itemize}

In order to support the user studies, we developed an online website.
The interface is shown in Fig.~\ref{fig:supp_user_study_interface}.
As illustrated, for each task, we display the query description at the top.
Below the description, the video with the keyframe of each shot are shown to the users.
After playing the video, users put down their scores to each question shown below the video player.
We design the user study as answering three 6 score Likert-scale questions.
Specifically, each score stands for a certain opinion:
(i) \textbf{1} -- strongly disagree;
(ii) \textbf{2} -- disagree;
(iii) \textbf{3} -- weakly disagree;
(iv) \textbf{4} -- weakly agree;
(v) \textbf{5} -- agree;
(vi) \textbf{6} -- strongly agree.
It takes about $18$ min for inexperienced users to finish $40$ tasks
and $100$ min for experts to finish all the $200$ tasks.

\begin{figure}[t]
    \vspace{-3px}
	\centering
	\includegraphics[width=0.99\linewidth]{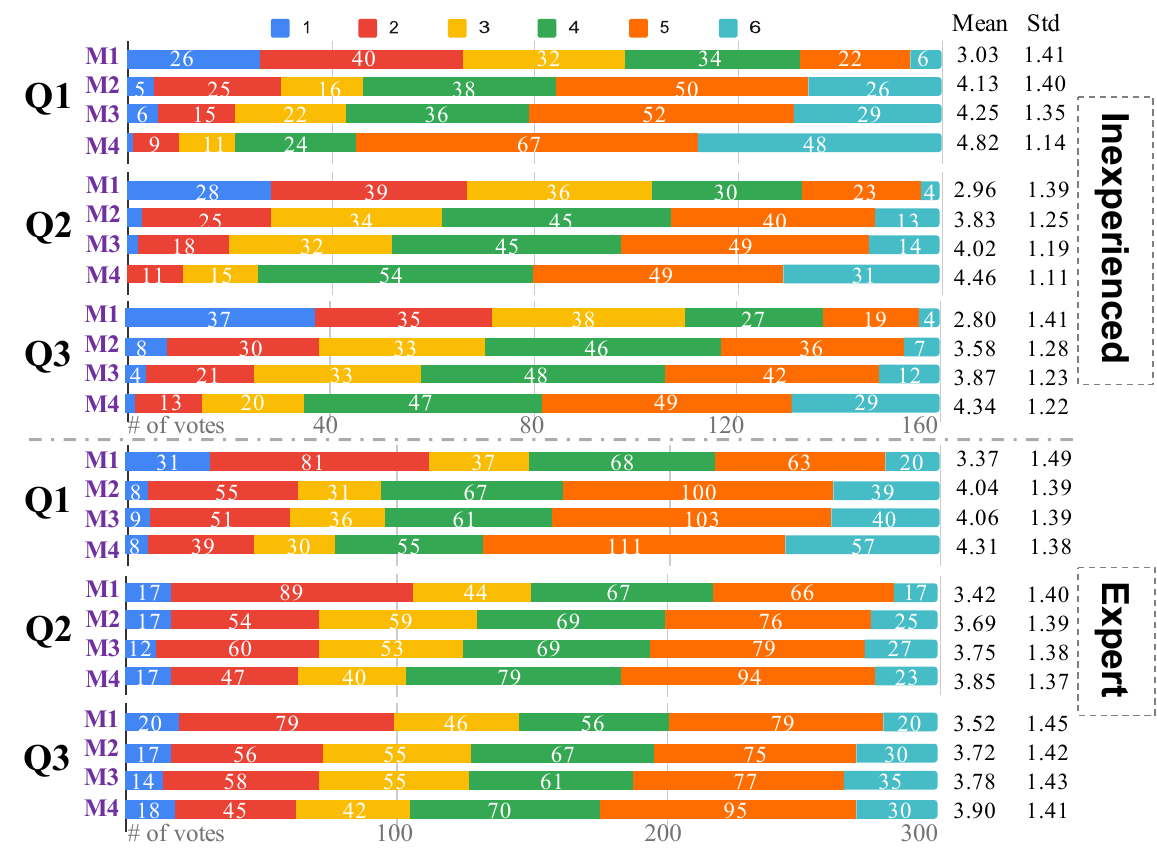}
	\vspace{-7px}
	\caption{\small
    	\textbf{User study score distribution.}
    	M1:PE+HTM+greedy;
    	M2:PE+WT+greedy;
    	M3:PE+WT+beam;
    	M4:AE+WT+beam.
	}
	\label{fig:userstudy_stat}
\end{figure}

Fig.~\ref{fig:userstudy_stat} shows the detailed score distribution of each method
(Zoom in for better view).

\begin{table*}[!t]
    \centering
    \small
\begin{tabular}{ccccc||ccccc||c}
\hline
\multicolumn{5}{c||}{Configurations}                                                                                                                                                               & \multicolumn{5}{c||}{sequence generation}           & \multicolumn{1}{|c}{seq. comp.} \\ \hline
\multicolumn{1}{c|}{Loss}                   & \multicolumn{1}{c|}{Arch.}                         & \multicolumn{1}{c|}{Data}                              & Search & Distortion   & Recall & AOP-1 & AOP-2 & AOP-3 & AOP-S & Acc.      \\ \hline
\multicolumn{1}{c|}{\multirow{4}{*}{MIL-NCE~\cite{milnce}}} & \multicolumn{1}{c|}{\multirow{3}{*}{Parallel}}    & \multicolumn{1}{c|}{\multirow{3}{*}{+WT}}              & beam   & -        & 7.28   & 6.89  & 0.49  & 0.05  & 7.43  & 2.63      \\
\multicolumn{1}{c|}{}                         & \multicolumn{1}{c|}{}                             & \multicolumn{1}{c|}{}                                  & beam   & repl. & 7.33   & 6.96  & 0.63  & 0.11  & 7.70  & 4.01       \\ 
\multicolumn{1}{c|}{}                         & \multicolumn{1}{c|}{}                             & \multicolumn{1}{c|}{}                                  & beam   & repl. + jit. & 7.39   & 7.06  & 0.62  & 0.10  & 7.77  & 4.14     \\ \cline{2-11} 
\multicolumn{1}{c|}{}                         & \multicolumn{1}{c|}{\multirow{3}{*}{Adaptive}} & \multicolumn{1}{c|}{\multirow{3}{*}{+WT}}              & beam   & -        & 7.20   & 6.85  & 0.33  & 0.00  & 7.19  & 2.22      \\
\multicolumn{1}{c|}{}                         & \multicolumn{1}{c|}{}                             & \multicolumn{1}{c|}{}                                  & beam   & repl.        & 7.37   & 7.04  & 0.63  & 0.05  & 7.72  & 4.24      \\
\multicolumn{1}{c|}{}                         & \multicolumn{1}{c|}{}                             & \multicolumn{1}{c|}{}                                  & beam   & repl. + jit. & 7.42   & 7.09  & 0.57  & 0.05  & 7.71  & 4.42      \\ \hline
\end{tabular}
    \caption{\small
        \textbf{Sequence generation and completion results.} 
        +WT means pretrained on HowTo100M~\cite{howto100m} and finetuning on WeTravel; 
        repl. denotes shot replacement while jit. denotes color jittering.
        Parallel denotes Parallel Encoder and Adaptive denotes Adaptive Encoder.
        Comparing the results that uses/removes the Temporal Coherence module,
        we find that TCM is the key for beam search under noisy retrieval results.
    }
    \label{tab:supp-exp-tcm}
\end{table*}

%% file: supp_articles/exp.tex

%
\begin{table}[t]
\small
        \begin{tabular}{c|c|ccc|c}
        \hline
        Dist.                   & Test set   & w/o dist. & repl. & jit.  & Acc   \\ \hline
        \multirow{2}{*}{repl.}       & WT-expedia & 89.15     & 54.32 & -     & 71.74 \\
                                     & WT-test     & 80.60     & 86.10 & -     & 83.35 \\ \hline
        \multirow{2}{*}{\shortstack{repl.\\+jit.}} & WT-expedia & 88.56     & 50.48 & 95.92 & 78.32 \\
                                     & WT-test     & 81.49     & 85.38 & 94.90 & 87.26 \\ \hline
        \end{tabular}
    \caption{\small
    \textbf{Distortion classification results.} Dist. denotes distortion, repl.
    denotes shot replacement, jit. denotes color jittering.
    WT-test denotes the whole testing set.
    WT-expedia denotes the expedia testing set. 
    We present the performance of temporal coherence models on the distortion classification task.
    The classification results on two models -- traning using one distortion (repl.) or two (repl. + jit.) are given here.
    We could learn from the results that the pretext task is learnable.
    }
    \label{tab:exp-dist-cls}
        
\end{table}

%
\begin{table}[t]
    \small
    \centering
        \begin{tabular}{c|c|ccc}
        \hline
        Distortion                          & Test set   & R@1   & R@2   & R@3   \\ \hline
        Random                          & -          & 4.17  & 8.33  & 12.50 \\ \hline
        \multirow{2}{*}{repl.}      & WT-expedia & 12.14 & 21.06 & 28.23 \\
                                        & WT-test     & 10.63 & 19.27 & 26.59 \\ \hline
        \multirow{2}{*}{repl.+jit.} & WT-expedia & 8.90  & 15.67 & 21.58 \\
                                        & WT-test     & 7.76  & 14.44 & 20.56 \\ \hline
        \end{tabular}
    \caption{\small
    \textbf{Sequence ranking results} of different distortion classification models on two testing sets.
    The sequence ranking task aims at testing the model's ability to recognize the valid ordering
    from multiple sequences generated by permuting a valid multi-shot clip.
    Dist. denotes distortion, repl.
    denotes shot replacement, jit. denotes color jittering.
    WT-test denotes the whole testing set.
    WT-expedia denotes the Expedia testing set. 
    We can find that Temporal coherence models are able to distinguish real ordered sequences
    against fake ones, demonstrating its ability to encouraging valid sequencing styles at inference.
    }
    \label{tab:exp-order-rank}
\end{table}

\section{Experiments}
\label{supp:exp}
In this section, we extend the experiment section in the main paper by 
providing extra analysis on sequence generation, sequence completion,
distortion classification and sequence ranking.

%
\noindent\textbf{Dataset.}
We randomly split the WeTravel dataset into \emph{train, val, test} 
subsets using $6$:$1$:$3$ ratios. 
Note that for tasks like video retrieval, 
well-written captions and relatively aligned clip-text pairs are needed. 
Among all the videos in the testing set, 
we found that those from the YouTube channel \emph{Expedia} closely follow our requirements. 
Accordingly, we create an additional test set called WT-expedia, 
different from the original test set called WT-test, 
where all the videos are from the Expedia channel. 
Finally, there are $130$ videos with $4.3$K captions and $13$K shots in WT-expedia.
Note that for tasks using transcript sentences (\ie, 
clip retrieval, sequence generation and sequence completion),
we conduct the experiments on WT-expedia without short videos (number of shots $<$ 50)
since these videos can not provide enough candidate shots as gallery.
Hence we filtered out $4.3$K captions in WT-expedia.
For tasks like distortion classification and sequence ranking,
we conduct experiments both on WT-test (the whole testing set)
and WT-expedia (only expedia videos).
We generate 13K and 996K testing samples in WT-expedia and WT-test respectively
for both distortion classification and sequence ranking.

\noindent\textbf{Training vision-language models.}
We use a two-layer MLP for encoding shot features. 
The input-output sizes of the backbone MLP layers are [512, 2048], [2048, 512].
We use AvgPooling over the time-step dimension as the Consensus Aggregation Function. 
For the text encoder, we follow the same setting as in ~\cite{milnce}.
In Adaptive Encoder, the Context Interaction Function $\sigma(\cdot)$
uses $\mW \in \mathbb{R}^{512\times512}$ as the projection matrix 
and $m(a,b)=\max(0, \text{cos}(a,b))$ as the similarity measurement. 
We adopt a two-layer MLP as the head after text encoder for final text output.
The sizes of the fc layers are [512, 512], [512, 512].
Given a positive pair of video clip and text,
\emph{at most} $3$ closest neighbor texts are sampled as positive candidates, 
\ie, any overlapping sentence within a 3-second interval from the query text.
The number of shots in each video clip is restricted to $10$.
The max length of a text is $16$ words. 

We train all the models using SGD with an initial learning rate of $0.001$ and apply a linear warm-up scheme for the first $1k$ steps. The batch size is $1024$.
We train the models in a single server with $8$ GPUs for $100$ epochs.
The learning rate is multiplied by $0.1$ at the $40$th and $80$th epochs.

\noindent\textbf{Training the temporal coherence module.}
We also use feature-level representations to train the temporal coherence module via the pretext distortion classification task.
For each shot, S3D features and RGB histograms are extracted.
The shot replacement randomly replaces shot features from other videos,
while color jittering is implemented by histogram-level channel shuffle
and histogram stretching.
We randomly crop a sequence with various shots from a video 
and apply distortions with equal probability on at most half of the shots.
Before passing it to the model, we first multiply each feature 
by the similarity between itself and its neighbor features. 
We do this to add temporal correlation priors to the model, 
which helps stabilize the training process.
Later on, two separated two-layer LSTMs encode S3D features and histogram features, respectively.
The input/hidden dimensions of the LSTM for S3D features are 512/512; for histogram: 384/384.
The outputs of each LSTM are fed through a two-layer MLP for distortion prediction.
And the dimensions of the MLP are [896, 128], [128, 2 or 3].
The dimension of the final output depends on the number of applied distortion.

We train all the models using SGD with an initial learning rate of $0.001$ and apply a linear warm-up scheme for the first $1k$ steps. The batch size is 128.
We train the models in a single server with $8$ GPUs for $100$ epochs.
The learning rate is multiplied by $0.1$ at the $40$th and $80$th epochs.

\noindent\textbf{Baseline models.}
For vision-language related tasks, we adopt VSE~\cite{vse} and VSE++~\cite{vsepp}
as baselines. We train them with a pairwise ranking loss
following the same settings as our Parallel/Adaptive encoders.
That being said, the model structures, the training schemes are the same as 
Parallel/Adaptive encoders and the only difference is the loss function.
In VSE, the loss function is defined as:
\begin{equation}
\cL = \sum_i\sum_{i\neq j} [0, f(c_j)^\top g(h_i) - f(c_i)^\top g(h_i)]_+ 
\end{equation}
For VSE++, the loss function is given as:
\begin{equation}
\label{eq:vsepp}
\cL = \sum_i\max_{i\neq j} [0, f(c_j)^\top g(h_i) - f(c_i)^\top g(h_i)]_+ 
\end{equation}
where $[x]_+=max(0,x)$.
For clip retrieval, we also add the frame-based MIL-NCE model as a baseline; we follow its original paper \cite{milnce} details to extract text and video features.
To be specific, we use models in ~\cite{milnce} to extract video feature given as input
uniformly sampled 32 frames of a clip. 

\noindent \textbf{Analysis of Clip Retrieval results.}
Here we provide extra analysis of clip retrieval task.
We could see from the results table in the main paper (main paper, Table.1)
that the performance of VSE models is relatively low in absolute number compared with the random baselines.
By authors' estimation, even after deleting talking-to-camera shots,
there are still $60\%$ of WeTravel clip-text pairs are not aligned.
Hence traditional VSEs are inferior
than noise-contrastive-estimation based methods. 
For instance, the max operation in the VSE++ loss function (Eq.~\ref{eq:vsepp}) produces unstable training due to noisy outliers. 
The testing set is cleaner, but it still has noisy alignments, which
added with the noisy training, results in poor retrieval.

\noindent \textbf{Analysis of Temporal Coherence Module.}
Here we provide the performance of sequence  generation and sequence completion
on the situation that only Beam Search are used, but without the re-ranking steps using the TCM module. 
The results are shown in Tab.~\ref{tab:supp-exp-tcm}. 
We can observe that for both Parallel Encoder and Adaptive Encoder, 
when using beam search without TCM, the performance of sequence completion and 
sequence generation drops a lot. For example, for Parallel Encoder,
AOP-S drops from 7.70/7.77 to 7.43 and sequence compeletion accuracy from 4.14/4.01 to 2.63.
Similar findings appear in results of Adaptive Encoder.
The results show that TCM is the key
for beam search under noisy retrieval results.

\noindent\textbf{Performance of distortion classification.}
In this section, we evaluate distortion classification models only.
Given a testing video, we crop a clip with 2 to 4 shots as a testing sample.
In WT-test, we obtain 996K testing samples while in WT-expedia, the number is 13K.
We test the models on both WT-expedia and WT-test.
As results in Tab.~\ref{tab:exp-dist-cls} show, the distortion classification task can be learned. We also see from the results that accuracy
for shot replacement in WT-expedia is lower than others.
This is because videos within Expedia are more visually similar than those in other channels, making it hard for models to detect the shot replacement distortion.

\noindent\textbf{Sequence Ranking.}
Besides distortion classification, we also test the Temporal Coherence module by ranking a sequence with its permutations. 
To be specific, given a clip with $M$ shots, we generate $M!$ sequences using permutation.
Therefore, there are $1$ clip with valid ordering and another $M!-1$ are considered as clips with fake (or bad) sequencing styles.
The task is to predict which clip is the one with valid sequencing style given all the $M!$ sequences.
And we use $M=4$ here. 
Experimental results
in Tab.~\ref{tab:exp-order-rank} show that models trained even without random shuffling distortion can determine the right order, demonstrating that the models have learned temporal structures.

%% file: supp_articles/dataset.tex

\begin{figure*}[!ht]
\vspace{-10pt}
\captionsetup[subfloat]{captionskip=-3pt}
\centering
	\subfloat[Distribution of shot duration on HowTo100M videos.\label{fig:shot_dur_htm}]{
		\includegraphics[width=0.95\linewidth]{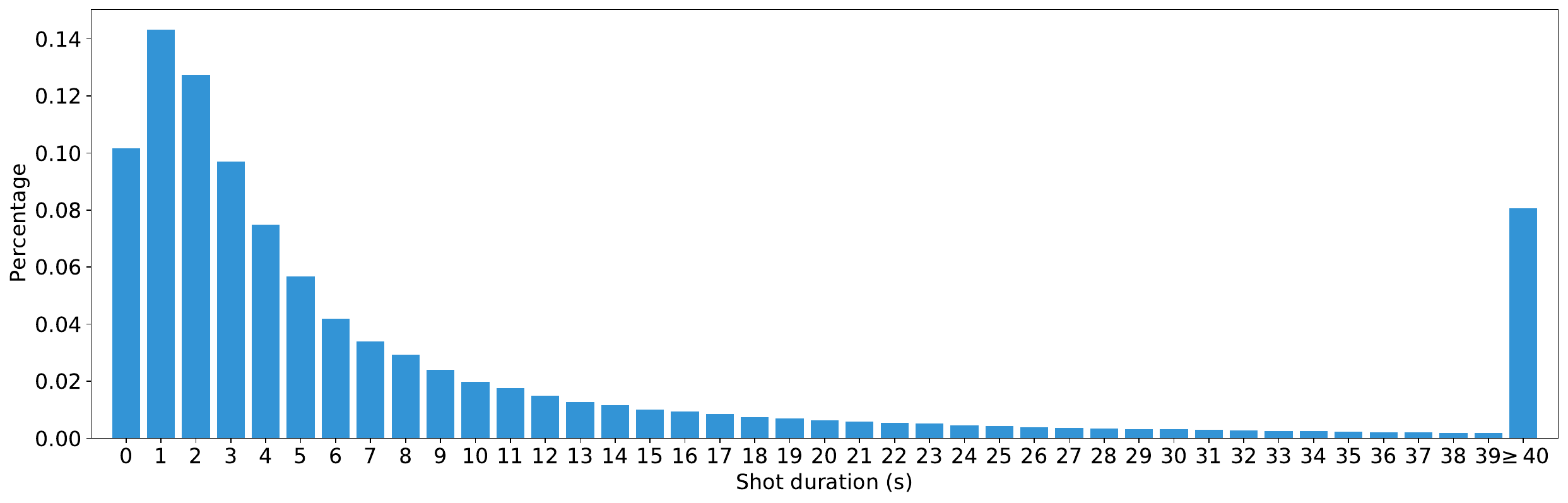}
		} \hfill
		\vspace{-10pt}
	\subfloat[Distribution of shot duration on WeTravel videos. \label{fig:shot_dur_wt}]{
		\includegraphics[width=0.95\linewidth]{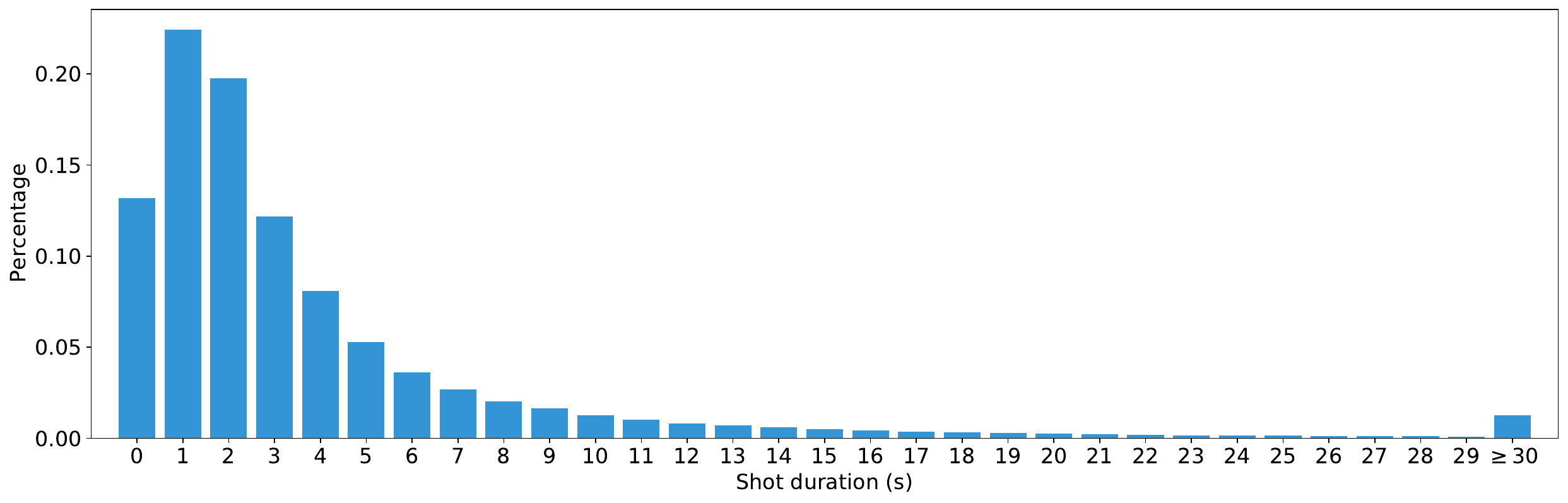}
	}
	\vspace{-10pt}
	\caption{\small
	\textbf{Shot duration statistics.} We compare the distribution of shot duration on both
	HowTo100M (5\% of the videos) and WeTravel. We can see that there are many long shots in HowTo100M.
	The average and median shot duration of HowTo100M videos are \textbf{17.0s} and \textbf{4.4s}.
	The average and median shot duration of WeTravel videos are \textbf{4.6s} and \textbf{2.7s}.}
	\label{fig:shot_dur}
\end{figure*}

\begin{figure*}[!ht]
\vspace{-5pt}
\centering
    \includegraphics[width=0.95\linewidth]{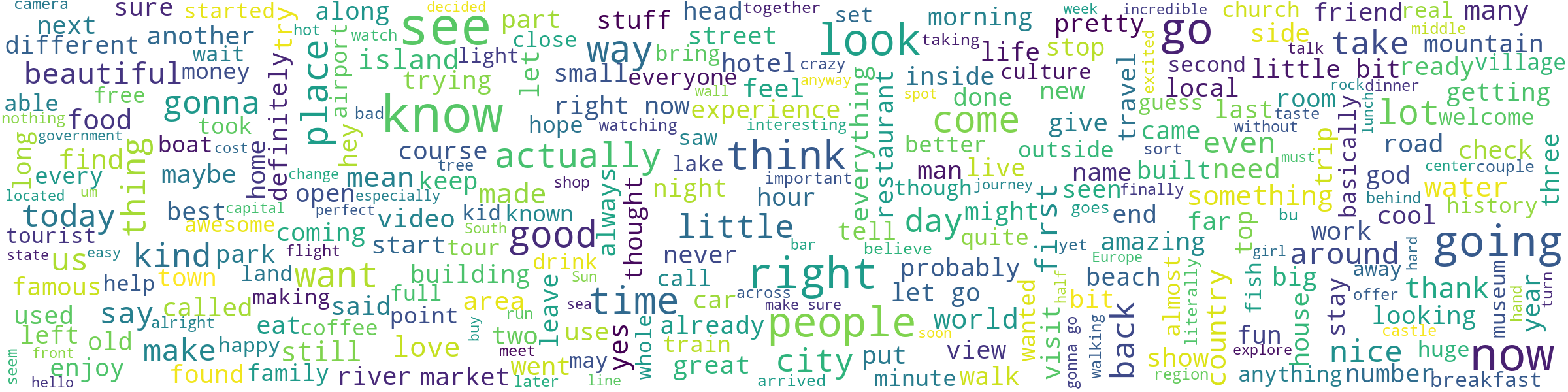}
    \vspace{-10pt}
	\caption{\small
	\textbf{Wordcloud visualization for the transcript texts.}
	We can see many travel related words like \emph{food}, \emph{beautiful} and \emph{beach} \etc.
	}
	\label{fig:wordcloud}
\end{figure*}

\begin{figure*}[!ht]
\vspace{-5pt}
\centering
    \includegraphics[width=0.95\linewidth]{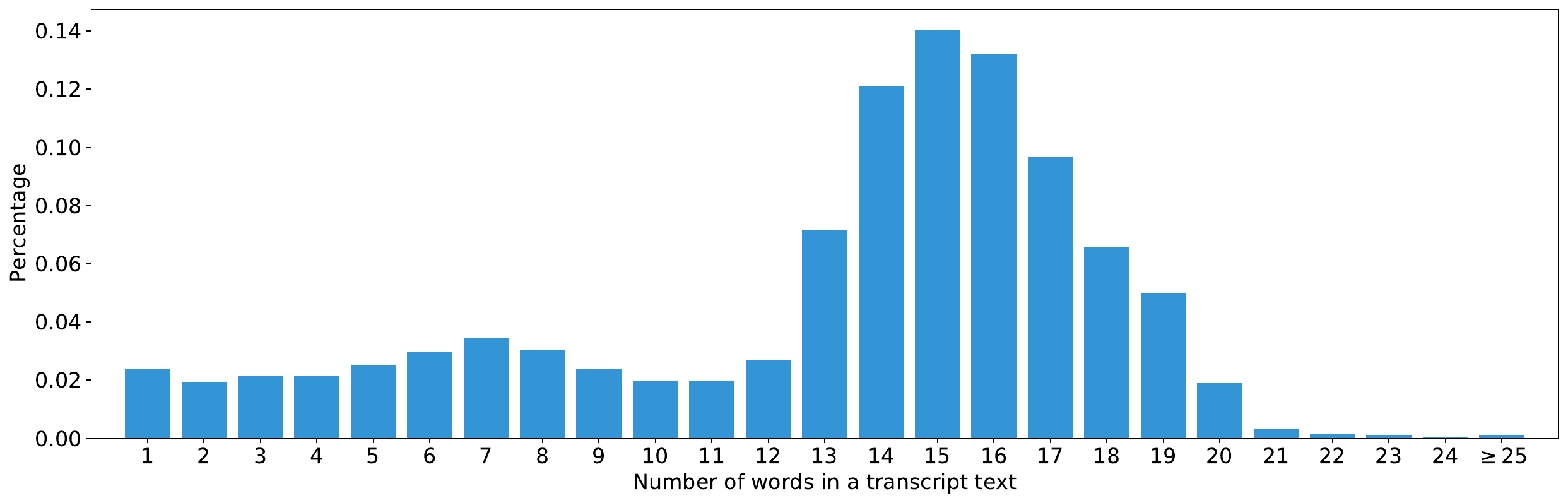}
    \vspace{-10pt}
	\caption{\small
	\textbf{Distribution of text length.} The average and median number of words in a transcript text
	are \textbf{13.1} and \textbf{15} after merging the transcript sentences. 
	Also, \textbf{81.2\%} of the video content are covered with transcript texts.
	}
	\label{fig:cap_dist}
	\vspace{-25pt}
\end{figure*}

\section{WeTravel Dataset}
\label{supp:dataset}

\subsection{Dataset Statistics}
In this section, we provide the detailed statistics of WeTravel.
WeTravel is a dataset to benchmark text-based multi-shot clip creation.
It contains $31.2$K narrated travel videos from YouTube
with transcripts generated by YouTube using Automatic Speech Recognition (ASR).
We choose travel videos
because they tend to be carefully edited to engage audiences, which results in a
large ratio of creatively sequenced clips per video.
Another option is instructional videos. However, instructional videos have two limitations. 
First, being instructional, these videos contain a vast number of talking-to-camera shots. 
Second, given that they depict the actions required to complete a goal, the shots from instructional videos are very long.
The percentage of carefully edited clips could be reflected by shot duration
because \emph{timing} is also one of the key component utilized by editors to shape rhythmic sequences~\cite{pearlman2012cutting}.
That being said, the shots are usually short for well-edited videos.
It's also demonstrated in Fig.~\ref{fig:shot_dur} by comparing
the distribution of shot durations on HowTo100M~\cite{howto100m}(Fig.~\ref{fig:shot_dur_htm}) 
and on our WeTravel(Fig.~\ref{fig:shot_dur_wt}).
We can see that there are many long shots in instructional videos while we can not observe
similar patterns in travel videos.

To further explore the WeTravel dataset, we provide more statistics below:
(1) The wordcloud visualization is shown in Fig.~\ref{fig:wordcloud}
where we see many travel related words are frequently mentioned.
(2) The distribution of word number in a transcript text is shown in Fig.~\ref{fig:cap_dist}.

\subsection{Dataset Examples.}
\begin{figure*}[t]
	\centering
	\includegraphics[width=\linewidth]{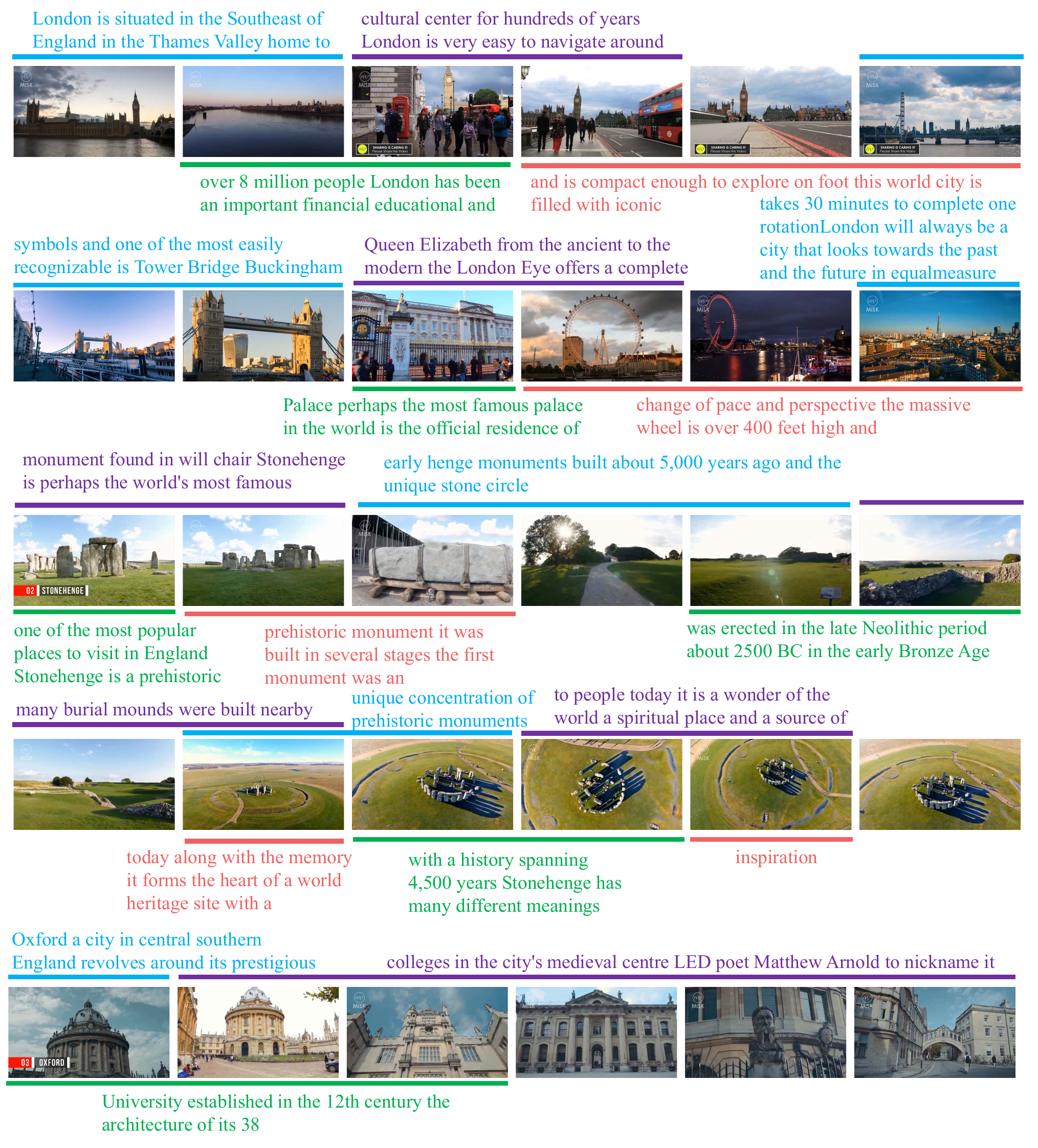}
	\caption{\small
		\textbf{Example pair from WeTravel video (London).}
		We use lines and color texts to show the mapping between each
		transcript text and video content.
		For each line, it covers a clip with multiple complete shots.
		The text in the same color denotes the corresponding
		transcript narration for this clip.
		The video link is \url{https://www.youtube.com/watch?v=hbQbaUeOkDQ}.
		We can see from the example that 
		(i) Transcripts are randomly broken in ASR results in order
		to fit the player width.
		(ii) Some content like \emph{financial educational center ... easy to navigate around}
		are visually irrelevant; over half of the text content are not related to its visual counterpart.
		(iii) Misalignment, \eg \emph{Buckingham} are mentioned in previous caption while shown in the next clip.
	}
	\label{fig:data-england}
\end{figure*}

\begin{figure*}[t]
	\centering
	\includegraphics[width=\linewidth]{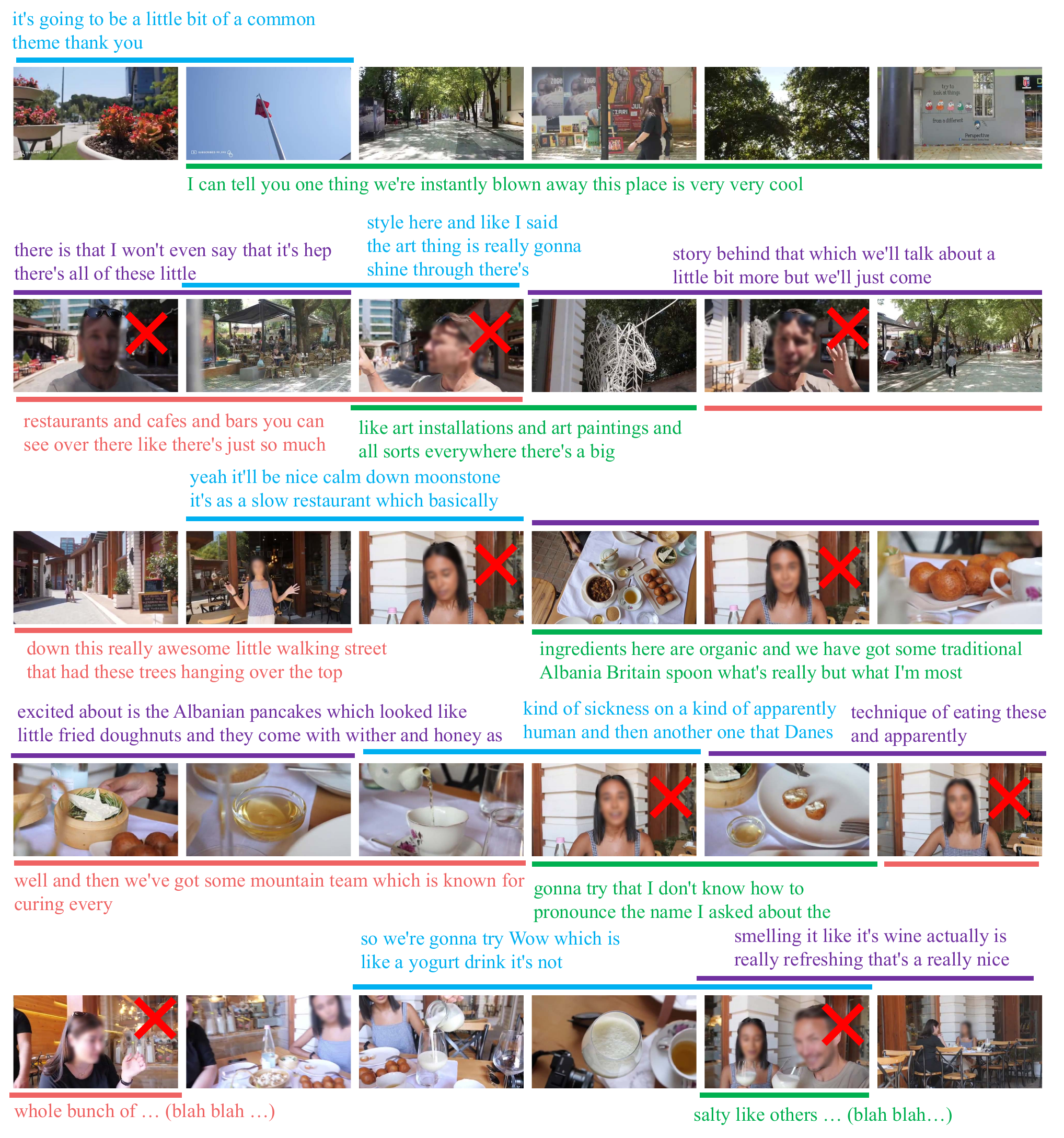}
	\caption{\small
		\textbf{Example pair from WeTravel video (Albania).}
		We use lines and color texts to show the mapping between each
		transcript text and video content.
		For each line, it covers a clip with multiple complete shots.
		The text in the same color denotes the corresponding
		transcript narration for this clip.
		The video link is \url{https://www.youtube.com/watch?v=N2ab7XwTNpg}.
		This example demonstrates a typical scene in travel video -- eating food.
		Apart from misalignment and irrelevant content issue, 
		we can see from the example that there are many talking-to-camera
		shots in this kind of travel vlog videos.
		In WeTravel, we directly filtered out these shots by face detection.
		``\xmark''~denotes the detected talking-to-camera shots.
	}
	\label{fig:data-abania}
	\vspace{30px}
\end{figure*}

\begin{figure*}[t]
	\centering
	\includegraphics[width=\linewidth]{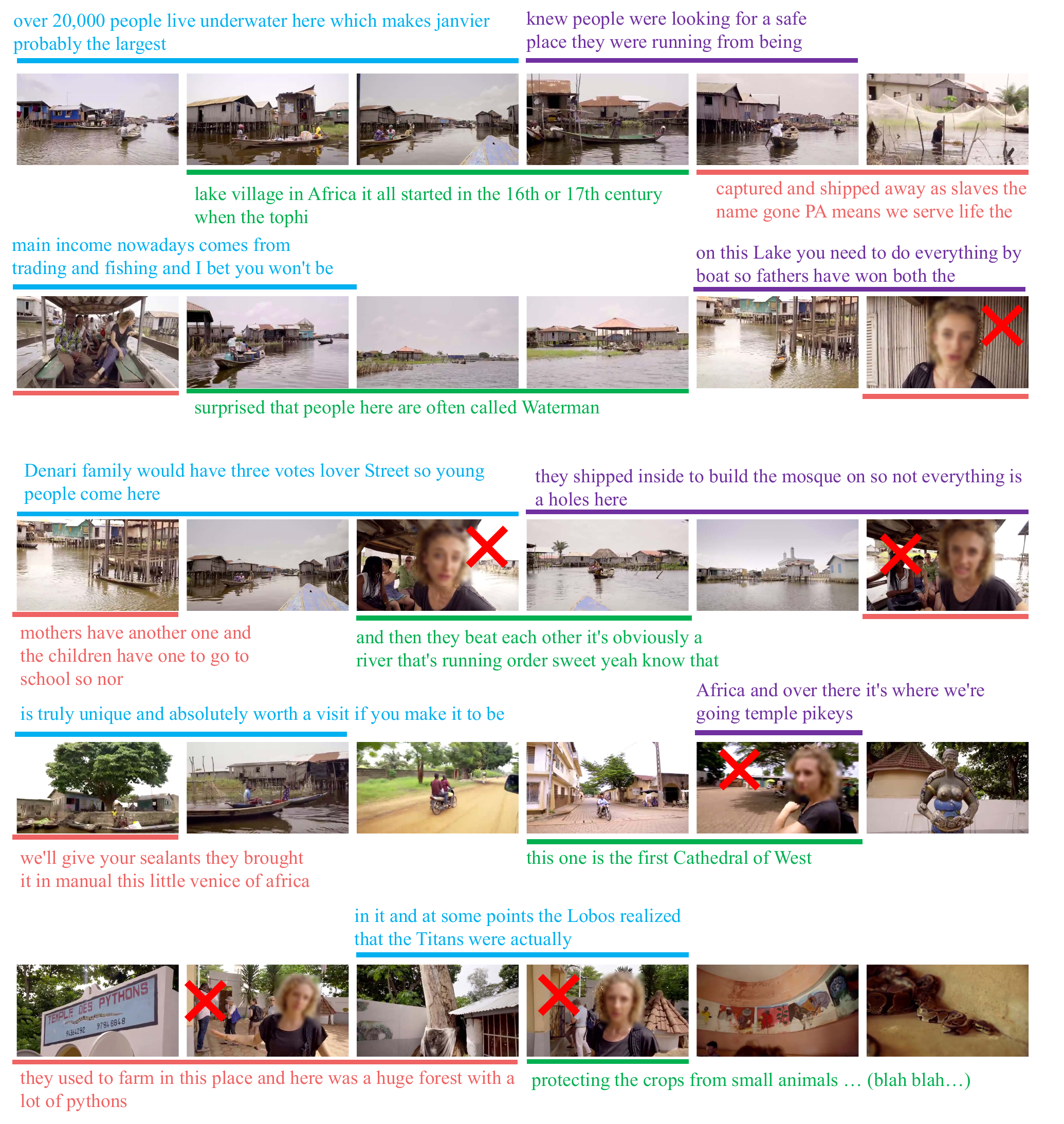}
	\caption{\small
		\textbf{Example pair from WeTravel video (Benin).}
We use lines and color texts to show the mapping between each
transcript text and video content.
For each line, it covers a clip with multiple complete shots.
The text in the same color denotes the corresponding
transcript narration for this clip.
The video link is \url{https://www.youtube.com/watch?v=5Tb0Q3BmfzI}.
This example demonstrates a typical scene in travel video -- local lifestyle.
Other typical topics include \emph{Activities}, \emph{Culture and places of famous}, \etc.
Apart from misalignment and irrelevant content issue, 
we can see from the example that there are many talking-to-camera
shots in this kind of travel vlog videos.
In WeTravel, we directly filtered out these shots by face detection.
``\xmark''~denotes the detected talking-to-camera shots.
	}
	\label{fig:data-benin}
\end{figure*}

\begin{figure*}[t]
	\centering
	\includegraphics[width=\linewidth]{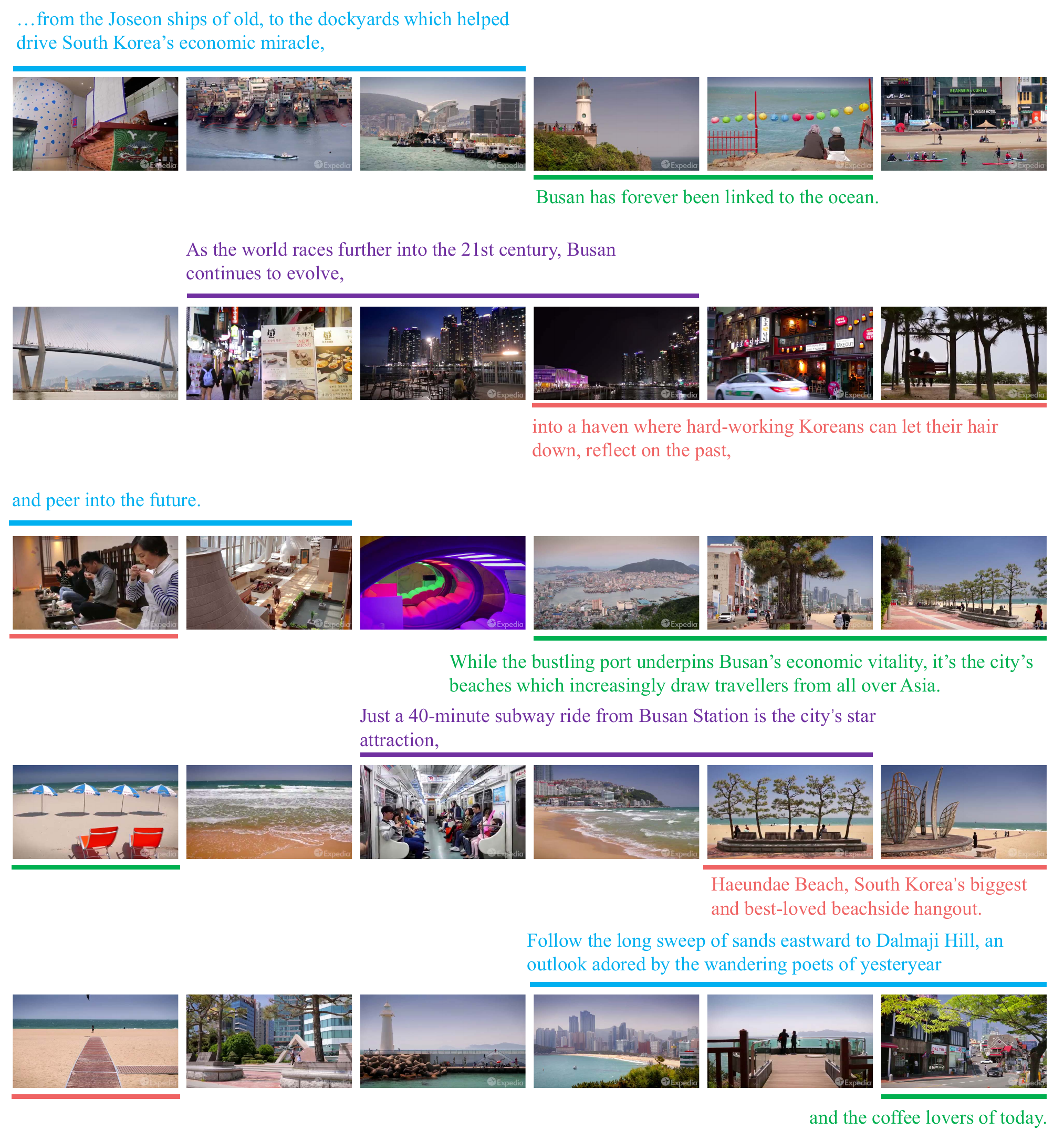}
	\caption{\small
		\textbf{Example pair from WeTravel video (Busan) -- Expedia.}
We use lines and color texts to show the mapping between each
transcript text and video content.
For each line, it covers a clip with multiple complete shots.
The text in the same color denotes the corresponding
transcript narration for this clip.
The video link is \url{https://www.youtube.com/watch?v=Qh3wrmSUqaI}.
This video comes from YouTube Channel \emph{Expedia}.
We can see the difference of this example from previous ones that
(i) the transcripts are manually added and are well-written.
(ii) there are more visually related texts.
(iii) most of the transcript texts are aligned or closely aligned with its counterpart clip.
	}
	\label{fig:data-busan}
\end{figure*}

In this section, we provide $4$ examples of video-transcript pair from WeTravel.
The first three are typical videos with ASR transcripts (Fig.~\ref{fig:data-england},\ref{fig:data-abania},\ref{fig:data-benin}).
While the last one (Fig.~\ref{fig:data-busan}) comes from YouTube Channel \emph{Expedia}.
It can be observed from the examples that Expedia captions are usually well written 
and closely aligned with the visual context.
For other videos, there are issues with

(i) Noisy sentence break;

(ii) Talking-to-camera shots with a lot of unrelated information;

(iii) Misaligned or no related texts for visual content.

%% file: supp_articles/visualization.tex

\section{Visualization}
\label{supp:visualization}
In this section, we present the visualization of created multi-shot clips.
For each video, we display the keyframe of each shot.
We compare the multi-shot clips generated by different methods to show the effectiveness
of each proposed component.
Specifically, the comparisons and demonstrations include:
\begin{itemize}
\vspace{-5pt}
\setlength\itemsep{-0.2em}
    \item Comparisons of Parallel Encoder trained on HowTo100M~\cite{howto100m}
    with Parallel Encoder finetuned on WeTravel in Fig.~\ref{fig:vis_htm_wt_animal},\ref{fig:vis_htm_wt_street}.
    \item Comparisons of inference using greedy search or beam search in Fig.~\ref{fig:vis_greedy_beam_animal},\ref{fig:vis_greedy_beam_st},\ref{fig:vis_greedy_beam_florida},\ref{fig:vis_greedy_beam_float}.
    This is for demonstrating the effectiveness of proposed beam search with temporal coherence module.
    \item Comparison of using different models: Parallel Encoder v.s. Adaptive Encoder
    in Fig.~\ref{fig:vis_pvsi_barcelona},\ref{fig:vis_pvsi_fruit}.
    \item Comparison of all the four methods in Fig.~\ref{fig:vis_all_lantern},~\ref{fig:vis_all_japan},~\ref{fig:vis_all_marina},~\ref{fig:vis_all_beer},~\ref{fig:vis_all_sky}: 
    (i) Parallel Encoder + HTM + Greedy Search; 
    (ii) Parallel Encoder + HTM + WT + Greedy Search; 
    (iii) Parallel Encoder + HTM + WT + Beam Search; 
    (iv) Adaptive Encoder + HTM + WT + Beam Search; 
    \item Visualization results using the model Adaptive Encoder + HTW + WT + Beam Search, shown in Fig.~\ref{fig:vis_final}.
\vspace{-5pt}
\end{itemize}

From the above visualization results, we can observe that:
(1) Fine-tuning on WeTravel brings better visualization results, especially for content matching.
(2) Using beam search with Temporal Coherence Module improves transition smoothness.
(3) Results from Adaptive Encoder are better than Parallel Encoder, especially when
the queries are complex.

Besides, we also provide visualization results on the whole testing set with $1.02$M gallery shots
in Fig.~\ref{fig:vis_whole_set_a},\ref{fig:vis_whole_set_b}.
Apart from YouTube videos, we also download $6$K Vimeo videos 
and construct a gallery called Vimeo-HD with $450$K shots in it.
The results are shown in Fig.~\ref{fig:vis_vimeo_a},\ref{fig:vis_vimeo_b}.

%% file: supp_articles/extra.tex

\section{Extra Information}
\label{supp:extra}

\subsection{Inference Time}
\vspace{-5pt}
To evaluate the efficiency of the proposed
inference pipeline, we present the search engine’s run-time
performance using the full models. 
We show the chart of single GPU inference time in Fig.~\ref{fig:infer_gpu_chart}
and the chart for comparison of GPU and CPU run-time in Fig.~\ref{fig:infer_both_chart}.
Both the GPU and CPU run-time increases linearly when gallery size enlarges.
The search engine is efficient that
GPU run-time goes from 16ms to 243ms
and CPU run-time goes from 46ms to 2.7s when gallery size increases from 10K to 2M.
It's clear that even when the gallery size are large (2M shots) the run-time
on single thread CPU is only 2.7s, demonstrating that the framework is practical appealing.

\input{supp_articles/fig_extra}

\subsection{Story Video}
In the supplementary video,
we show a story video generated by the following paragraph:

\emph{The waterfall in the valley attracts many visitors.
Here is a tip for you to travel there on the weekend.
Drive through the country road and approach to the mountain.
Walk through the trail and you can enjoy the sceneries at the same time.
Then you can see the waterfall.
Let's continue the journey to see more.
Climb up the exuberant forest hills and stay until the nightfall.
Immerse yourselves in the warm sunset at the hilltop.
Until the second day, pick up your package and go home.
You can walk down through the dust road in the forest.
And eat the breakfast in the wooden house at the foot of the mountain.
To finish the journey, you drive home; you can stay somewhere for a while on the way.
There are some temples and old houses in the village near the mountain.
Restart driving, then maybe you will meet animals on the road.
Finally, your car approaches the city and enters the crowded downtown road.
Welcome back end enjoy the dinner in restaurant at the night street.
What a wonderful journey!
}

We use the content-relevant sentences as queries and ignore the transition sentences.
The results are the raw output from the model, without any further post-processing.
By analysis, we find that the video consists of 47 shots from 18 different videos.
It demonstrates that our method can retrieve diverse shots from varied video sources.

%% file: supp_articles/fig_extra.tex
\begin{figure}[!ht]
	\centering
	\includegraphics[width=0.95\linewidth]{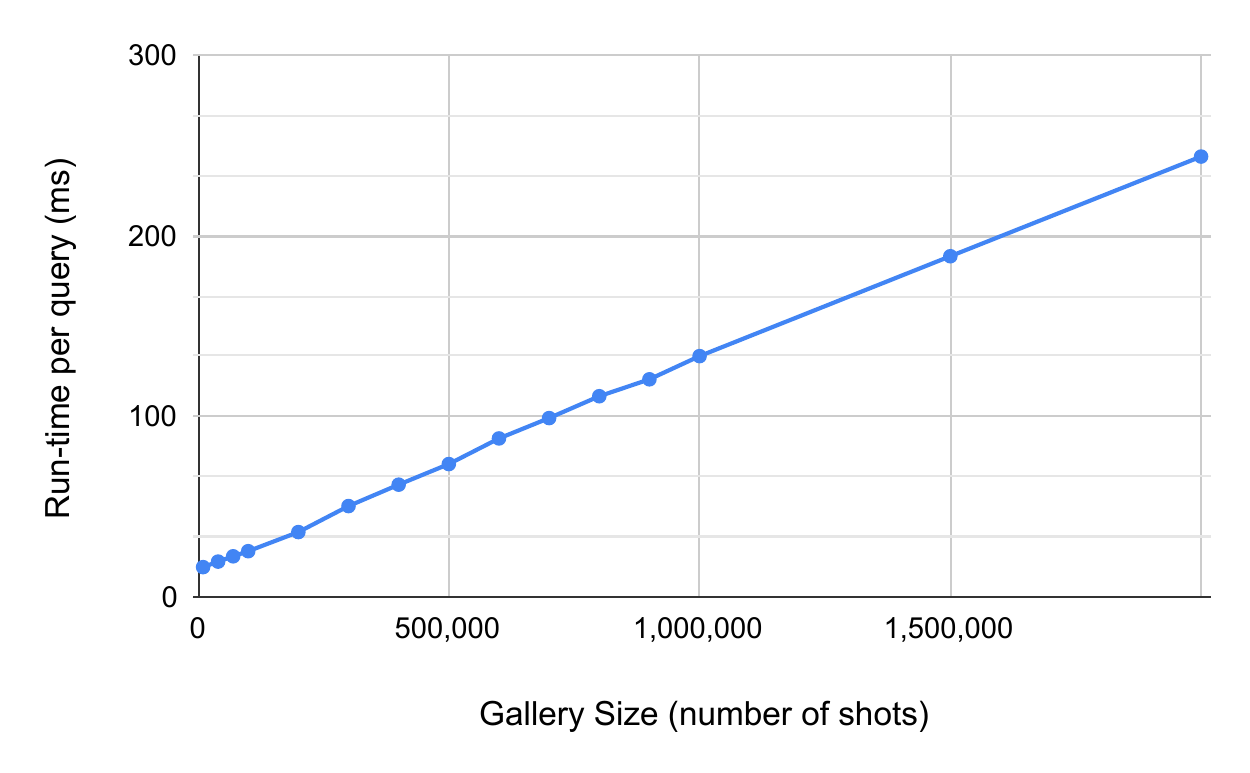}
	\vspace{-15pt}
	\caption{\small
		\textbf{Single GPU inference time.}
	}
	\label{fig:infer_gpu_chart}
	\vspace{-10pt}
\end{figure}

\begin{figure}[!ht]
\vspace{-5pt}
	\centering
	\includegraphics[width=0.95\linewidth]{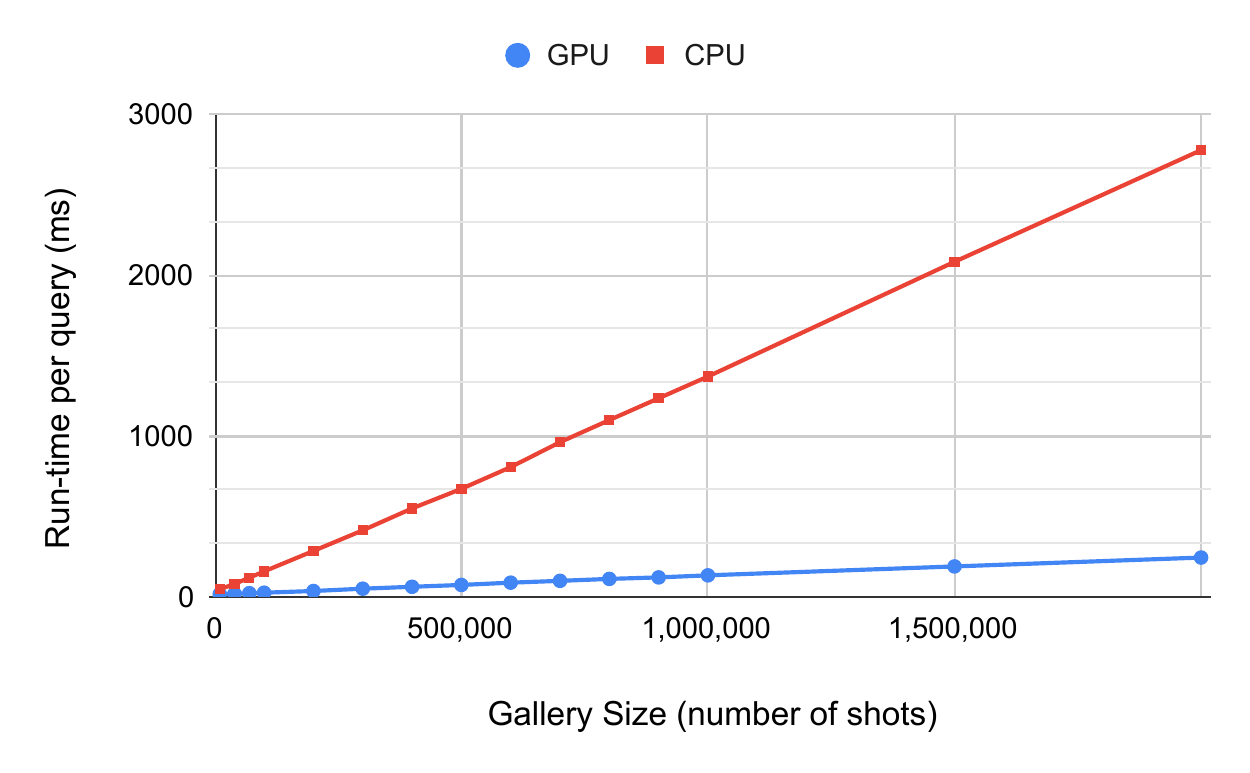}
	\vspace{-15pt}
	\caption{\small
		\textbf{Single GPU inference time and single thread CPU inference time.}
	}
	\label{fig:infer_both_chart}
	\vspace{-10pt}
\end{figure}

%% file: supp_articles/fig_htm_vs_wt.tex
\begin{figure*}[!t]
\captionsetup[subfloat]{captionskip=-3pt}
	\centering
	\begin{minipage}[t]{\linewidth}
		\centering
		\subfloat[Results of Parallel Encoder pretrained on HowTo100M. (Parallel Encoder + HTM + Greedy Search)]{
			\begin{tabular}[t]{c}
				\includegraphics[width=0.24\linewidth]{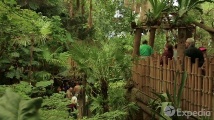}
				\includegraphics[width=0.24\linewidth]{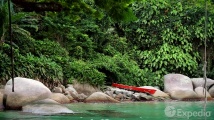}
				\includegraphics[width=0.24\linewidth]{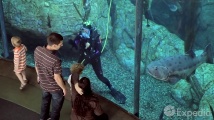}
				\includegraphics[width=0.24\linewidth]{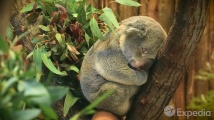}\\
			\end{tabular}
		}
	\end{minipage}
	\quad
	\begin{minipage}[t]{\linewidth}
	\centering
	\subfloat[Results of Parallel Encoder fine-tuned on WeTravel. (Parallel Encoder + HTM + WT + Greedy Search)]{
		\begin{tabular}[t]{c}
			\includegraphics[width=0.24\linewidth]{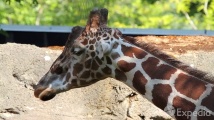}
			\includegraphics[width=0.24\linewidth]{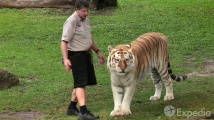}
			\includegraphics[width=0.24\linewidth]{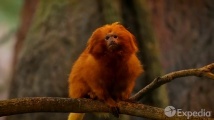}
			\includegraphics[width=0.24\linewidth]{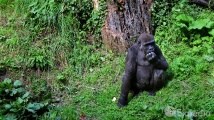}\\	
		\end{tabular}
	}
\end{minipage}
\vspace{-10pt}
	\caption{\small
	\textbf{Can fine-tuning on WeTravel bring better results?}
	Query text: \emph{Animals in the zoo eat and play happily.}
	From the results we see that clip in (a) fails to retrieve exactly relevant shots. 
	The first three shots are related with zoo but do not capture the whole content in the text query.
	While in (b), all the retrieved shots are related with an animal that is either eating or playing.
	So we can see that Parallel Encoder finetuned on WeTravel is better than directly training on HTM.
	}
	\label{fig:vis_htm_wt_animal}
\end{figure*}

\begin{figure*}[!t]
\captionsetup[subfloat]{captionskip=-3pt}
	\centering
	\begin{minipage}[t]{\linewidth}
		\centering
		\subfloat[Results of Parallel Encoder pretrained on HowTo100M. (Parallel Encoder + HTM + Greedy Search)]{
			\begin{tabular}[t]{c}
				\includegraphics[width=0.24\linewidth]{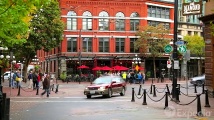}
				\includegraphics[width=0.24\linewidth]{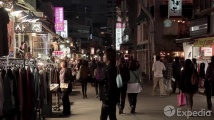}
				\includegraphics[width=0.24\linewidth]{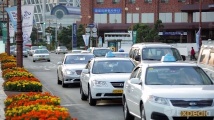}
				\includegraphics[width=0.24\linewidth]{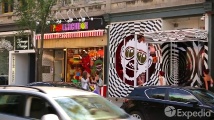}\\
			\end{tabular}
		}
	\end{minipage}
	\quad
	\begin{minipage}[t]{\linewidth}
	\centering
	\subfloat[Results of Parallel Encoder fine-tuned on WeTravel. (Parallel Encoder + HTM + WT + Greedy Search)]{
		\begin{tabular}[t]{c}
			\includegraphics[width=0.24\linewidth]{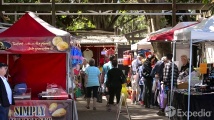}
			\includegraphics[width=0.24\linewidth]{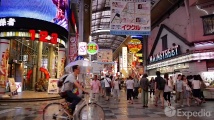}
			\includegraphics[width=0.24\linewidth]{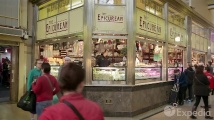}
			\includegraphics[width=0.24\linewidth]{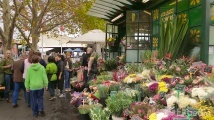}\\	
		\end{tabular}
	}
\end{minipage}
\vspace{-10pt}
	\caption{\small
	\textbf{Can fine-tuning on WeTravel bring better results?}
	Query text: \emph{In the happy bustle of its shopping streets we see many stores and restaurants.}
	In this example, we see that (a) contains some shots showing streets, 
	but not very precisely meet the requirement of ``shopping street'' (\eg shot\#1, \#3).
	But for (b) we see all the places are market streets, hence is more accurate in terms of content retrieval.
	Therefore, we see that Parallel Encoder finetuned on WeTravel is better than directly training on HTM.
	}
	\label{fig:vis_htm_wt_street}
\end{figure*}

%% file: supp_articles/fig_greedy_vs_beam.tex
\begin{figure*}[!t]
\captionsetup[subfloat]{captionskip=-3pt}
	\centering
	\begin{minipage}[t]{\linewidth}
		\centering
		\subfloat[Results of inference using greedy search. (Parallel Encoder + HTM + WT + Greedy Search)]{
			\begin{tabular}[t]{c}
				\includegraphics[width=0.24\linewidth]{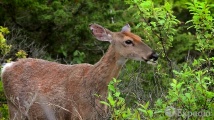}
				\includegraphics[width=0.24\linewidth]{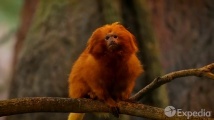}
				\includegraphics[width=0.24\linewidth]{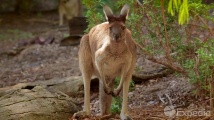}
				\includegraphics[width=0.24\linewidth]{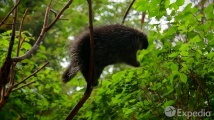}\\
			\end{tabular}
		}
	\end{minipage}
	\quad
	\begin{minipage}[t]{\linewidth}
	\centering
	\subfloat[Results of inference using beam search. (Parallel Encoder + HTM + WT + Beam Serach)]{
		\begin{tabular}[t]{c}
			\includegraphics[width=0.24\linewidth]{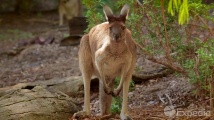}
			\includegraphics[width=0.24\linewidth]{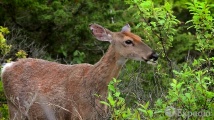}
			\includegraphics[width=0.24\linewidth]{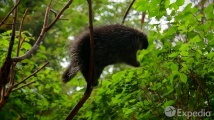}
			\includegraphics[width=0.24\linewidth]{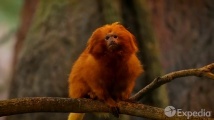}\\	
		\end{tabular}
	}
\end{minipage}
\vspace{-10pt}
	\caption{\small
	\textbf{Can beam search with temporal coherence module bring better sequencing styles?}
	Query text: \emph{Small animals prefer resting on the tree.}
	Firstly, the retrieved shots in (a) and (b) are the same using eigher greedy search or beam search with TCM.
	However, the sequencing styles in (b) are considered better since (i) the colors transit smoothly and 
	(ii) the shot scales change from medium to closer linearly.
	Hence we can see that Temporal Coherence Module can encourage better sequencing styles.
	}
	\label{fig:vis_greedy_beam_animal}
\end{figure*}

\begin{figure*}[!t]
\captionsetup[subfloat]{captionskip=-3pt}
	\centering
	\begin{minipage}[t]{\linewidth}
		\centering
		\subfloat[Results of inference using greedy search. (Parallel Encoder + HTM + WT + Greedy Serach)]{
			\begin{tabular}[t]{c}
				\includegraphics[width=0.24\linewidth]{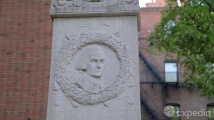}
				\includegraphics[width=0.24\linewidth]{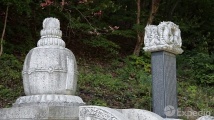}
				\includegraphics[width=0.24\linewidth]{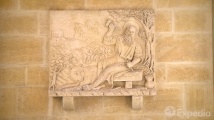}
				\includegraphics[width=0.24\linewidth]{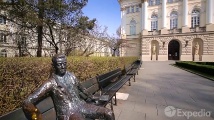}\\
			\end{tabular}
		}
	\end{minipage}
	\quad
	\begin{minipage}[t]{\linewidth}
	\centering
	\subfloat[Results of inference using beam search. (Parallel Encoder + HTM + WT + Beam Search)]{
		\begin{tabular}[t]{c}
			\includegraphics[width=0.24\linewidth]{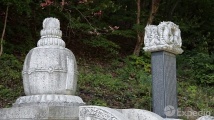}
			\includegraphics[width=0.24\linewidth]{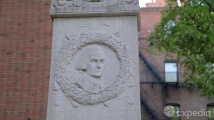}
			\includegraphics[width=0.24\linewidth]{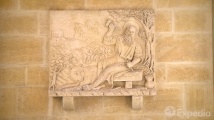}
			\includegraphics[width=0.24\linewidth]{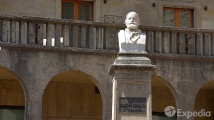}\\	
		\end{tabular}
	}
\end{minipage}
\vspace{-10pt}
	\caption{\small
	\textbf{Can beam search with temporal coherence module bring better sequencing style?}
	Query text: \emph{St Petersburg has become something of a center for sculpture.}
	From this example, we see that TCM help adjust sequencing style by replacing the last shot to another one
	with a similar scale as first three shots. 
	So that shots in (b) have similar shot scales, making the generated clip look better as a whole.
	}
	\label{fig:vis_greedy_beam_st}
\end{figure*}

\begin{figure*}[!t]
\captionsetup[subfloat]{captionskip=-3pt}
	\centering
	\begin{minipage}[t]{\linewidth}
		\centering
		\subfloat[Results of inference using greedy search. (Parallel Encoder + HTM + WT + Greedy Search)]{
			\begin{tabular}[t]{c}
				\includegraphics[width=0.24\linewidth]{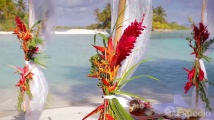}
				\includegraphics[width=0.24\linewidth]{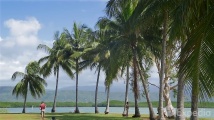}
				\includegraphics[width=0.24\linewidth]{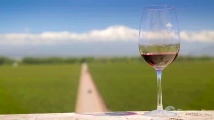}
				\includegraphics[width=0.24\linewidth]{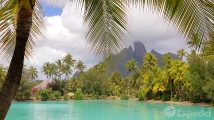}\\
			\end{tabular}
		}
	\end{minipage}
	\quad
	\begin{minipage}[t]{\linewidth}
	\centering
	\subfloat[Results of inference using beam search. (Parallel Encoder + HTM + WT + Beam Search)]{
		\begin{tabular}[t]{c}
			\includegraphics[width=0.24\linewidth]{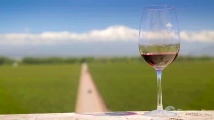}
			\includegraphics[width=0.24\linewidth]{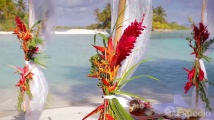}
			\includegraphics[width=0.24\linewidth]{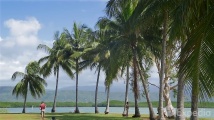}
			\includegraphics[width=0.24\linewidth]{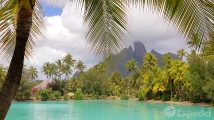}\\	
		\end{tabular}
	}
\end{minipage}
\vspace{-10pt}
	\caption{\small
	\textbf{Can beam search with temporal coherence module bring better sequencing style?}
	Query text: \emph{Soaking up all that color under the Florida sun can be thirsty work.}
	Comparing clips from (a) and (b), we see that (b) is generally better because the shot scale changes from close
	to wide gradually without sudden changes. 
	Also, similar contents are placed closer (\eg the trees), encouraging the continuity in video editing.
	}
	\label{fig:vis_greedy_beam_florida}
\end{figure*}

\begin{figure*}[!t]
\captionsetup[subfloat]{captionskip=-3pt}
	\centering
	\begin{minipage}[t]{\linewidth}
		\centering
		\subfloat[Results of inference using greedy search. (Parallel Encoder + HTM + WT + Greedy Search)]{
			\begin{tabular}[t]{c}
				\includegraphics[width=0.24\linewidth]{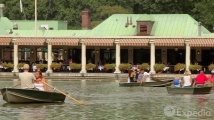}
				\includegraphics[width=0.24\linewidth]{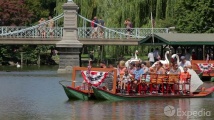}
				\includegraphics[width=0.24\linewidth]{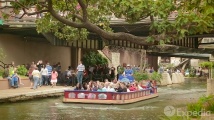}
				\includegraphics[width=0.24\linewidth]{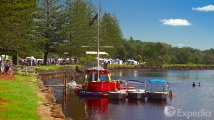}\\
			\end{tabular}
		}
	\end{minipage}
	\quad
	\begin{minipage}[t]{\linewidth}
	\centering
	\subfloat[Results of inference using beam search. (Parallel Encoder + HTM + WT + Beam Search)]{
		\begin{tabular}[t]{c}
			\includegraphics[width=0.24\linewidth]{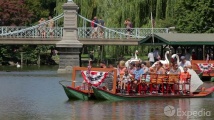}
			\includegraphics[width=0.24\linewidth]{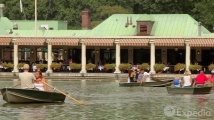}
			\includegraphics[width=0.24\linewidth]{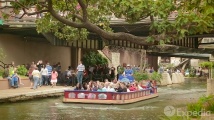}
			\includegraphics[width=0.24\linewidth]{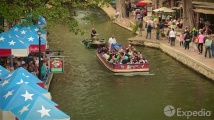}\\	
		\end{tabular}
	}
\end{minipage}
\vspace{-10pt}
	\caption{\small
	\textbf{Can beam search with temporal coherence module bring better sequencing style?}
	Query text: \emph{a floating feast of some of the finest civic buildings, bridges and parks.}
	We can observe from (b) that the color grows warmer across shots but can not find similar patterns in (a), which demonstrates
	a better sequencing style is observed by sequence generated using beam search. Another finding in this example
	is that the third and fourth shots are from the same scene, but with different camera positions.
	}
	\label{fig:vis_greedy_beam_float}
\end{figure*}

%% file: supp_articles/fig_p_vs_i.tex
\begin{figure*}[!t]
\captionsetup[subfloat]{captionskip=-3pt}
	\centering
	\begin{minipage}[t]{\linewidth}
		\centering
		\subfloat[Results of Parallel Encoder. (Parallel Encoder + HTM + WT + Beam Search)]{
			\begin{tabular}[t]{c}
				\includegraphics[width=0.24\linewidth]{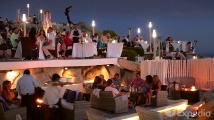}
				\includegraphics[width=0.24\linewidth]{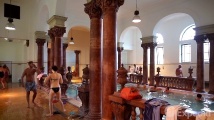}
				\includegraphics[width=0.24\linewidth]{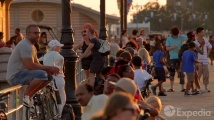}
				\includegraphics[width=0.24\linewidth]{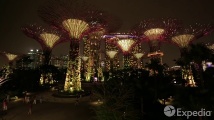}\\
			\end{tabular}
		}
	\end{minipage}
	\quad
	\begin{minipage}[t]{\linewidth}
	\centering
	\subfloat[Results of of Adaptive Encoder. (Adaptive Encoder + HTM + WT + Beam Search)]{
		\begin{tabular}[t]{c}
			\includegraphics[width=0.24\linewidth]{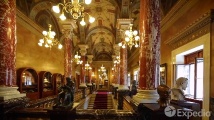}
			\includegraphics[width=0.24\linewidth]{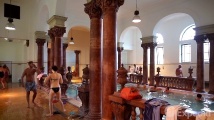}
			\includegraphics[width=0.24\linewidth]{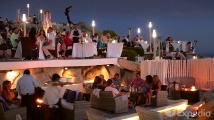}
			\includegraphics[width=0.24\linewidth]{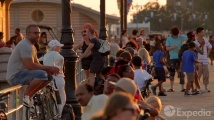}\\	
		\end{tabular}
	}
\end{minipage}
\vspace{-10pt}
	\caption{\small
	\textbf{Comparison of results on Parallel Encoder with Adaptive Encoder.}
	Query text: \emph{Barcelona is sometimes chaotic, often intense, and always, always seductive.}
	In this comparison, we see that Adaptive Encoder helps retrieve more accurate content since all the shots in
	(b) are European style while the last shot in (a) is Singapore style.
	}
	\label{fig:vis_pvsi_barcelona}
\end{figure*}

\begin{figure*}[!t]
\captionsetup[subfloat]{captionskip=-3pt}
	\centering
	\begin{minipage}[t]{\linewidth}
		\centering
		\subfloat[Results of Parallel Encoder. (Parallel Encoder + HTM + WT + Beam Search)]{
			\begin{tabular}[t]{c}
				\includegraphics[width=0.24\linewidth]{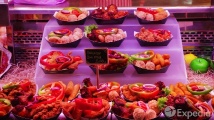}
				\includegraphics[width=0.24\linewidth]{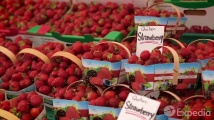}
				\includegraphics[width=0.24\linewidth]{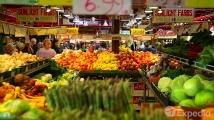}
				\includegraphics[width=0.24\linewidth]{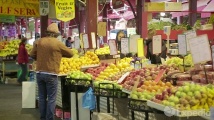}\\
			\end{tabular}
		}
	\end{minipage}
	\quad
	\begin{minipage}[t]{\linewidth}
	\centering
	\subfloat[Results of of Adaptive Encoder. (Adaptive Encoder + HTM + WT + Beam Search)]{
		\begin{tabular}[t]{c}
			\includegraphics[width=0.24\linewidth]{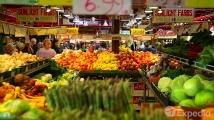}
			\includegraphics[width=0.24\linewidth]{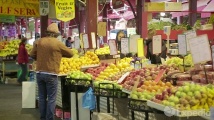}
			\includegraphics[width=0.24\linewidth]{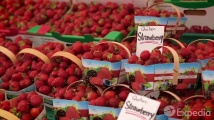}
			\includegraphics[width=0.24\linewidth]{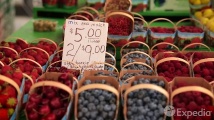}\\	
		\end{tabular}
	}
\end{minipage}
\vspace{-10pt}
	\caption{\small
	\textbf{Comparison of results on Parallel Encoder with Adaptive Encoder.}
	Query text: \emph{or fill your basket with the fruits of Bordeaux’s fields, forests and waters, at Capucins.}
	In this example, we see both the results of Parallel Encoder and Adaptive Encoder benefit
	from temporal coherence module that similar contents are placed successively. 
	But the retrieval results in (b) is better than in (a) because the first shot in (a) displays more objects
	than fruits.
	}
	\label{fig:vis_pvsi_fruit}
\end{figure*}

%% file: supp_articles/fig_all.tex
\begin{figure*}[!t]
\captionsetup[subfloat]{captionskip=-3pt}
	\centering
	\begin{minipage}[t]{\linewidth}
		\centering
		\subfloat[Results of Parallel Encoder+ HTM + Greedy Search]{
			\begin{tabular}[t]{c}
				\includegraphics[width=0.24\linewidth]{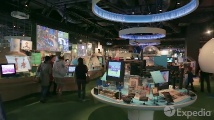}
				\includegraphics[width=0.24\linewidth]{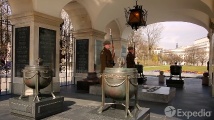}
				\includegraphics[width=0.24\linewidth]{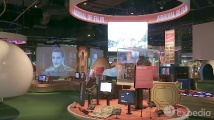}
				\includegraphics[width=0.24\linewidth]{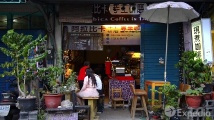}\\
			\end{tabular}
		}
	\end{minipage}
	\quad
	\begin{minipage}[t]{\linewidth}
	\centering
	\subfloat[Results of Parallel Encoder+ HTM + WT + Greedy Search]{
		\begin{tabular}[t]{c}
			\includegraphics[width=0.24\linewidth]{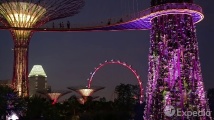}
			\includegraphics[width=0.24\linewidth]{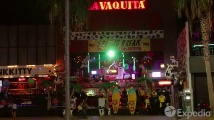}
			\includegraphics[width=0.24\linewidth]{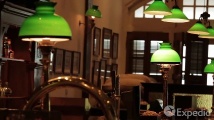}
			\includegraphics[width=0.24\linewidth]{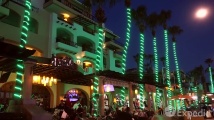}\\	
		\end{tabular}
	}
\end{minipage}
	\quad
	\begin{minipage}[t]{\linewidth}
	\centering
	\subfloat[Results of Parallel Encoder + HTM + WT + Beam Search]{
		\begin{tabular}[t]{c}
			\includegraphics[width=0.24\linewidth]{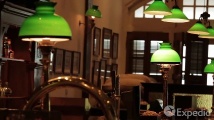}
			\includegraphics[width=0.24\linewidth]{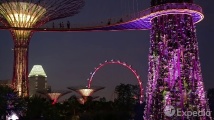}
			\includegraphics[width=0.24\linewidth]{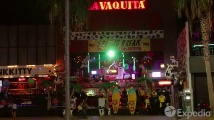}
			\includegraphics[width=0.24\linewidth]{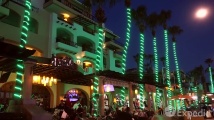}\\	
		\end{tabular}
	}
\end{minipage}
	\quad
	\begin{minipage}[t]{\linewidth}
	\centering
	\subfloat[Results of Adaptive Encoder + HTM + WT + Beam Search]{
		\begin{tabular}[t]{c}
			\includegraphics[width=0.24\linewidth]{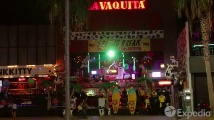}
			\includegraphics[width=0.24\linewidth]{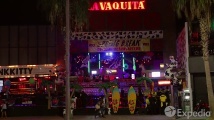}
			\includegraphics[width=0.24\linewidth]{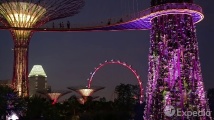}
			\includegraphics[width=0.24\linewidth]{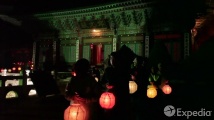}\\	
		\end{tabular}
	}
\end{minipage}
\vspace{-10pt}
	\caption{\small
	\textbf{Comparison of results on all the four models.}
	Query text: \emph{The city erupts in light each night, just like the displays of its lantern festival.}
	This example compares the results from all the four models. 
	The result in (a) demonstrate the failure of models pretrained on HTM that could not retrieve relevant content.
	All the other results from (b) - (d) are more reasonable in terms of content retrieval,
	showing the improvement of using WeTravel for finetuning.
	Compare with (b) and (c), the sequencing style in (c) is considered better because it starts from
	a medium shot with indoor scene while changes to wide shots with outdoor night scenes.
	It demonstrates the effectiveness of proposed beam search with TCM module.
	While for the result in (d), we could observe that the shots displayed more accurately describe
	the visual content in text query.
	But a flaw appears that the first and the second shots are similar.
	This is because the shot boundary detector accidentally cut the whole shot into two parts.
	}
	\label{fig:vis_all_lantern}
\end{figure*}

\begin{figure*}[!t]
\captionsetup[subfloat]{captionskip=-3pt}
	\centering
	\begin{minipage}[t]{\linewidth}
		\centering
		\subfloat[Results of Parallel Encoder+ HTM + Greedy Search]{
			\begin{tabular}[t]{c}
				\includegraphics[width=0.24\linewidth]{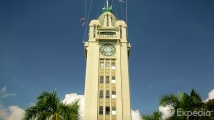}
				\includegraphics[width=0.24\linewidth]{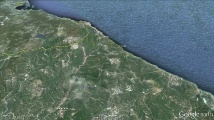}
				\includegraphics[width=0.24\linewidth]{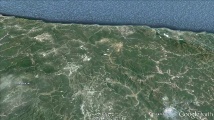}
				\includegraphics[width=0.24\linewidth]{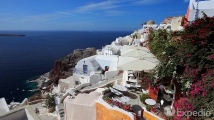}\\
			\end{tabular}
		}
	\end{minipage}
	\quad
	\begin{minipage}[t]{\linewidth}
	\centering
	\subfloat[Results of Parallel Encoder+ HTM + WT + Greedy Search]{
		\begin{tabular}[t]{c}
			\includegraphics[width=0.24\linewidth]{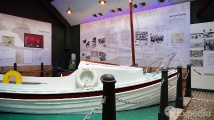}
			\includegraphics[width=0.24\linewidth]{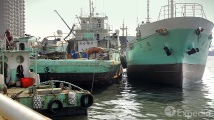}
			\includegraphics[width=0.24\linewidth]{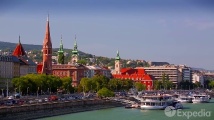}
			\includegraphics[width=0.24\linewidth]{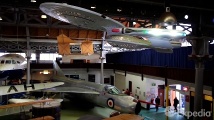}\\	
		\end{tabular}
	}
\end{minipage}
	\quad
	\begin{minipage}[t]{\linewidth}
	\centering
	\subfloat[Results of Parallel Encoder + HTM + WT + Beam Search]{
		\begin{tabular}[t]{c}
			\includegraphics[width=0.24\linewidth]{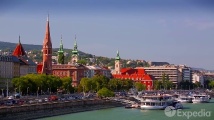}
			\includegraphics[width=0.24\linewidth]{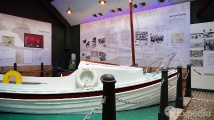}
			\includegraphics[width=0.24\linewidth]{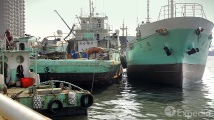}
			\includegraphics[width=0.24\linewidth]{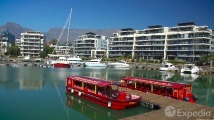}\\	
		\end{tabular}
	}
\end{minipage}
	\quad
	\begin{minipage}[t]{\linewidth}
	\centering
	\subfloat[Results of Adaptive Encoder + HTM + WT + Beam Search]{
		\begin{tabular}[t]{c}
			\includegraphics[width=0.24\linewidth]{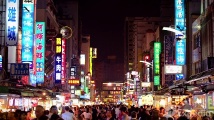}
			\includegraphics[width=0.24\linewidth]{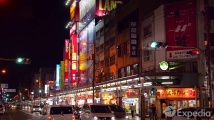}
			\includegraphics[width=0.24\linewidth]{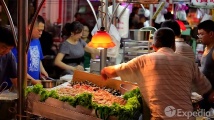}
			\includegraphics[width=0.24\linewidth]{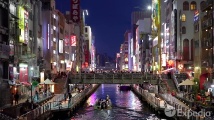}\\	
		\end{tabular}
	}
\end{minipage}
\vspace{-10pt}
	\caption{\textbf{Comparison of results on all the four models.}
	Query text: \emph{This harbor city is now a dynamic fusion of Japanese sophistication.}
	In this case, we see that the resulting multi-shot clip in (a) almost fail to retrieve relevant shots.
	While for (b) and (c), we could see some concept of \emph{harbor} shown in the shots.
	Compared with (b), the clip in (c) retrieves more reasonable last shot thanks to the TCM module
	that help to raise the score of the last shot.
	The overall score in (d) are considered way better than others in the user study because it generates content
	related to ``a city'' with ``Japanese sophistication''.
	}
	\label{fig:vis_all_japan}
\end{figure*}

\begin{figure*}[!t]
\captionsetup[subfloat]{captionskip=-3pt}
	\centering
	\begin{minipage}[t]{\linewidth}
		\centering
		\subfloat[Results of Parallel Encoder + HTM + Greedy Search]{
			\begin{tabular}[t]{c}
				\includegraphics[width=0.24\linewidth]{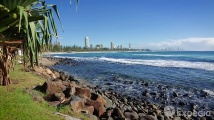}
				\includegraphics[width=0.24\linewidth]{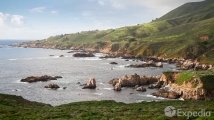}
				\includegraphics[width=0.24\linewidth]{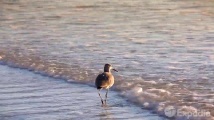}
				\includegraphics[width=0.24\linewidth]{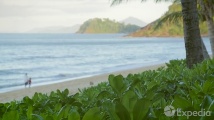}\\
			\end{tabular}
		}
	\end{minipage}
	\quad
	\begin{minipage}[t]{\linewidth}
	\centering
	\subfloat[Results of Parallel Encoder + HTM + WT + Greedy Search]{
		\begin{tabular}[t]{c}
			\includegraphics[width=0.24\linewidth]{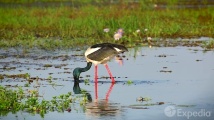}
			\includegraphics[width=0.24\linewidth]{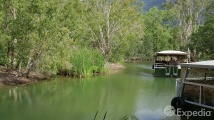}
			\includegraphics[width=0.24\linewidth]{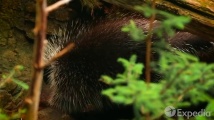}
			\includegraphics[width=0.24\linewidth]{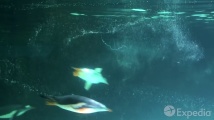}\\	
		\end{tabular}
	}
\end{minipage}
	\quad
	\begin{minipage}[t]{\linewidth}
	\centering
	\subfloat[Results of Parallel Encoder + HTM + WT + Beam Search]{
		\begin{tabular}[t]{c}
			\includegraphics[width=0.24\linewidth]{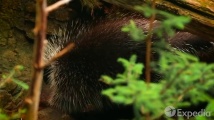}
			\includegraphics[width=0.24\linewidth]{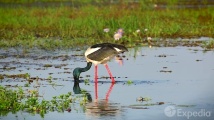}
			\includegraphics[width=0.24\linewidth]{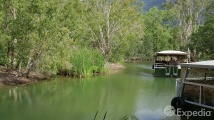}
			\includegraphics[width=0.24\linewidth]{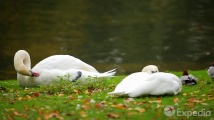}\\	
		\end{tabular}
	}
\end{minipage}
	\quad
	\begin{minipage}[t]{\linewidth}
	\centering
	\subfloat[Results of Adaptive Encoder + HTM + WT + Beam Search]{
		\begin{tabular}[t]{c}
			\includegraphics[width=0.24\linewidth]{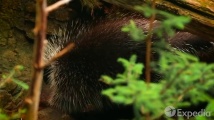}
			\includegraphics[width=0.24\linewidth]{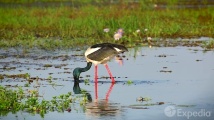}
			\includegraphics[width=0.24\linewidth]{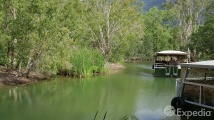}
			\includegraphics[width=0.24\linewidth]{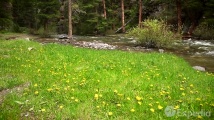}\\	
		\end{tabular}
	}
\end{minipage}
\vspace{-10pt}
	\caption{\textbf{Comparison of results on all the four models.}
	Query text: \emph{Then head to the marina to see these sea creatures and mammals in the wild.}
	This example illustrates a failure case of the proposed methods.
	The clip in (a) is considered more precisely described the visuals in text query, 
	\ie, more related to \emph{marina}.
	While for (b) - (d), the retrieved contents are more related to ``creatures in the wetland''.
	The sequencing styles in (c) and (d) are better than that in (b) 
	because the content transition is more smooth. 
	}
	\label{fig:vis_all_marina}
\end{figure*}

\begin{figure*}[!t]
\captionsetup[subfloat]{captionskip=-3pt}
	\centering
	\begin{minipage}[t]{\linewidth}
		\centering
		\subfloat[Results of Parallel Encoder + HTM + Greedy Search]{
			\begin{tabular}[t]{c}
				\includegraphics[width=0.24\linewidth]{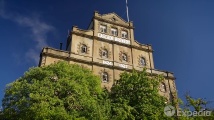}
				\includegraphics[width=0.24\linewidth]{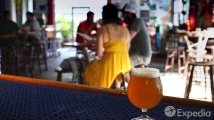}
				\includegraphics[width=0.24\linewidth]{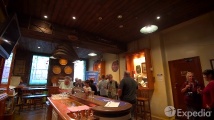}
				\includegraphics[width=0.24\linewidth]{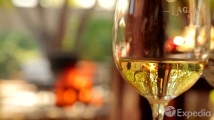}\\
			\end{tabular}
		}
	\end{minipage}
	\quad
	\begin{minipage}[t]{\linewidth}
	\centering
	\subfloat[Results of Parallel Encoder + HTM + WT + Greedy Search]{
		\begin{tabular}[t]{c}
			\includegraphics[width=0.24\linewidth]{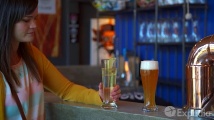}
			\includegraphics[width=0.24\linewidth]{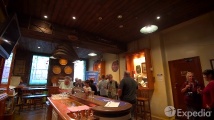}
			\includegraphics[width=0.24\linewidth]{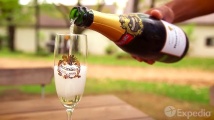}
			\includegraphics[width=0.24\linewidth]{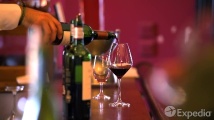}\\	
		\end{tabular}
	}
\end{minipage}
	\quad
	\begin{minipage}[t]{\linewidth}
	\centering
	\subfloat[Results of Parallel Encoder + HTM + WT + Beam Search]{
		\begin{tabular}[t]{c}
			\includegraphics[width=0.24\linewidth]{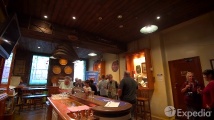}
			\includegraphics[width=0.24\linewidth]{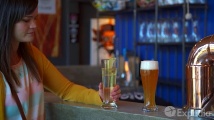}
			\includegraphics[width=0.24\linewidth]{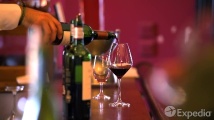}
			\includegraphics[width=0.24\linewidth]{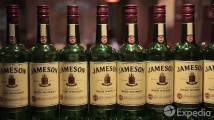}\\	
		\end{tabular}
	}
\end{minipage}
	\quad
	\begin{minipage}[t]{\linewidth}
	\centering
	\subfloat[Results of Apdaptive Encoder + HTM + WT + Beam Search]{
		\begin{tabular}[t]{c}
			\includegraphics[width=0.24\linewidth]{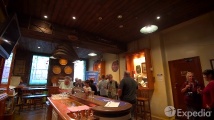}
			\includegraphics[width=0.24\linewidth]{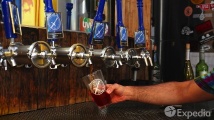}
			\includegraphics[width=0.24\linewidth]{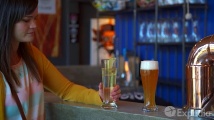}
			\includegraphics[width=0.24\linewidth]{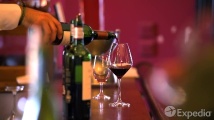}\\	
		\end{tabular}
	}
\end{minipage}
\vspace{-10pt}
	\caption{\textbf{Comparison of results on all the four models.}
	Query text: \emph{and share refreshing ales and conversation with fellow beer lovers.}
	This example reveals the following findings: 
	(i) Without finetuning on WeTravel, the generated clip may contain more irrelevant information (see (a) and (b)).
	(ii) Clips in (c) and (d) have better sequencing styles compared with (b). The clip in (b) jumps from inside scene
	to outside scene and jumps back across the shots.
	(iii) Results in (c) and (d) are both good with matched visual content and valid sequencing styles.
	}
	\label{fig:vis_all_beer}
\end{figure*}

\begin{figure*}[!t]
\captionsetup[subfloat]{captionskip=-3pt}
	\centering
	\begin{minipage}[t]{\linewidth}
		\centering
		\subfloat[Results of Parallel Encoder + HTM + Greedy Search]{
			\begin{tabular}[t]{c}
				\includegraphics[width=0.24\linewidth]{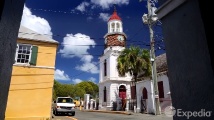}
				\includegraphics[width=0.24\linewidth]{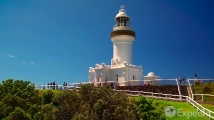}
				\includegraphics[width=0.24\linewidth]{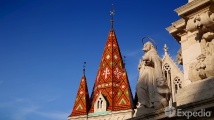}
				\includegraphics[width=0.24\linewidth]{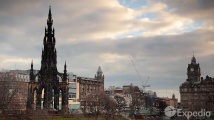}\\
			\end{tabular}
		}
	\end{minipage}
	\quad
	\begin{minipage}[t]{\linewidth}
	\centering
	\subfloat[Results of Parallel Encoder + HTM + WT + Greedy Search]{
		\begin{tabular}[t]{c}
			\includegraphics[width=0.24\linewidth]{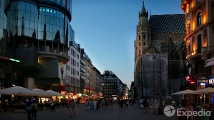}
			\includegraphics[width=0.24\linewidth]{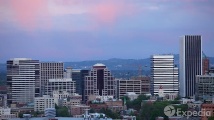}
			\includegraphics[width=0.24\linewidth]{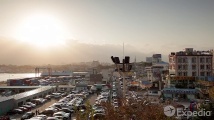}
			\includegraphics[width=0.24\linewidth]{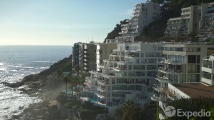}\\	
		\end{tabular}
	}
\end{minipage}
	\quad
	\begin{minipage}[t]{\linewidth}
	\centering
	\subfloat[Results of Parallel Encoder + HTM + WT + Beam Search]{
		\begin{tabular}[t]{c}
			\includegraphics[width=0.24\linewidth]{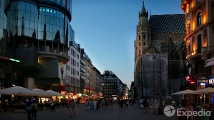}
			\includegraphics[width=0.24\linewidth]{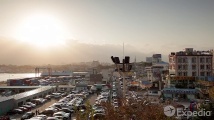}
			\includegraphics[width=0.24\linewidth]{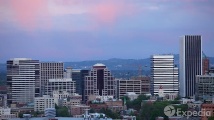}
			\includegraphics[width=0.24\linewidth]{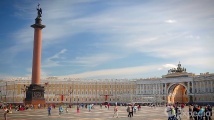}\\	
		\end{tabular}
	}
\end{minipage}
	\quad
	\begin{minipage}[t]{\linewidth}
	\centering
	\subfloat[Results of Adaptive Encoder + HTM + WT + Beam Search]{
		\begin{tabular}[t]{c}
			\includegraphics[width=0.24\linewidth]{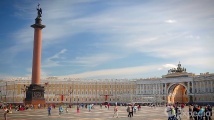}
			\includegraphics[width=0.24\linewidth]{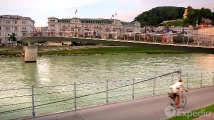}
			\includegraphics[width=0.24\linewidth]{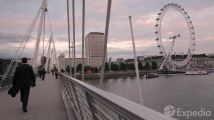}
			\includegraphics[width=0.24\linewidth]{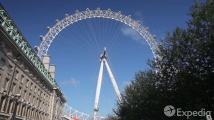}\\	
		\end{tabular}
	}
\end{minipage}
\vspace{-10pt}
	\caption{\textbf{Comparison of results on all the four models.}
	Query text: \emph{or watch the sun splash the buildings and skywheels across the sky.}
    This example illustrates the effectiveness of proposed content retrieval module.
    Although the clips in (a) - (c) demonstrates the concept of ``sun splash the buildings'' but fail
    to display ``skywheels across the sky''. 
    The query content is relatively complex and the Adaptive Encoder is helpful for dynamically
    retrieving the shots acknowledging the existing ones.
	}
	\label{fig:vis_all_sky}
\end{figure*}

%% file: supp_articles/fig_mil_nce_i.tex
\begin{figure*}[!t]
\vspace{-20pt}
\captionsetup[subfloat]{captionskip=-3pt}
	\centering
	\begin{minipage}[t]{\linewidth}
		\centering
		\subfloat[Query:\textbf{Explore the wind-swept bluffs and secluded coves of the Park.}
		The original captions for each shot are: (i) Hobart is a place where the old and new meet, right at the edge a pristine wilderness. It’s a city where cutting-edge art lives amongst historic sandstone warehouses; (ii) to sheltered beaches like Petit Bot Bay.
		(iii) far vaster and more beautiful than any built by the hand of man. (iv) just for you.]{
			\begin{tabular}[t]{c}
				\includegraphics[width=0.2\linewidth]{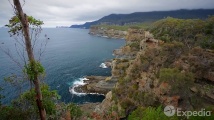}
				\includegraphics[width=0.2\linewidth]{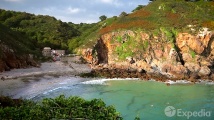}
				\includegraphics[width=0.2\linewidth]{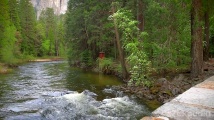}
				\includegraphics[width=0.2\linewidth]{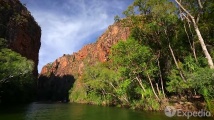}\\
			\end{tabular}
		}
	\end{minipage}
	\quad
	\begin{minipage}[t]{\linewidth}
	\centering
	\subfloat[Query:\textbf{This market sells fresh seafood: shellfish, shrimp and crab.}
	The original captions for each shot are: (i) From the shellfish which drew Paleolithic gatherers to its rugged shores, ... Jagalchi Fish Market; (ii) Nil; (iii) Or just make a selection from restaurants that have been serving up the fruits of these waters for generations. (iv) and maybe, pick up a few snacks for later.]{
		\begin{tabular}[t]{c}
			\includegraphics[width=0.2\linewidth]{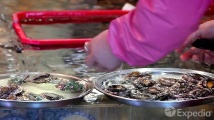}
			\includegraphics[width=0.2\linewidth]{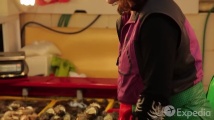}
			\includegraphics[width=0.2\linewidth]{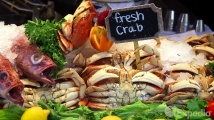}
			\includegraphics[width=0.2\linewidth]{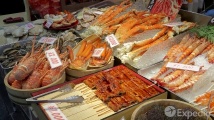}\\	
		\end{tabular}
	}
\end{minipage}
	\quad
	\begin{minipage}[t]{\linewidth}
	\centering
	\subfloat[Query:\textbf{The sea is beautiful with houses built upon the rock.}
	The original captions for each shot are: (i) While Guernsey’s dramatic coastline often steals the show, you’ll find plenty to inspire inland too. (ii) and most spectacular stretches of undeveloped coastline in the USA. ... (iii) When the Venetian warships finally arrived, laden with men and materials, ... (iv) Dubrovnik’s citizens mobilized and within months the first walls of Fort Lawrence were in place. ...]{
		\begin{tabular}[t]{c}
			\includegraphics[width=0.2\linewidth]{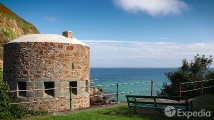}
			\includegraphics[width=0.2\linewidth]{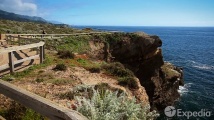}
			\includegraphics[width=0.2\linewidth]{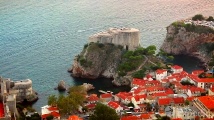}
			\includegraphics[width=0.2\linewidth]{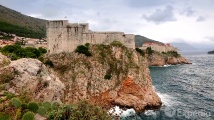}\\	
		\end{tabular}
	}
\end{minipage}
	\quad
	\begin{minipage}[t]{\linewidth}
	\centering
	\subfloat[Query:\textbf{a hearty breakfast is all you will need to freshen up.}
	The original captions for each shot are: (i) For classical Austrian fare served with a side of Opera, take the short walk to Stiftskeller St. Peter (ii) Pair each course with a different drop  for unexpected bursts of flavor. (ii) and there are many vineyards lying within the city boundaries where you can enjoy traditional meals like Wiener Schnitzel matched with local wines.
	(iv) ...restaurants that use locally grown ingredients and offer stunning views.]{
		\begin{tabular}[t]{c}
			\includegraphics[width=0.2\linewidth]{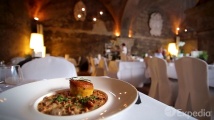}
			\includegraphics[width=0.2\linewidth]{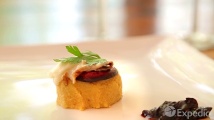}
			\includegraphics[width=0.2\linewidth]{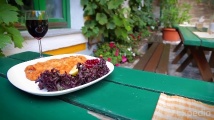}
			\includegraphics[width=0.2\linewidth]{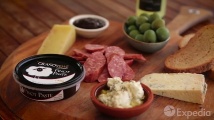}\\	
		\end{tabular}
	}
\end{minipage}
	\quad
	\begin{minipage}[t]{\linewidth}
	\centering
	\subfloat[Query:\textbf{and take the easy 1-mile walk through dense old-growth forests to the Beach.}
	The original captions for each shot are: (i) Those who do, will be embraced by Mother Nature, at her wildest, purest and most beautiful best. (ii) ...rainforest, drop in at a crocodile farm to see small fresh water crocs (iii) Take a short boat ride from Soufriere to one of the island’s finest resort beaches, Anse Chastanet. (iv) its ancient Banyan trees and attractive beachfront. The park is home to the historic Honolulu Zoo.]{
		\begin{tabular}[t]{c}
			\includegraphics[width=0.2\linewidth]{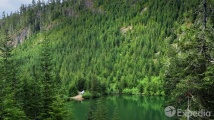}
			\includegraphics[width=0.2\linewidth]{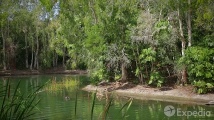}
			\includegraphics[width=0.2\linewidth]{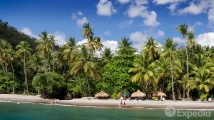}
			\includegraphics[width=0.2\linewidth]{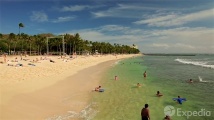}\\	
		\end{tabular}
	}
\end{minipage}
	\quad
	\begin{minipage}[t]{\linewidth}
	\centering
	\subfloat[Query:\textbf{People come for the Japanese style street food.}
	The original captions for each shot are: (i) Nil; (ii) Osakans love to say, kui-daore, which means, eat til you drop
	(iii) Osaka is a city to be feasted upon, and shared…over and over, day after day, until you drop!
	(iv) But for the ultimate in retro eye-candy, take the ten-minute walk south to the bright lights of Shin-sekai.]{
		\begin{tabular}[t]{c}
			\includegraphics[width=0.2\linewidth]{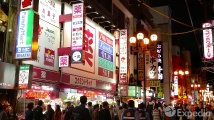}
			\includegraphics[width=0.2\linewidth]{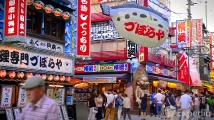}
			\includegraphics[width=0.2\linewidth]{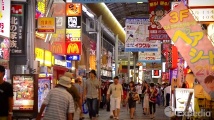}
			\includegraphics[width=0.2\linewidth]{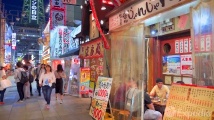}\\	
		\end{tabular}
	}
\end{minipage}
	\quad
	\begin{minipage}[t]{\linewidth}
	\centering
	\subfloat[Query:\textbf{Climb up the snow mountain and enjoy the beautiful ice world.}
	The original captions for each shot are: (i) Nil; (ii) Nil; (iii) Nil; (iv) While you're here take a ride into the Canadian Rockies on the Banff Gondola to see panoramic views of beautiful Bow Valley and Bow River, as well as the Banff township below.]{
		\begin{tabular}[t]{c}
			\includegraphics[width=0.2\linewidth]{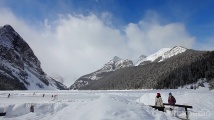}
			\includegraphics[width=0.2\linewidth]{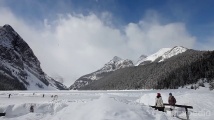}
			\includegraphics[width=0.2\linewidth]{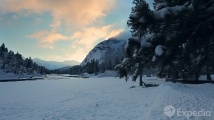}
			\includegraphics[width=0.2\linewidth]{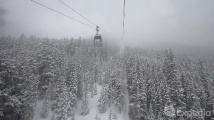}\\	
		\end{tabular}
	}
\end{minipage}
\vspace{-10pt}
	\caption{\small
	\textbf{Results of Adaptive Encoder + HTM + WT + Beam Search.} According to the authors' estimation, there are about $60\%$
	of the texts are not aligned with the visual content. Hence we also present the original transcript texts attached in the retrieved shots.}
	\label{fig:vis_final}
	\vspace{-30pt}
\end{figure*}

%% file: supp_articles/fig_whole_test_set.tex
\begin{figure*}[!t]

\captionsetup[subfloat]{captionskip=-3pt}
	\centering
	\begin{minipage}[t]{\linewidth}
		\centering
		\subfloat[Query:\textbf{a broad hill laced with trails, gardens and historic treasures.}]{
			\begin{tabular}[t]{c}
				\includegraphics[width=0.24\linewidth]{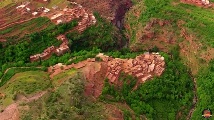}
				\includegraphics[width=0.24\linewidth]{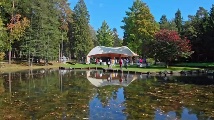}
				\includegraphics[width=0.24\linewidth]{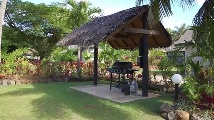}
				\includegraphics[width=0.24\linewidth]{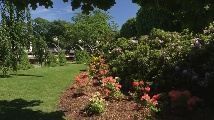}\\
			\end{tabular}
		}
	\end{minipage}
	\quad
	\begin{minipage}[t]{\linewidth}
	\centering
	\subfloat[Query:\textbf{a city that continues to work hard, play harder, and eat, like there's no tomorrow!}]{
		\begin{tabular}[t]{c}
			\includegraphics[width=0.24\linewidth]{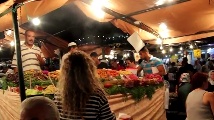}
			\includegraphics[width=0.24\linewidth]{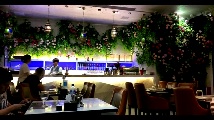}
			\includegraphics[width=0.24\linewidth]{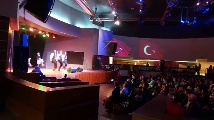}
			\includegraphics[width=0.24\linewidth]{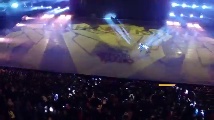}\\	
		\end{tabular}
	}
\end{minipage}
	\quad
	\begin{minipage}[t]{\linewidth}
	\centering
	\subfloat[Query:\textbf{and in the city's most popular shopping street.}]{
		\begin{tabular}[t]{c}
			\includegraphics[width=0.24\linewidth]{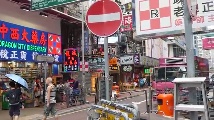}
			\includegraphics[width=0.24\linewidth]{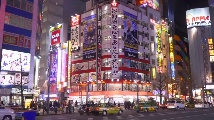}
			\includegraphics[width=0.24\linewidth]{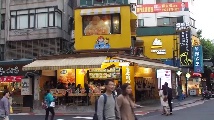}
			\includegraphics[width=0.24\linewidth]{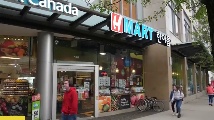}\\	
		\end{tabular}
	}
\end{minipage}
	\quad
	\begin{minipage}[t]{\linewidth}
	\centering
	\subfloat[Query:\textbf{cool off in one of the many plazas and let the city come to you.}]{
		\begin{tabular}[t]{c}
			\includegraphics[width=0.24\linewidth]{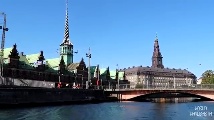}
			\includegraphics[width=0.24\linewidth]{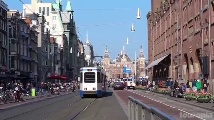}
			\includegraphics[width=0.24\linewidth]{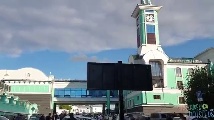}
			\includegraphics[width=0.24\linewidth]{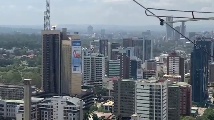}\\	
		\end{tabular}
	}
\end{minipage}
	\quad
	\begin{minipage}[t]{\linewidth}
	\centering
	\subfloat[Query:\textbf{Eating, drinking, chating, what a wonderful life!}]{
		\begin{tabular}[t]{c}
			\includegraphics[width=0.24\linewidth]{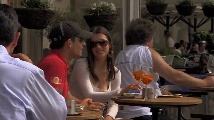}
			\includegraphics[width=0.24\linewidth]{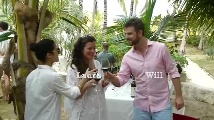}
			\includegraphics[width=0.24\linewidth]{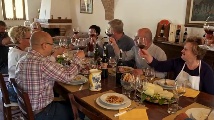}
			\includegraphics[width=0.24\linewidth]{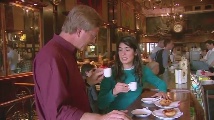}\\	
		\end{tabular}
	}
\end{minipage}
	\quad
	\begin{minipage}[t]{\linewidth}
	\centering
	\subfloat[Query:\textbf{Enjoy the golden sunrise and sunset at anywhere.}]{
		\begin{tabular}[t]{c}
			\includegraphics[width=0.24\linewidth]{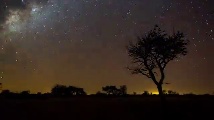}
			\includegraphics[width=0.24\linewidth]{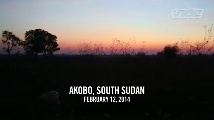}
			\includegraphics[width=0.24\linewidth]{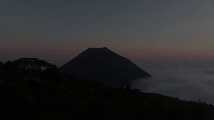}
			\includegraphics[width=0.24\linewidth]{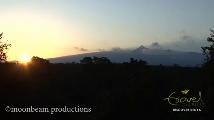}\\	
		\end{tabular}
	}
\end{minipage}
	\quad
	\begin{minipage}[t]{\linewidth}
	\centering
	\subfloat[Query:\textbf{explore the rock pools and the hidden caves.}]{
		\begin{tabular}[t]{c}
			\includegraphics[width=0.24\linewidth]{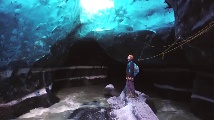}
			\includegraphics[width=0.24\linewidth]{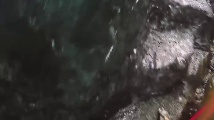}
			\includegraphics[width=0.24\linewidth]{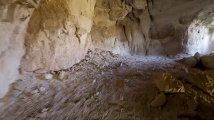}
			\includegraphics[width=0.24\linewidth]{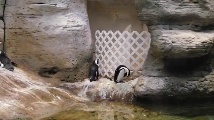}\\	
		\end{tabular}
	}
\end{minipage}
\vspace{-10pt}
	\caption{\small
	\textbf{Results of Adaptive Encoder + HTM + WT + Beam Search on the whole testing set with 1.02M gallery shots.}
	From the results we see that the proposed method is able to retrieve relevant shots to cover the visual information
	in the text queries as a whole. Also reasonable sequencing styles can be observed from the results.
	Even if the gallery is large with more than 1M shots, we argue that the inference speed is $<0.2$s in GPU or  $<2$s in CPU,
	making the whole inference pipeline practical appealing.
	}
	\label{fig:vis_whole_set_a}

\end{figure*}

\begin{figure*}[!t]

\captionsetup[subfloat]{captionskip=-3pt}
	\centering
	\begin{minipage}[t]{\linewidth}
		\centering
		\subfloat[Query:\textbf{From the port, take the ferry across the lagoon to the main island.}]{
			\begin{tabular}[t]{c}
				\includegraphics[width=0.24\linewidth]{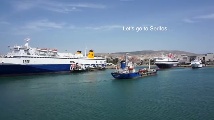}
				\includegraphics[width=0.24\linewidth]{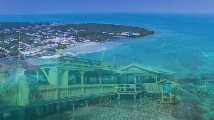}
				\includegraphics[width=0.24\linewidth]{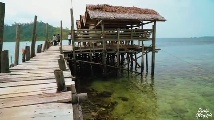}
				\includegraphics[width=0.24\linewidth]{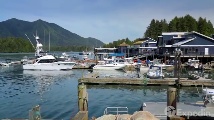}\\
			\end{tabular}
		}
	\end{minipage}
	\quad
	\begin{minipage}[t]{\linewidth}
	\centering
	\subfloat[Query:\textbf{or fill your basket with the fruits of Bordeaux's fields, forests and waters, at Capucins Market.}]{
		\begin{tabular}[t]{c}
			\includegraphics[width=0.24\linewidth]{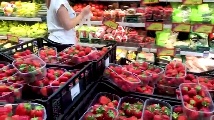}
			\includegraphics[width=0.24\linewidth]{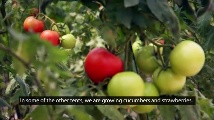}
			\includegraphics[width=0.24\linewidth]{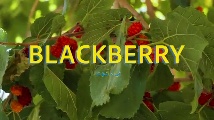}
			\includegraphics[width=0.24\linewidth]{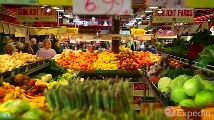}\\	
		\end{tabular}
	}
\end{minipage}
	\quad
	\begin{minipage}[t]{\linewidth}
	\centering
	\subfloat[Query:\textbf{People come for the Thailand style street market.}]{
		\begin{tabular}[t]{c}
			\includegraphics[width=0.24\linewidth]{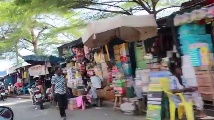}
			\includegraphics[width=0.24\linewidth]{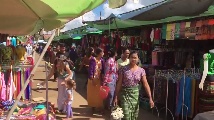}
			\includegraphics[width=0.24\linewidth]{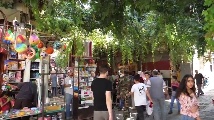}
			\includegraphics[width=0.24\linewidth]{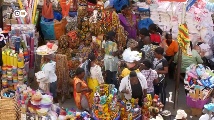}\\	
		\end{tabular}
	}
\end{minipage}
	\quad
	\begin{minipage}[t]{\linewidth}
	\centering
	\subfloat[Query:\textbf{Small animals prefer resting on the tree.}]{
		\begin{tabular}[t]{c}
			\includegraphics[width=0.24\linewidth]{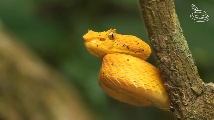}
			\includegraphics[width=0.24\linewidth]{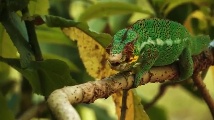}
			\includegraphics[width=0.24\linewidth]{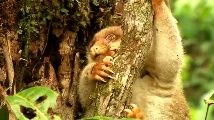}
			\includegraphics[width=0.24\linewidth]{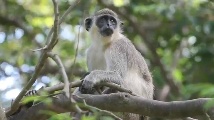}\\	
		\end{tabular}
	}
\end{minipage}
	\quad
	\begin{minipage}[t]{\linewidth}
	\centering
	\subfloat[Query:\textbf{The castle was built in the 16th century with protective walls.}]{
		\begin{tabular}[t]{c}
			\includegraphics[width=0.24\linewidth]{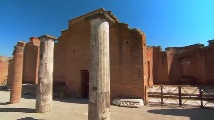}
			\includegraphics[width=0.24\linewidth]{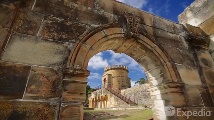}
			\includegraphics[width=0.24\linewidth]{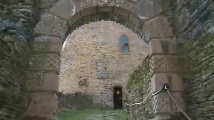}
			\includegraphics[width=0.24\linewidth]{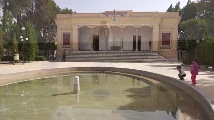}\\	
		\end{tabular}
	}
\end{minipage}
	\quad
	\begin{minipage}[t]{\linewidth}
	\centering
	\subfloat[Query:\textbf{The mountains are tall with picks near the clouds.}]{
		\begin{tabular}[t]{c}
			\includegraphics[width=0.24\linewidth]{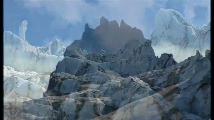}
			\includegraphics[width=0.24\linewidth]{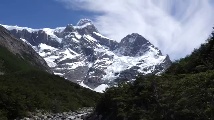}
			\includegraphics[width=0.24\linewidth]{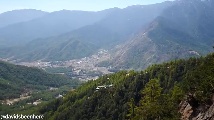}
			\includegraphics[width=0.24\linewidth]{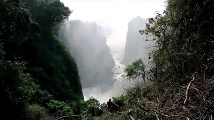}\\	
		\end{tabular}
	}
\end{minipage}
	\quad
	\begin{minipage}[t]{\linewidth}
	\centering
	\subfloat[Query:\textbf{Then, follow the roar of blowtorches to Kuromon Market, and feast upon mountains of BBQ.}]{
		\begin{tabular}[t]{c}
			\includegraphics[width=0.24\linewidth]{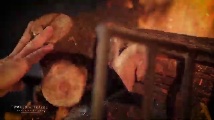}
			\includegraphics[width=0.24\linewidth]{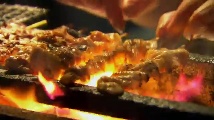}
			\includegraphics[width=0.24\linewidth]{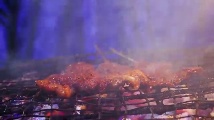}
			\includegraphics[width=0.24\linewidth]{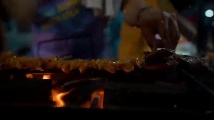}\\	
		\end{tabular}
	}
\end{minipage}
\vspace{-10pt}
	\caption{\small
	\textbf{Results of Adaptive Encoder + HTM + WT + Beam Search on the whole testing set with 1.02M gallery shots.}
	From the results we see that the proposed method is able to retrieve relevant shots to cover the visual information
	in the text queries as a whole. Also reasonable sequencing styles can be observed from the results.
	Even if the gallery is large with more than 1M shots, we argue that the inference speed is $<0.2$s in GPU or  $<2$s in CPU,
	making the whole inference pipeline practical appealing.
	}
	\label{fig:vis_whole_set_b}

\end{figure*}

%% file: supp_articles/fig_vimeo.tex
\begin{figure*}[!t]

\captionsetup[subfloat]{captionskip=-3pt}
	\centering
	\begin{minipage}[t]{\linewidth}
		\centering
		\subfloat[Query:\textbf{and share refreshing ales and conversation with fellow beer lovers.}]{
			\begin{tabular}[t]{c}
				\includegraphics[width=0.24\linewidth]{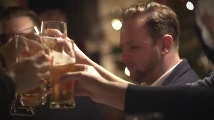}
				\includegraphics[width=0.24\linewidth]{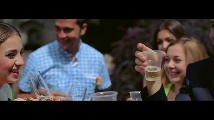}
				\includegraphics[width=0.24\linewidth]{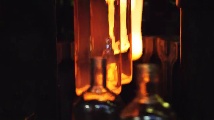}
				\includegraphics[width=0.24\linewidth]{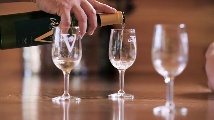}\\
			\end{tabular}
		}
	\end{minipage}
	\quad
	\begin{minipage}[t]{\linewidth}
	\centering
	\subfloat[Query:\textbf{and take the easy 1-mile walk through dense old-growth forests to the Beach.}]{
		\begin{tabular}[t]{c}
			\includegraphics[width=0.24\linewidth]{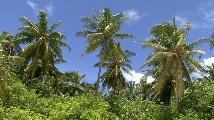}
			\includegraphics[width=0.24\linewidth]{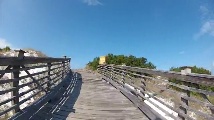}
			\includegraphics[width=0.24\linewidth]{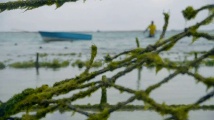}
			\includegraphics[width=0.24\linewidth]{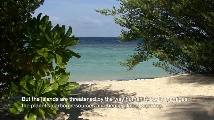}\\	
		\end{tabular}
	}
\end{minipage}
	\quad
	\begin{minipage}[t]{\linewidth}
	\centering
	\subfloat[Query:\textbf{As the trail leads you deep into the Park, breathe in the fresh scents of cedar.}]{
		\begin{tabular}[t]{c}
			\includegraphics[width=0.24\linewidth]{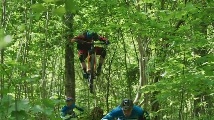}
			\includegraphics[width=0.24\linewidth]{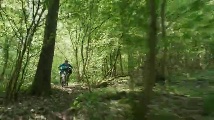}
			\includegraphics[width=0.24\linewidth]{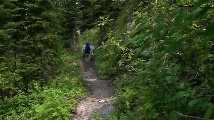}
			\includegraphics[width=0.24\linewidth]{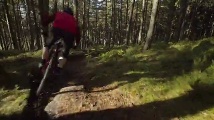}\\	
		\end{tabular}
	}
\end{minipage}
	\quad
	\begin{minipage}[t]{\linewidth}
	\centering
	\subfloat[Query:\textbf{For like everything they do, they have turned the humble snack into an art form.}]{
		\begin{tabular}[t]{c}
			\includegraphics[width=0.24\linewidth]{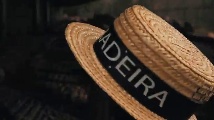}
			\includegraphics[width=0.24\linewidth]{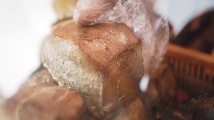}
			\includegraphics[width=0.24\linewidth]{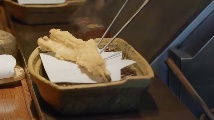}
			\includegraphics[width=0.24\linewidth]{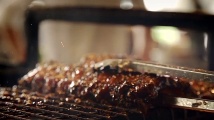}\\	
		\end{tabular}
	}
\end{minipage}
	\quad
	\begin{minipage}[t]{\linewidth}
	\centering
	\subfloat[Query:\textbf{Slow down in this colorful town and let the sizzling aromas of street food wash over you.}]{
		\begin{tabular}[t]{c}
			\includegraphics[width=0.24\linewidth]{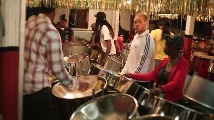}
			\includegraphics[width=0.24\linewidth]{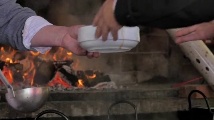}
			\includegraphics[width=0.24\linewidth]{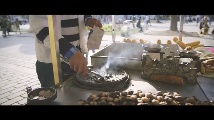}
			\includegraphics[width=0.24\linewidth]{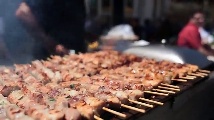}\\	
		\end{tabular}
	}
\end{minipage}
	\quad
	\begin{minipage}[t]{\linewidth}
	\centering
	\subfloat[Query:\textbf{The grounds are scented by the medicinal fragrance of lavender, laurel, lemon and other flowers.}]{
		\begin{tabular}[t]{c}
			\includegraphics[width=0.24\linewidth]{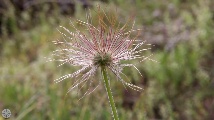}
			\includegraphics[width=0.24\linewidth]{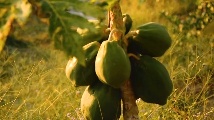}
			\includegraphics[width=0.24\linewidth]{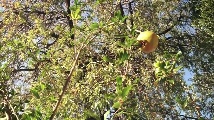}
			\includegraphics[width=0.24\linewidth]{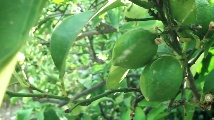}\\	
		\end{tabular}
	}
\end{minipage}
	\quad
	\begin{minipage}[t]{\linewidth}
	\centering
	\subfloat[Query:\textbf{Then head to the marina to see these sea creatures and mammals in the wild.}]{
		\begin{tabular}[t]{c}
			\includegraphics[width=0.24\linewidth]{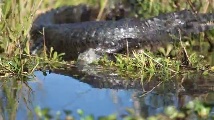}
			\includegraphics[width=0.24\linewidth]{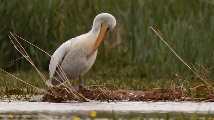}
			\includegraphics[width=0.24\linewidth]{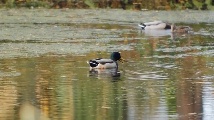}
			\includegraphics[width=0.24\linewidth]{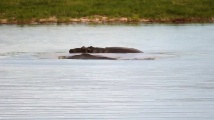}\\	
		\end{tabular}
	}
\end{minipage}
\vspace{-10pt}
	\caption{\small
	\textbf{Results of Adaptive Encoder + HTM + WT + Beam Search on Vimeo-HD with 450K gallery shots.}
	From the results we see that the proposed method is able to retrieve relevant shots to cover the visual information
	in the text queries as a whole. Also reasonable sequencing styles can be observed from the results.
	Even if the gallery is large with more than 1M shots, we argue that the inference speed is $<0.1$s in GPU or  $<1$s in CPU,
	making the whole inference pipeline practical appealing. 
	Besides, the Vimeo videos are uploaded by professional editors and are almost music oriented videos (without transcripts). 
	We train the models on YouTube videos and directly tested in Vimeo gallery shots, showing the generalization ability of our models.
	}
	\label{fig:vis_vimeo_a}

\end{figure*}

\begin{figure*}[!t]

\captionsetup[subfloat]{captionskip=-3pt}
	\centering
	\begin{minipage}[t]{\linewidth}
		\centering
		\subfloat[Query:\textbf{The mountains are tall with picks near the clouds.}]{
			\begin{tabular}[t]{c}
				\includegraphics[width=0.24\linewidth]{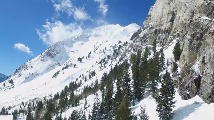}
				\includegraphics[width=0.24\linewidth]{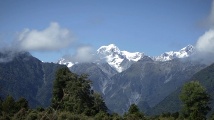}
				\includegraphics[width=0.24\linewidth]{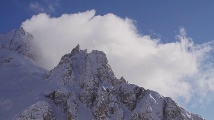}
				\includegraphics[width=0.24\linewidth]{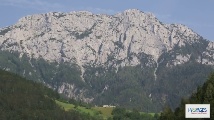}\\
			\end{tabular}
		}
	\end{minipage}
	\quad
	\begin{minipage}[t]{\linewidth}
	\centering
	\subfloat[Query:\textbf{People sing and dance to celebrate the wedding.}]{
		\begin{tabular}[t]{c}
			\includegraphics[width=0.24\linewidth]{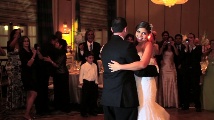}
			\includegraphics[width=0.24\linewidth]{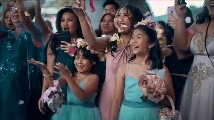}
			\includegraphics[width=0.24\linewidth]{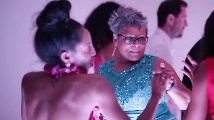}
			\includegraphics[width=0.24\linewidth]{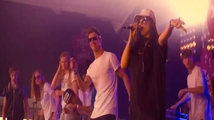}\\	
		\end{tabular}
	}
\end{minipage}
	\quad
	\begin{minipage}[t]{\linewidth}
	\centering
	\subfloat[Query:\textbf{Tons of fish and fresh seafood are sold in this market..}]{
		\begin{tabular}[t]{c}
			\includegraphics[width=0.24\linewidth]{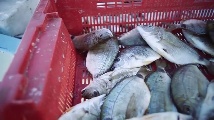}
			\includegraphics[width=0.24\linewidth]{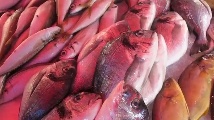}
			\includegraphics[width=0.24\linewidth]{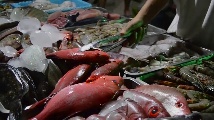}
			\includegraphics[width=0.24\linewidth]{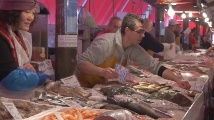}\\	
		\end{tabular}
	}
\end{minipage}
	\quad
	\begin{minipage}[t]{\linewidth}
	\centering
	\subfloat[Query:\textbf{the waves are big enough and people are surfing.}]{
		\begin{tabular}[t]{c}
			\includegraphics[width=0.24\linewidth]{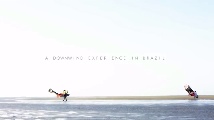}
			\includegraphics[width=0.24\linewidth]{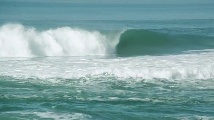}
			\includegraphics[width=0.24\linewidth]{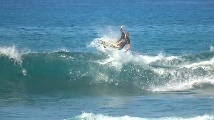}
			\includegraphics[width=0.24\linewidth]{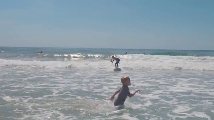}\\	
		\end{tabular}
	}
\end{minipage}
	\quad
	\begin{minipage}[t]{\linewidth}
	\centering
	\subfloat[Query:\textbf{any of various invertebrate animals resembling a plant such as a sea anemone or coral or sponge.}]{
		\begin{tabular}[t]{c}
			\includegraphics[width=0.24\linewidth]{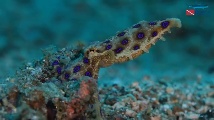}
			\includegraphics[width=0.24\linewidth]{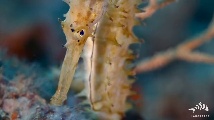}
			\includegraphics[width=0.24\linewidth]{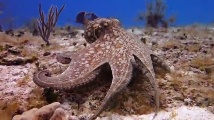}
			\includegraphics[width=0.24\linewidth]{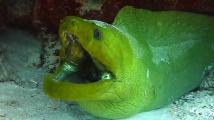}\\	
		\end{tabular}
	}
\end{minipage}
	\quad
	\begin{minipage}[t]{\linewidth}
	\centering
	\subfloat[Query:\textbf{The campground is about a 20 minute drive from here.}]{
		\begin{tabular}[t]{c}
			\includegraphics[width=0.24\linewidth]{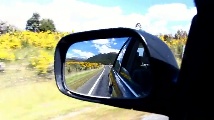}
			\includegraphics[width=0.24\linewidth]{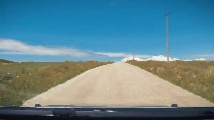}
			\includegraphics[width=0.24\linewidth]{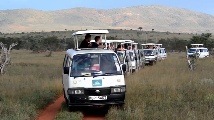}
			\includegraphics[width=0.24\linewidth]{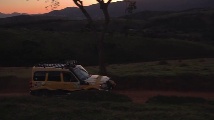}\\	
		\end{tabular}
	}
\end{minipage}
	\quad
	\begin{minipage}[t]{\linewidth}
	\centering
	\subfloat[Query:\textbf{We had to cross a large area of arid, featureless desert.}]{
		\begin{tabular}[t]{c}
			\includegraphics[width=0.24\linewidth]{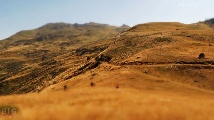}
			\includegraphics[width=0.24\linewidth]{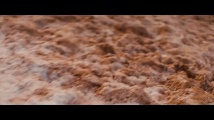}
			\includegraphics[width=0.24\linewidth]{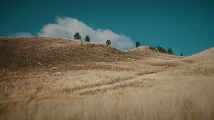}
			\includegraphics[width=0.24\linewidth]{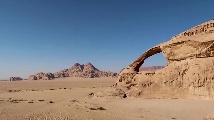}\\	
		\end{tabular}
	}
\end{minipage}
\vspace{-10pt}
	\caption{\small
    \textbf{Results of Adaptive Encoder + HTM + WT + Beam Search on Vimeo-HD with 450K gallery shots.}	
    From the results we see that the proposed method is able to retrieve relevant shots to cover the visual information
	in the text queries as a whole. Also reasonable sequencing styles can be observed from the results.
	Even if the gallery is large with more than 1M shots, we argue that the inference speed is $<0.1$s in GPU or  $<1$s in CPU,
	making the whole inference pipeline practical appealing.
	Besides, the Vimeo videos are uploaded by professional editors and are almost music oriented videos (without transcripts). 
	We train the models on YouTube videos and directly tested in Vimeo gallery shots, showing the generalization ability of our models.
	}
	\label{fig:vis_vimeo_b}

\end{figure*}